\documentclass{article}

\PassOptionsToPackage{numbers, compress}{natbib}

\usepackage[preprint]{neurips_2022}

\usepackage[utf8]{inputenc} 
\usepackage[T1]{fontenc}    
\usepackage[pagebackref,breaklinks,colorlinks]{hyperref}
\usepackage{url}            
\usepackage{booktabs}       
\usepackage{amsfonts}       
\usepackage{nicefrac}       
\usepackage{microtype}      
\usepackage{xcolor}         

\usepackage{graphicx}
\usepackage{amsmath, bm}
\usepackage{bbm}
\usepackage{breqn}
\usepackage{amsfonts}
\usepackage{algorithm,algorithmicx,algpseudocode}
\usepackage{dsfont}
\usepackage{caption}
\usepackage[position=b]{subcaption}
\usepackage{tabularx,booktabs}
\usepackage{subcaption}
\usepackage{wrapfig}

\usepackage{amsmath,amssymb,amsfonts}
\usepackage{scalerel,stackengine}

\newcommand\equalhat{\mathrel{\stackon[1.5pt]{=}{\stretchto{%
    \scalerel*[\widthof{=}]{\wedge}{\rule{1ex}{3ex}}}{0.5ex}}}}

\newtheorem{definition}{Definition}
\renewcommand*\backref[1]{\ifx#1\relax \else (cited on p. #1) \fi}

\usepackage{graphicx}
\usepackage{textcomp}
\usepackage{xcolor}
\usepackage{tabularx,booktabs}
\usepackage{tikz}
\usepackage[toc,page]{appendix}

\usepackage[utf8]{inputenc}
\usepackage[english]{babel}

\usepackage{soul}
\usepackage{xcolor}

\newcommand{\hlc}[2][yellow]{{%
    \colorlet{foo}{#1}%
    \sethlcolor{foo}\hl{#2}}%
}

\usepackage{wrapfig}

\usepackage{ragged2e}
\usepackage{mathtools, xparse}

\hypersetup{
	colorlinks = true,
    urlcolor  = cyan,
	citecolor = blue,
	linkcolor = blue}

\usetikzlibrary{positioning}
\usetikzlibrary{calc}

\title{Relational Local Explanations}

\author{%
  Vadim Borisov\thanks{Corresponding author: \href{mailto:vadim.borisov@uni-tuebingen.de}{vadim.borisov@uni-tuebingen.de}}  \\
  University of Tübingen\\
  \And
  Gjergji Kasneci\\
  University of Tübingen\\
}

\begin{document}

\maketitle

\begin{abstract}
The majority of existing post-hoc explanation approaches for machine learning models produce independent, per-variable feature attribution scores, ignoring a critical inherent characteristics of homogeneously structured data, such as visual or text data: there exist latent inter-variable relationships between features.
In response, we develop a novel model-agnostic and permutation-based feature attribution approach based on the relational analysis between input variables. As a result, we are able to gain a broader insight into the predictions and decisions of machine learning models. Experimental evaluations of our framework in comparison with state-of-the-art attribution techniques on various setups involving both image and text data modalities demonstrate the effectiveness and validity of our method. 
\end{abstract}

\section{Introduction}

The increasing reliance on machine learning (ML) models in various domains of our daily life has raised the need for explaining the inner workings automated prediction and decision-making processes based on such models \cite{xu2019explainable}. This is of particular relevance not only for deep convolutional neural networks (CNNs) in visual analytics domains, which have demonstrated superior performance on tasks such as object detection \cite{liu2020deep}, segmentation \cite{minaee2021image}, and classification \cite{he2016deep}, but also for natural language processing (NLP) applications, where self-attention models \cite{vaswani2017attention}, specifically deep Transformer-based models, have achieved state-of-the-art results on tasks such as text summarization, translation, and \cite{el2021automatic} and sentiment analysis \cite{birjali2021comprehensive}.


As a result, it is necessary to have confidence that black-box ML models are functioning as intended, and explanations that include \textit{inter-variable relational information} can help achieve this. Moreover, the interpretability of ML models is a vital aspect for numerous applications, particularly those evolving around life-critical use cases such as healthcare or autonomous driving~\cite{de2019local, zhang2020survey}.

Furthermore, in accordance with the General Data Protection Regulation (GDPR)~\cite{GDPR} and  California Consumer Privacy Act (CCPA) \cite{ccpa2021}, it is essential for real-world applications to not only provide accurate and reliable predictions but also to provide transparent and easy-to-understand explanations for automated decision systems. Additionally, there is a practical need for model-agnostic feature attribution methods that can be used with any machine learning system~\cite{bhatt2020explainable}. 

\textbf{Motivation.} Although numerous feature attribution approaches exist, the vast majority of them work with each input variable \textit{independently}, ignoring the latent relationships between attributes that exist in the many homogeneous data formats such as visual and textual ones. 

Another issue with the state-of-the-art feature attribution approaches is that many local explanation methods ``corrupt’’ a data sample to obtain local approximations of that sample \cite{ribeiro2016should, sundararajan2017axiomatic}, leading to the out-of-distribution problem \cite{slack2020fooling}. Further discussion on this topic is provided in Section \ref{sec:related_work} and Section \ref{sec:discussion}. 

In response, we propose a novel framework for post-hoc feature attributions that enables the generation of \textit{relational local explanations}. Our framework represents an input instance from a homogeneous dataset as a graph, where the features (or local feature combinations) represent the nodes and the pair-wise relations between features the edges. The weight of an edge then represents the value of the pair-wise local relationship between two features.

The proposed approach provides a two-fold consideration of feature attributions:
(1) The first one is that of general local explanations, which provides coefficients representing the influence of particular variables (words or a group of pixels) on the decision of the given ML model, positively or negatively related to other variables. 
(2) The second consideration is that of relational local explanations in the form of attention matrices, where the relationship between each input variable and other variables is represented as a coefficient. This type of explanation answers an important question -\textit{ How strongly is this variable related to all other variables?} - which adds another layer of depth to the explanation.

\begin{figure}[t]
\begin{tabular}{>{\centering\arraybackslash}m{3.6in}>{\centering\arraybackslash}m{1.5in}}
\centering
{\includegraphics[page=1,width=.57\textwidth]{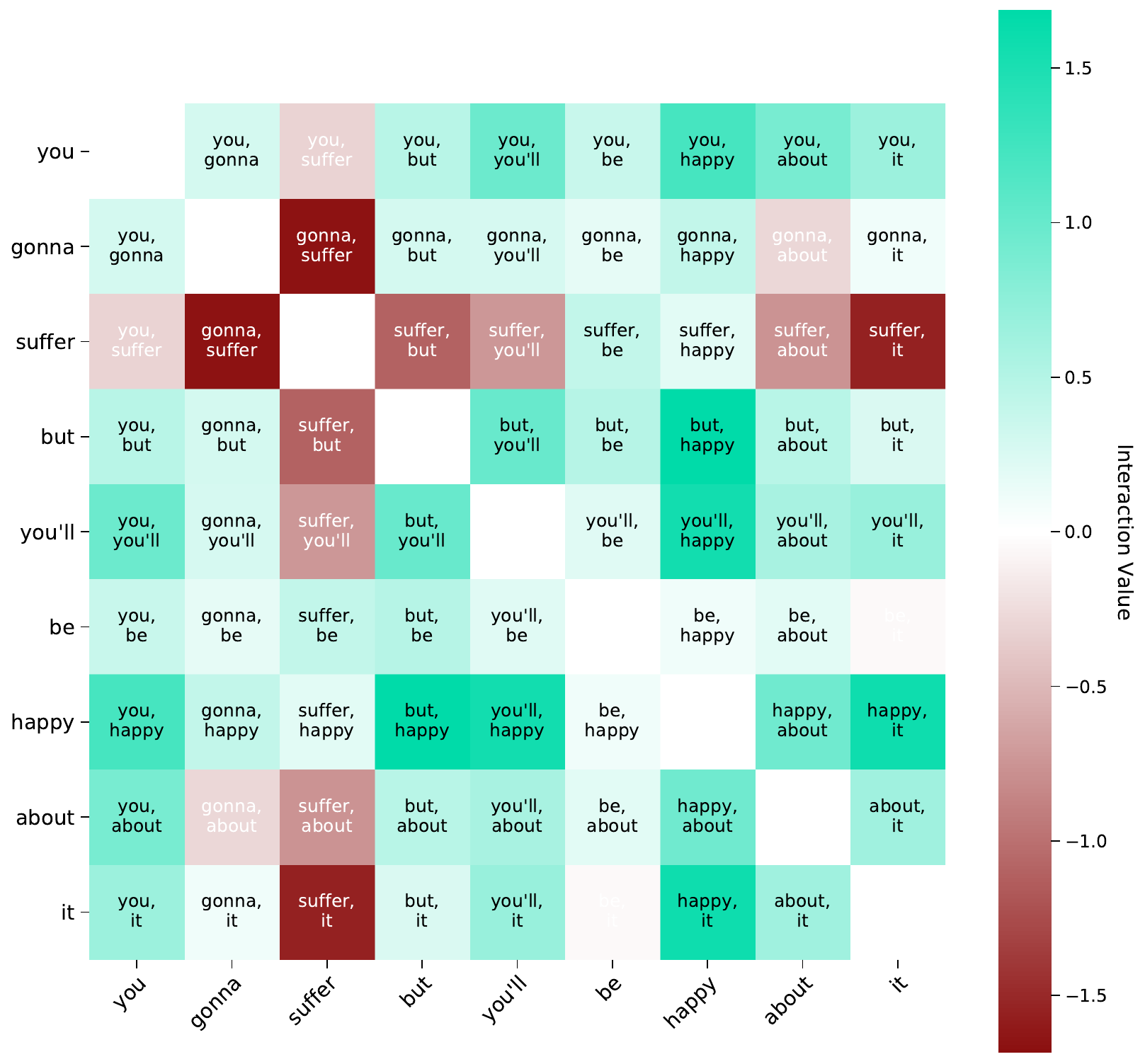}} & \texttt{you \hlc[red!50]{gonna} \hlc[red!50]{suffer} but \hlc[green!50]{you'll} be \hlc[green!50]{happy} about it}\\
\end{tabular}
\caption{An example of the relational local explanation (\textit{left}) and standard local explanation (\textit{right}) for textual data from the proposed RLE framework, where green color indicates positive influence and red negative. It can be seen that the relational local explanation allows the analysis of the pairwise influence of each word. For the task we select a pre-trained \texttt{DistilBERT} model \cite{sanh2019distilbert} for the sentiment analysis task. For more results, please refer to the Sec.~\ref{sec:experiments} and Appendix \ref{sec:app_experiments}.} 
\label{fig:example_text}
\end{figure}

\textbf{Contributions.} Below, we list the main contributions of our work:
\begin{itemize}
\item We highlight the importance of relational interactions between input features for local explanations. Since visual and textual data types are ``compositional’’ per nature i.e. the ``regional’’ information between variables naturally exists, it is crucial not only to understand what variable is important but also to spotlight and quantify the most critical combinations of variables in a given data sample. 

\item We develop a novel model-agnostic local feature attribution technique - coined Relational Local Explanations (RLE) - and formally describe it. To the best of our knowledge, this is the first model-agnostic local explanation algorithm based on relationships between input variables.

\item We extensively evaluate the proposed approach on image and text datasets and empirically show that it produces explanations that are superior to those produced by state-of-the-art attribution techniques.
\item We open-sourced the RLE implementation \url{https://github.com/unnir/rle}.
\end{itemize}

The remainder of this work is organized as follows. In Section~\ref{sec:related_work}, we give a short overview of the related methods for explaining machine learning models. Section~\ref{sec:rle} presents the proposed RLE algorithm. In Section~\ref{sec:experiments}, we visually and empirically compare our algorithm against other state-of-the-art feature attribution approaches. Finally, Section~\ref{sec:discussion} discusses the properties and limitations of the proposed method before concluding in Section~\ref{sec:conclusion}.

\section{Related Work}
\label{sec:related_work}

In recent years, there have been several studies that have focused on methods for explaining feature interactions and adjacency. Lundberg et al. \cite{lundberg2019explainable} propose an efficient local explanation method based on the SHAP framework \cite{SHAP} for decision tree-based models, which allows for the direct measurement of local feature interaction effects. Cui et al. \cite{cui2019learning} propose a probabilistic estimation method to assess the joint effect of two input features and the sum of their marginal effects in order to evaluate global feature pairwise interactions.

A number of studies have also explored feature interaction approaches specifically for deep neural networks (DNNs). For example, Greenside et al. \cite{greenside2018discovering} explore interactions between variables using deep feature interaction maps by calculating the difference between the attributions of two variables. Singh et al. \cite{singh2018hierarchical} present the generalization of the Contextual Decomposition \cite{murdoch2018beyond} to explain interactions for dense DNNs and CNNs.

More recently, Janizel et al. \cite{janizek2021explaining} propose an efficient method for local explanations based on feature interactions for DNNs called Integrated Hessians (IH). This method is based on an enhancement of the Integrated Gradients (IG) approach \cite{sundararajan2017axiomatic} and has been shown to produce trustworthy results. However, from a practical perspective, the Hessian matrix is significantly larger than gradient matrices and requires a sufficient memory size.




\textbf{Limitations of prior approaches.} Despite the progress made in understanding feature relationships through previous approaches, a major limitation of these methods is that they are often tied to specific model architectures or data types, making them not fully model agnostic. In addition, perturbation-based algorithms such as LIME \cite{ribeiro2016should}, SHAP \cite{SHAP}, and IG \cite{sundararajan2017axiomatic} rely on altering the data sample - and thus also altering the underlying distribution - in order to provide explanations. Our method, instead, aims to preserve as much information as possible by altering only the global structure of the input and not its features. We discuss this issue in more detail in Section \ref{sec:discussion}.


\begin{figure}[t]
  \centering
  \includegraphics[page=1,width=.5\textwidth]{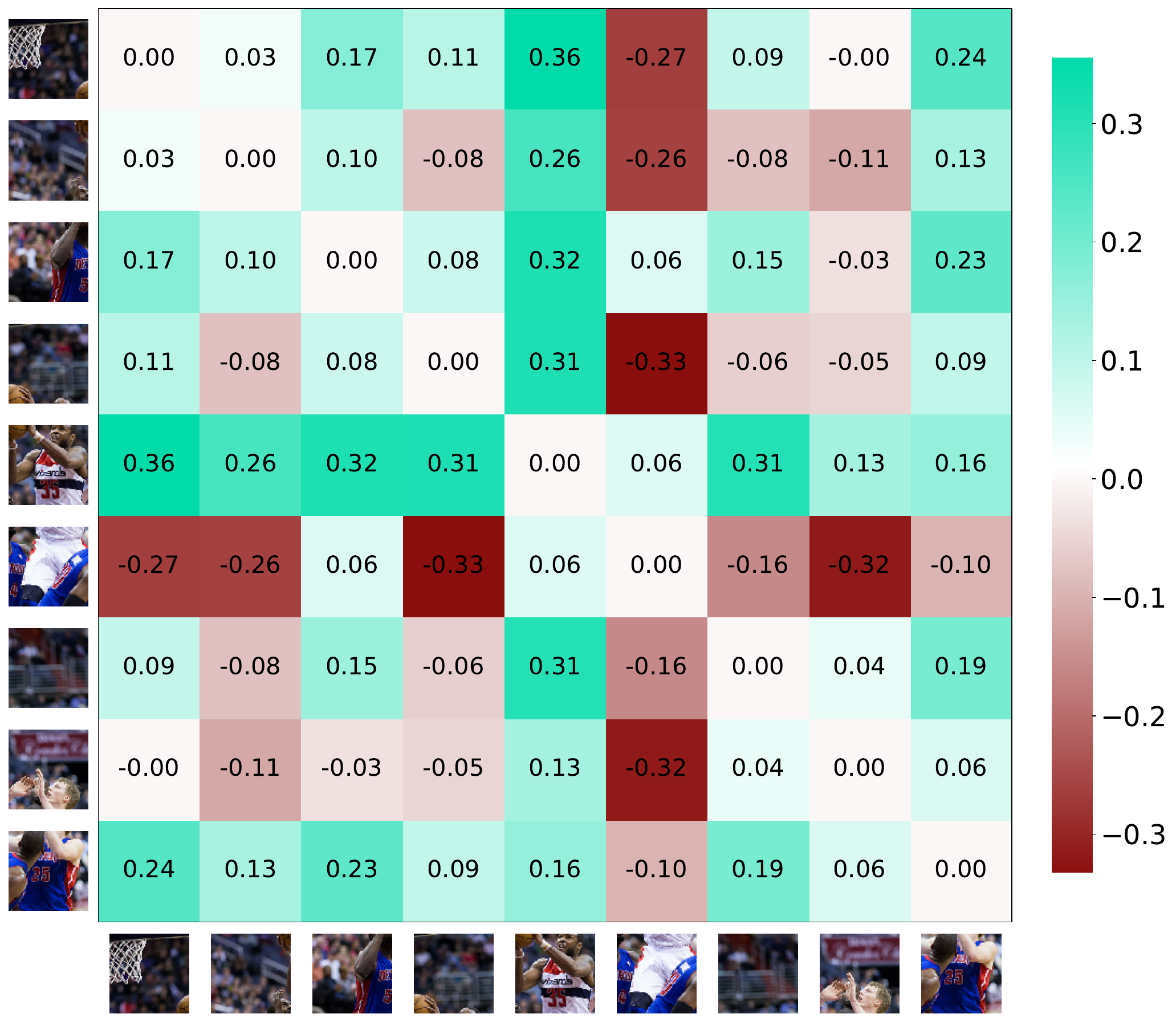}\hspace*{.05\textwidth}
  {\raisebox{15mm}{\includegraphics[width=.25\textwidth]{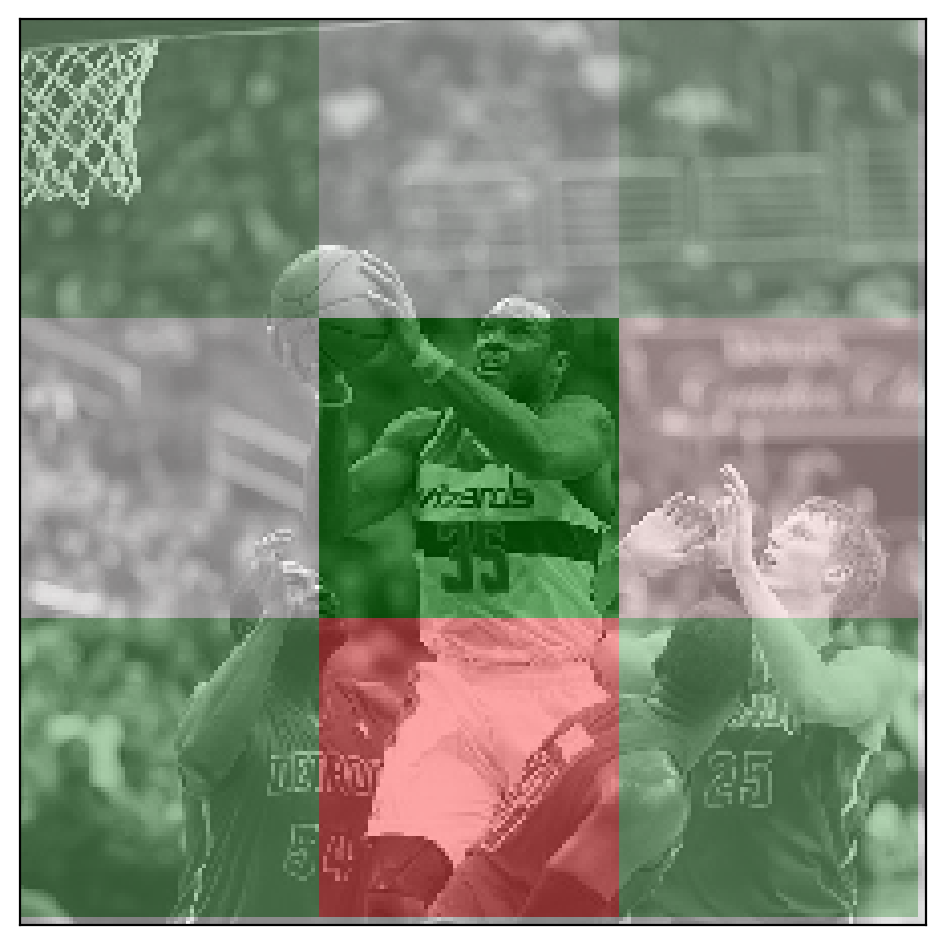}}}
  \caption{An example of the relational local explanation (\textit{left}) and standard local explanation (\textit{right}) for visual data from the proposed RLE framework, where green color indicates positive influence, and red negative. The relational local explanation can be used for a deeper feature analysis of the image data. For this example, the RLE approach is applied to a pre-trained ResNet-50 model~\cite{he2016deep} and an image of the class \texttt{basketball} from the ImageNet dataset~\cite{deng2009imagenet}, i.e., to uncover a combination of patches that had a high influence on the prediction of the model. For more results, we refer the reader to the Appendix.~\ref{sec:experiments}.}
  \label{fig:example_image}
\end{figure}

\section{Relational Local Explanation (RLE) Framework}
\label{sec:rle}

This section introduces the Relational Local Explanation (RLE) algorithm by first discussing its main components. In addition, we present how the RLE approach can be utilized for visual and textual data modalities.

\subsection{Formal definitions}

Before proceeding to the description of the proposed method, we introduce central definitions of our study. The definitions are based on existing works \cite{SHAP,sundararajan2017axiomatic}. 

\begin{definition}[Local Explanation]
\label{def:le}
A feature attribution function can be seen as  $\phi(f, \bm x, c_{\bm x})\in \mathbb{R}^{n}$, where $f: \mathbb{R}^{n} \to \mathbb{R}$ is a black-box model and $\bm x\in \mathbb{R}^{n}$ is an input sample belonging to a class $c_{\bm x}\subset \mathbb{R}$. The output of $\phi$ is an explanation representation vector $\mathbf{e}_{\bm x}\in\mathbb{R}^{n}$.
\end{definition}

Each element of $\mathbf{e}_{\bm x}$ is an importance score corresponding to a feature value in $\bm x$. A large positive or negative value in $\mathbf{e}_{\bm x}$ indicates that a corresponding feature greatly influences the outcome. Features with values close to zero in $\mathbf{e}_{\bm x}$ have little impact. Note that there are explanation methods that do not require a class specification; thus, for simpler and more general notation, a feature scoring function has the form $\phi(f, {\bm x})$. 

Taking the description of local explanations, we can extend it to the definition of relational local explanations.

\begin{definition}[Relational Local Explanation]
A relational local explanation function can be seen as $\Psi(f, {\bm x}, c_{x})\in \mathbb{R}^{n\times n}$, where $f: \mathbb{R}^{n} \to \mathbb{R}$ is a black-box model as above and ${\bm x}\in \mathbb{R}^{n}$ is an input sample belonging to a class $c_{\bm x}\subset \mathbb{R}$. The output of the $\Psi$ is an relational explanation representation in a form of a adjacency matrix $\mathbf{A}_{\bm x}$. 
\end{definition}

The relational local explanation matrix $\mathbf{A}_{\bm x}$ contains in each cell $\mathbf{A}_{\bm x}(i,j)$ the relational interaction between the $i$'th and $j$'th input feature. Note that $\mathbf{A}_{\bm x}$ is symmetric, i.e., $\mathbf{A}_x = \mathbf{A}^\top_{\bm x}$, and thus, the mean value of column or row elements corresponds to average local importance for the corresponding feature. Formally, let $\bar{\mathbf{A}}_{\bm x}$ denote the vector of mean values of the rows of matrix $\mathbf{A}_x$. Then:

\begin{equation}
    \bar{\mathbf{A}}_x \equalhat \mathbf{e}_{\bm x}.
\end{equation}

The symmetry property of relational local explanations is based on the assumption that the association between two variables has to be symmetrical. This was also indicated in the previous related works \cite{janizek2021explaining}.

\begin{algorithm}[t]
\caption{Relational Local Explanations (RLE)}
\label{alg:adjacency}
\begin{algorithmic}
\Require ML model $f$, Instance to explain $\bm x_o$, Number of permutations $n$
\State $\mathcal{X}^\prime \gets \{\}$ \Comment{New auxiliary dataset}
\For{$i \in \{1,2,3, ..., n\}$}
        \State ${\bm x}^p_i \gets permute({\bm x}_o)$ \Comment{Replace and shuffle the instance to explain}
        \State $\mathcal{G}_i \, \gets Graph({\bm x}^p_i)$ \Comment{Get the graph structure of patches}
        \State $\mathbf{A}_i \gets AdjacencyMatrix(\mathcal{G}_i)$ \Comment{Get the adjacency matrix}
        \State ${\bm x}^\prime \, \gets Lower(\mathbf{A}_i)$ \Comment{Get the lower triangle}
        \State $\mathcal{X}^\prime \gets \mathcal{X}^\prime \cup \langle x^\prime, f(\tilde{x}_i) \rangle$ 
\EndFor
\State $w \gets LinearModel(\mathcal{X}^\prime)$ \Comment{Train an explainable-by-design surrogate model}
\State \Return $w$
\end{algorithmic}
\end{algorithm}

\begin{figure}[t]
    \centering
    \includegraphics[width=\textwidth]{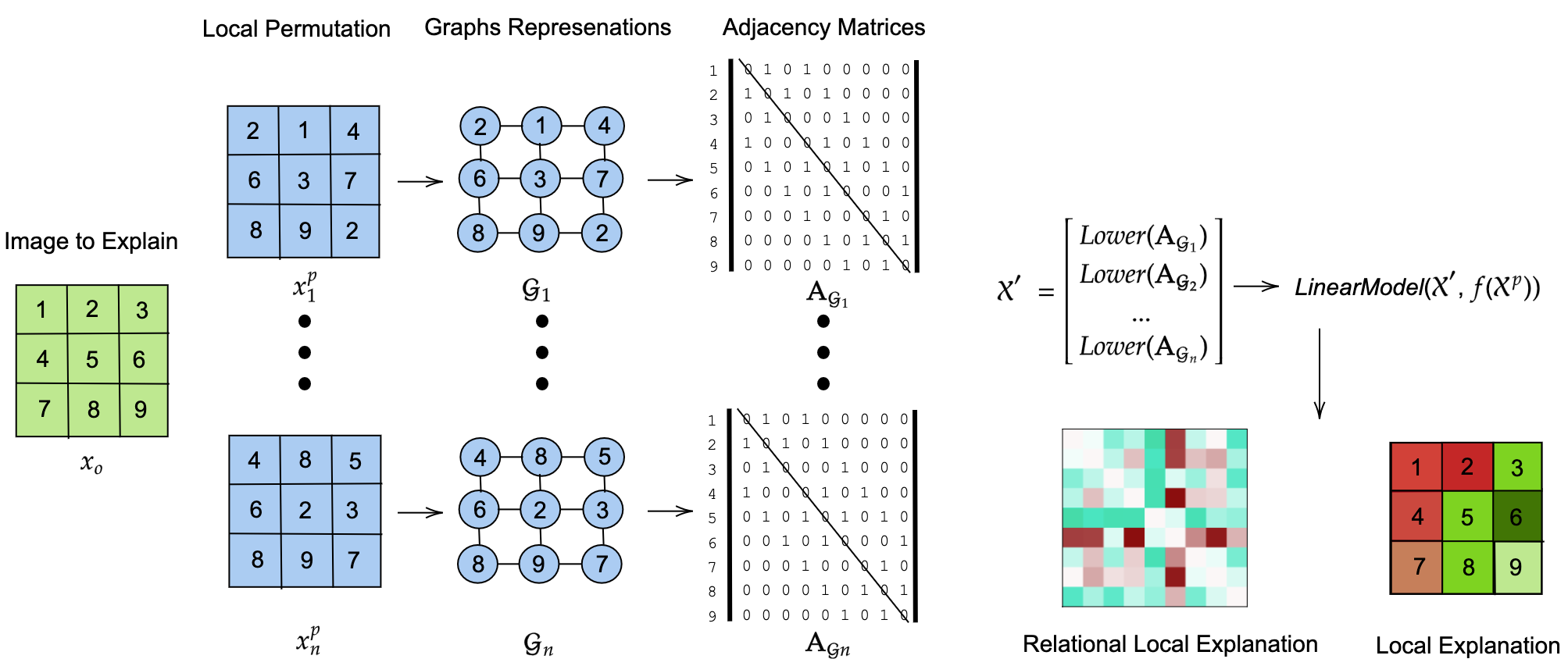}
    \caption{A relational local explanation of a data sample given a vision model using the RLE algorithm. Where $f$ is a black-box machine learning model to explain, $\bm x_o$ is a sample of interest, $\mathcal{G}_i$ is a graph of representation of a perturb image sample $x^p_i$.}
    \label{fig:rle_images}
\end{figure}

\subsection{RLE : The proposed framework}

Our approach follows common strategies for the generation of local explanations, as proposed in LIME \cite{ribeiro2016should}, SHAP \cite{SHAP}, and Anchors \cite{ribeiro2018anchors}, since they have a solid theoretical foundation~\cite{garreau2020explaining} and are quite popular in the ML community~\cite{xu2019explainable, dieber2020model}. 

The main idea of the RLE algorithm is to generate $n$ local permutations of a data sample to explain, then construct corresponding graph representations and adjacency matrices of the relationships between input features from the shuffled data. Thus we obtain a new dataset of local relations between features. Next, a linear model (that is explainable by design) is fitted to the new dataset - using information from the adjacency matrices to get the corresponding coefficients, which can be utilized for the relational local explanation.

Formally, for a given black-box model $f:\mathcal{X} \to \mathcal{Y}$ with $\mathcal{X}\subseteq \mathbb{R}^{n}$, $\mathcal{Y}\subset\mathbb{R}$, we may learn an interpretable \textit{surrogate} model $g$, which is a local approximation of $f$ for a given perturbation of a  particular input $\bm x_o \in \mathcal{X}$.
For this purpose, we first divide a data sample (image, text) into discrete elements of pixel patches for visual data, or groups of words for textual data. Then the chosen data sample $\bm x_o$ is perturbed $n$ times to generate $\bm x^p_{oi},  i=1..n$, whereby we randomly replace a single element (i.e., patch/group) $p_{oi}$ from $\bm x_{o}$ with another randomly selected element $p_{oj}$ from $\bm x_0$. Subsequently, an undirected graph representation of the shuffled sample is obtained as $\mathcal{G}_{x_o}^p$, where each vertex represents one of the discrete elements $\bm x^p_{oi},  i=1..n$ (e.g., superpixel or word), and each edge the relational connection between two such elements.

Further, an adjacency matrix $\mathbf{A}_{{{\mathcal{G}}_i}}$ for each $\bm x^p_{oi}$ is obtained. Since \textit{adjacency matrices} for undirected graphs are symmetric, only the lower triangle is utilized $Lower(\mathbf{A}_{{\mathcal{G}}_i})$ for the next step. 

Hence, the key idea of our approach is to permute a data sample but still keep the local features. We do not rely on feature perturbation techniques since they may lead to an undesired out-of-distribution problem \cite{slack2020fooling}; we examine this issue in detail in Section \ref{sec:discussion}. 

The procedure described above, as depicted in Figure~\ref{fig:rle_images}, leads to a new dataset $\mathcal{X}^\prime = \{ Lower(\mathbf{A}_{{\mathcal{G}}_i}), f(\bm x^p_i) \}^n_{i=1}$. We then learn a sparse linear regression $g_{\bm w_{\bm x_o}}(\bm x^p) = \bm w_{\bm x_0}^\top \bm x^p$ using the local dataset $\mathcal{X}^\prime$ by optimizing the following loss function with $\Omega(\cdot)$ as a measure of complexity.

\begin{equation}
w_{\bm x_0}=\mathop{\text{argmin}}_{\bm w} \mathcal{L}(f, g) + \Omega(\bm w),
\end{equation}
where $\mathcal{L}(f, g)$ is the mean squared loss,

\begin{equation}
\mathcal{L}(f, g) = \frac{1}{n}\sum\limits_{i=1}^n
\Big[f(x^{p}_{oi}) - g({\bm x}^{p}_{oi}, {\bm x}_0)\Big]^2.
\end{equation}

The RLE algorithm yields $g_{\bm w_{\bm x_0}}(\bm x^\prime)$, which approximates the complex model $f(\bm x^\prime)$ around $\bm x_0$.  In case $g$ is a linear model, the components of the weight vector $\bm w_{\bm x_0}$ indicate the relative influence of the relationship between features values of $\bm x_0$ based the sample $\mathcal{X}^\prime$ and can be used as the relational local explanation of $f(\bm x_0)$. The full approach is summarized in Algorithm \ref{alg:adjacency}. 

The following subsections discuss how the proposed algorithm can be adapted to visual and textual data modalities.

\subsection{Relational local explanations for multidimensional visual data}
\label{sec:visual_data}

\begin{figure}[t]
    \centering
    \includegraphics[width=\textwidth]{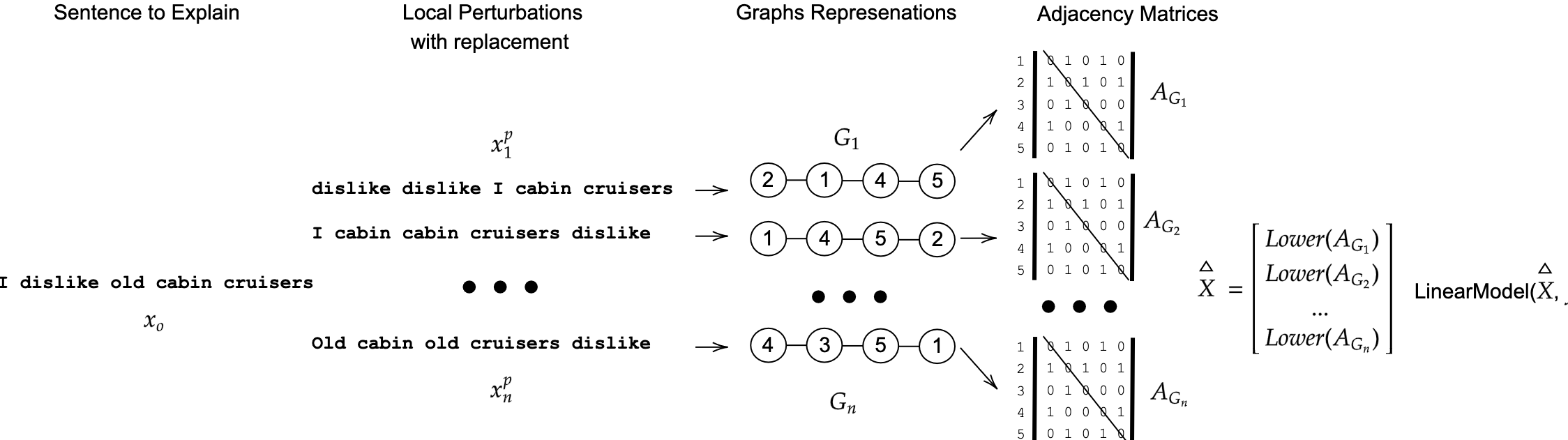}
    \caption{Relational local explanation of a data sample given a textual model using the RLE algorithm. Where $f$ is a black-box machine learning model to explain, $\bm x_o$ is a sentence of interest, $\mathcal{G}_i$ is a graph of representation of a perturbed sentence $x^p_i$.}
    \label{fig:rle_text}
\end{figure}

In the case of visual data, we divide an image into patches to disrupt the spatial layout of local image regions, this is a common approach for local explanation methods~\cite{ribeiro2016should, SHAP}. Formally, given an input image $I$, we first uniformly partition the image into $N\times N$ sub-regions denoted by $R_{i,j}$, where $i$ and $j$ are the horizontal and vertical indices respectively and $1 \leq i, j \leq N$. Thus, following this construction, a single patch $R_{i,j}$ represents a feature for the RLE algorithm. The procedure for an input image is presented in Fig \ref{fig:rle_images}. For textual data as input, the analogous procedure is depicted in Figure~\ref{fig:rle_text}.  

Note that the obtained graph representations for patched images do not consider the directed (e.g., causal) relationships between neighboring patches. But note, that the framework can be easily extended to include directional information; towards that goal, we may use the same binary encoding scheme for the adjacency matrix $\mathbf{A}_{{\mathcal{G}}_i}$, as depicted in the Figures~\ref{fig:rle_images,fig:rle_text}.

We argue that local details are much more important than the global structure for image recognition, as the differences between details are often the main driver of vision algorithms for distinguishing between different classes.
In most cases, various fine-grained categories tend to share similar global structures and vary only in specific local details \cite{chen2019destruction}; therefore, by random permutation, a given black-box model for computer vision is forced to focus on the local details. Multiple studies support our argument: the jigsaw puzzle pretext task for Self-Supervised Learning (SSL) approaches \cite{misra2020self}, a regularization scheme for Variational Autoencoders (VAEs) \cite{taghanaki2020jigsaw}, and last but not least for Vision Transformers (ViTs) follow the same strategy of dividing an image into patches as well \cite{dosovitskiy2020image, liu2021swin}. Additionally, also graph-based representations of visual data represent a common approach for many downstream tasks \cite{sanfeliu2002graph}.

\subsection{Relational local explanations for textual data}

For relational local explanations of  textual models, in particulate self-attention-based models \cite{vaswani2017attention}, we represent a sentence as a graph $\mathcal{G}_i$, where each word is expressed as a node. Then, as we presented before, we permute the sentence by changing the word order with replacement $n$ times. The whole procedure is illustrated in Figure \ref{fig:rle_text}. 
This operation allows learning the bidirectional context for a word. As a result, each position is contextualized, i.e., by the directional context from words occurring on its left and words occurring on its right. 

A similar permutation idea was successfully used for training an XLNet \cite{yang2019xlnet} and GReaT \cite{borisov2022language} Transformer models; thus, exploiting the insight that transformer models are able to learn and extract additional information from the semantics of a shuffled sentence. Furthermore, multiple studies \cite{battaglia2018relational, joshi2020transformers} made the same observation - a sentence can be seen as a graph, where words correspond to nodes and the computation of an attention score is the assignment of a weight to an edge between two words.


\begin{table*}[t]
\begin{tabularx}{\textwidth}{c @{\extracolsep{\fill}} c @{\extracolsep{\fill}} c@{\extracolsep{\fill}}c@{\extracolsep{\fill}} c@{\extracolsep{\fill}}}
\toprule
$\,\,$Original & IG \cite{sundararajan2017axiomatic} & LIME \cite{ribeiro2016should} &  SHAP  \cite{SHAP} & RLE (Ours) \\
\midrule

\includegraphics[width=23.9mm]{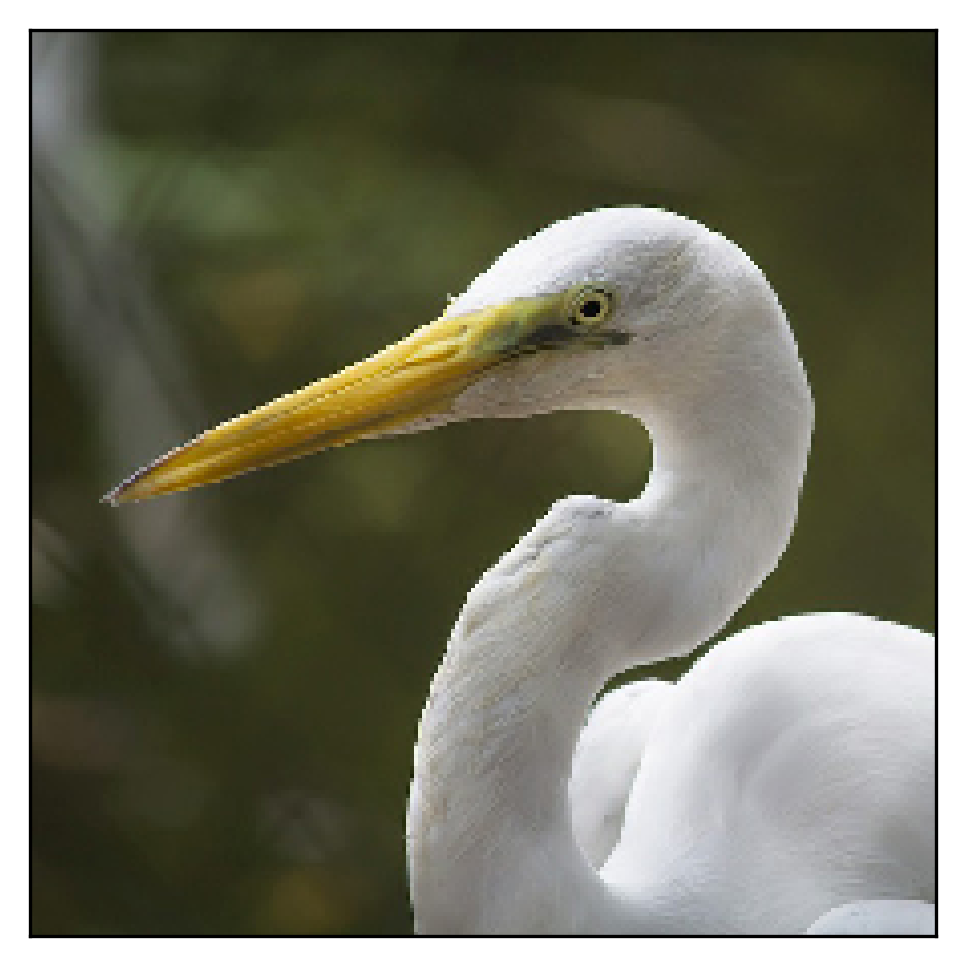} & 
\includegraphics[width=23.9mm]{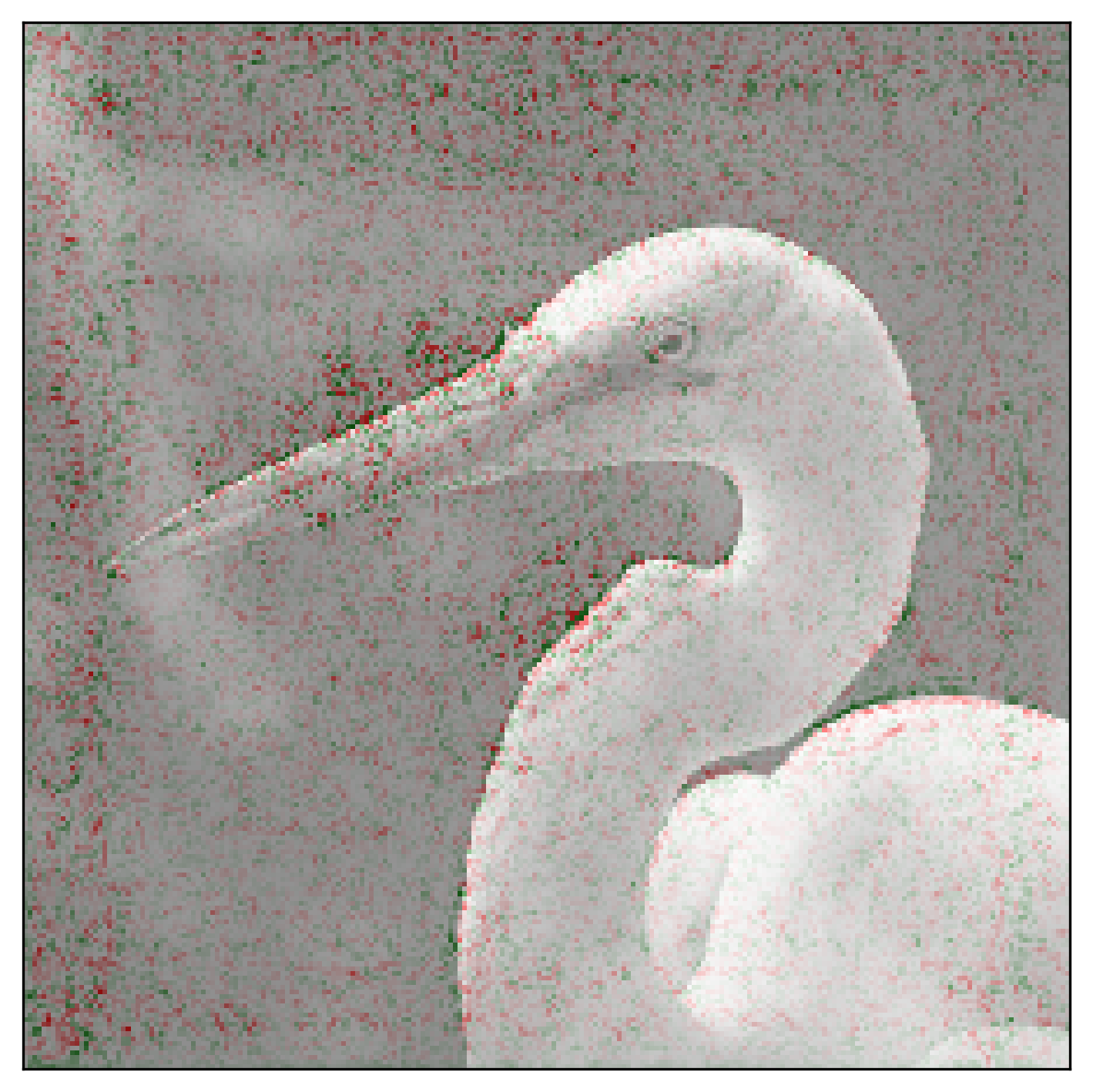} & 
\includegraphics[width=23.9mm]{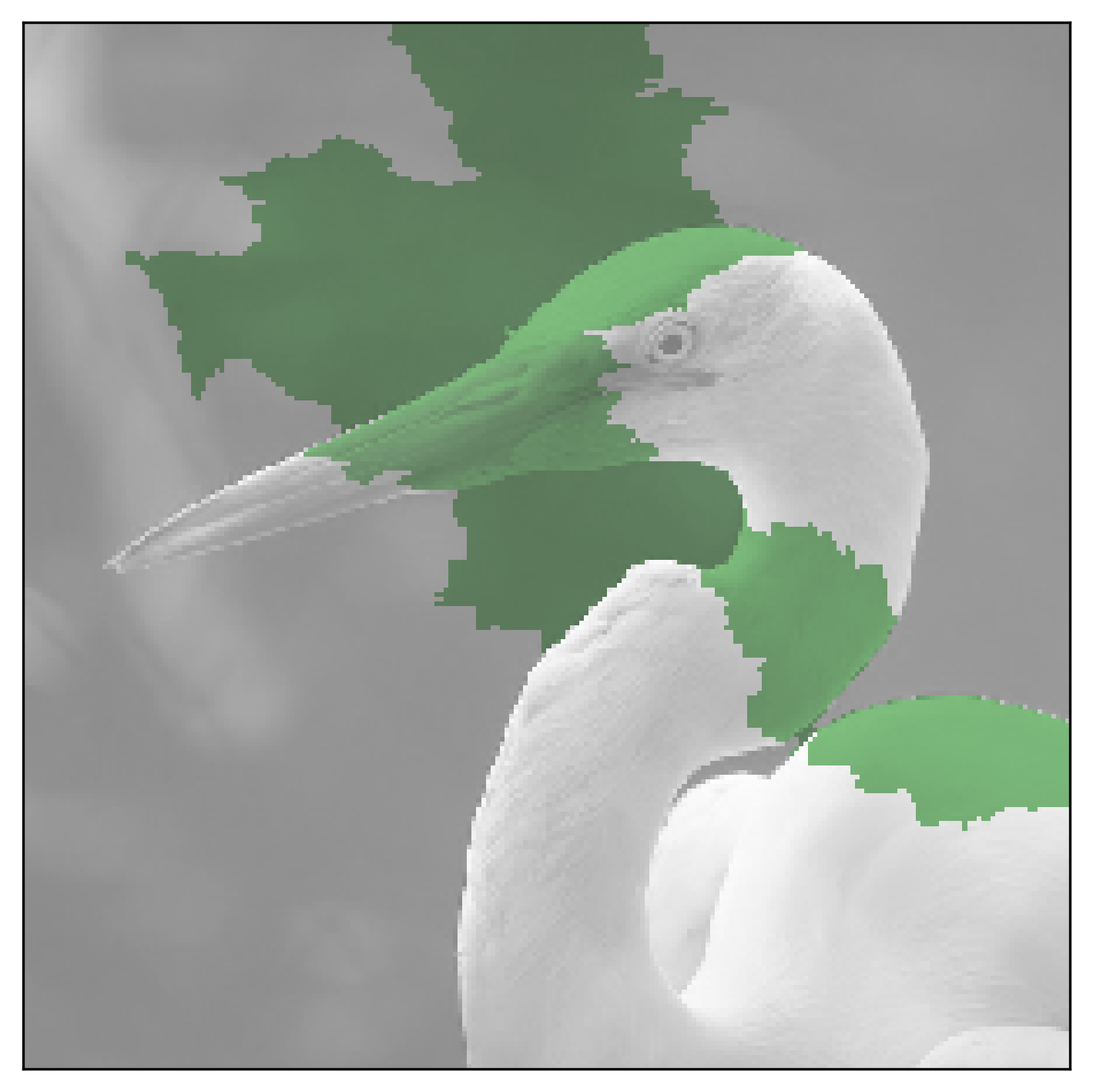} & 
\includegraphics[width=23.9mm]{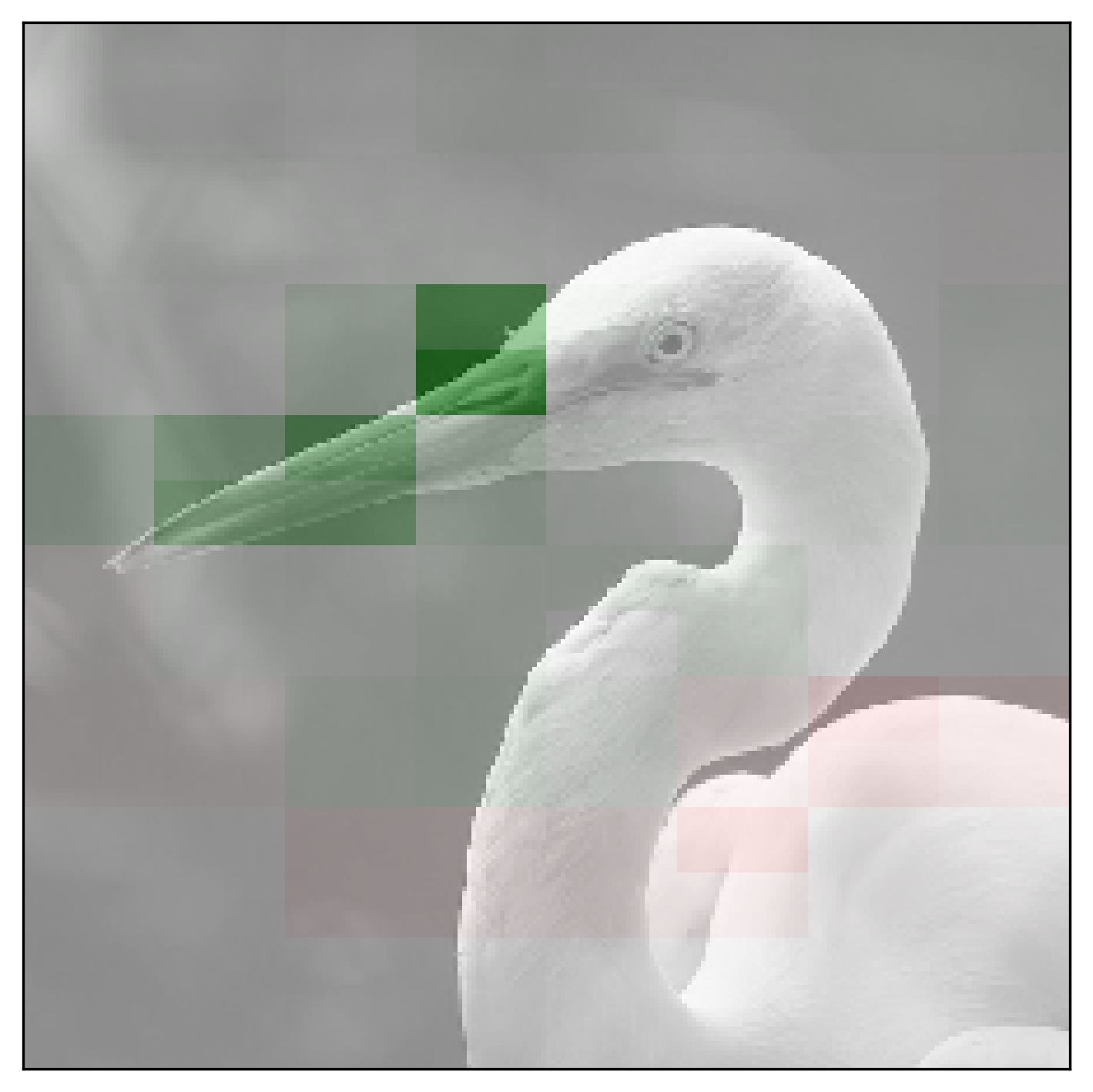} & 
\includegraphics[width=23.9mm]{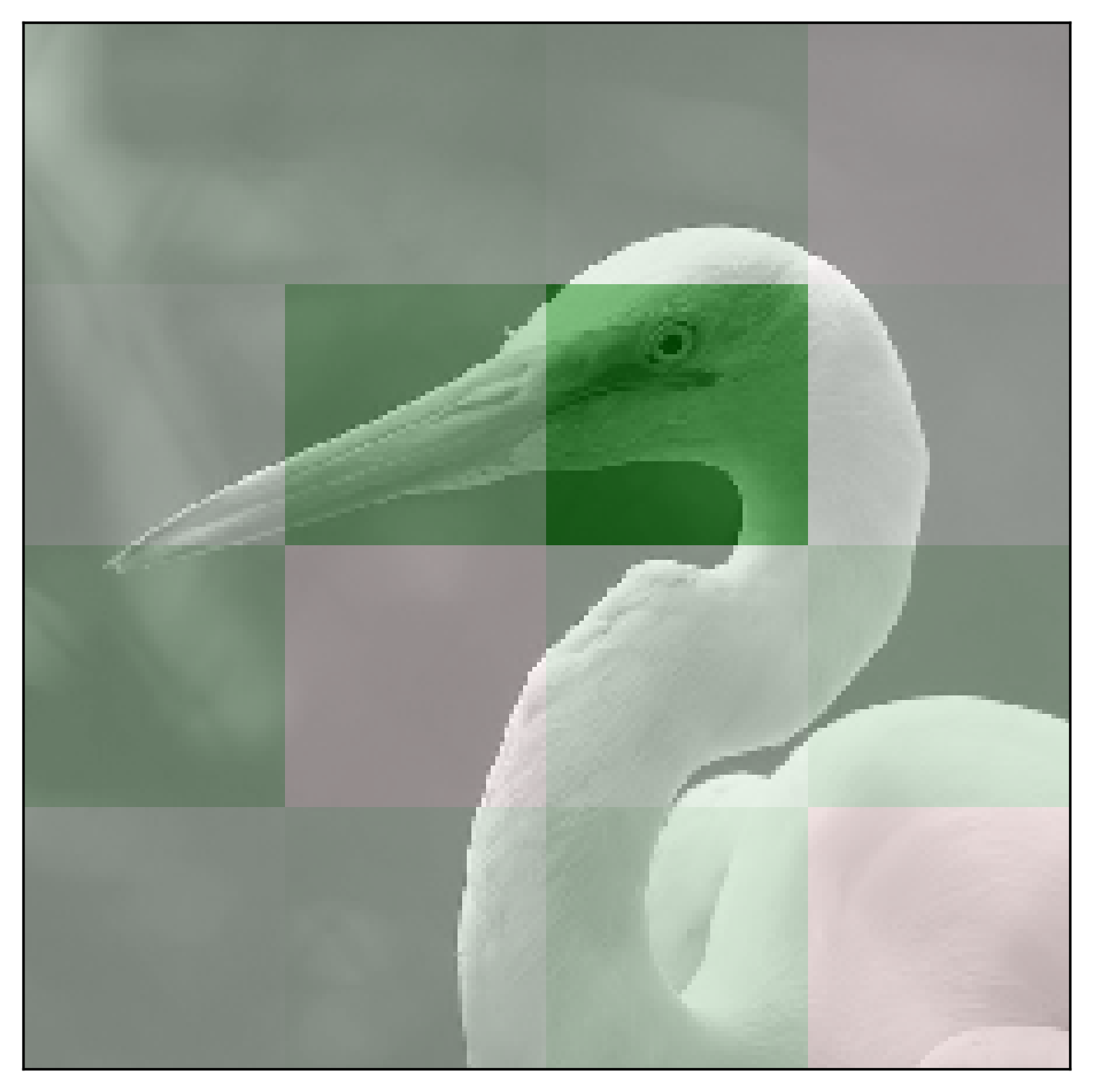} 
\\

\includegraphics[width=23.9mm]{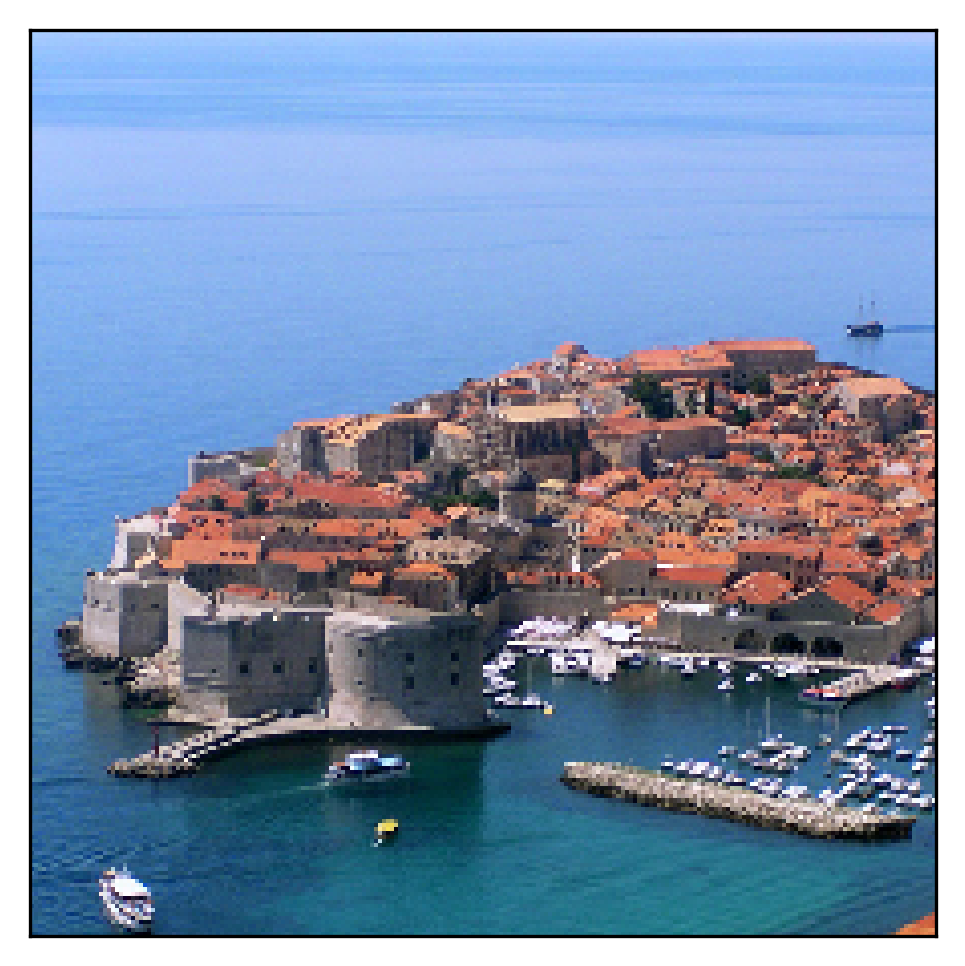} & 
\includegraphics[width=23.9mm]{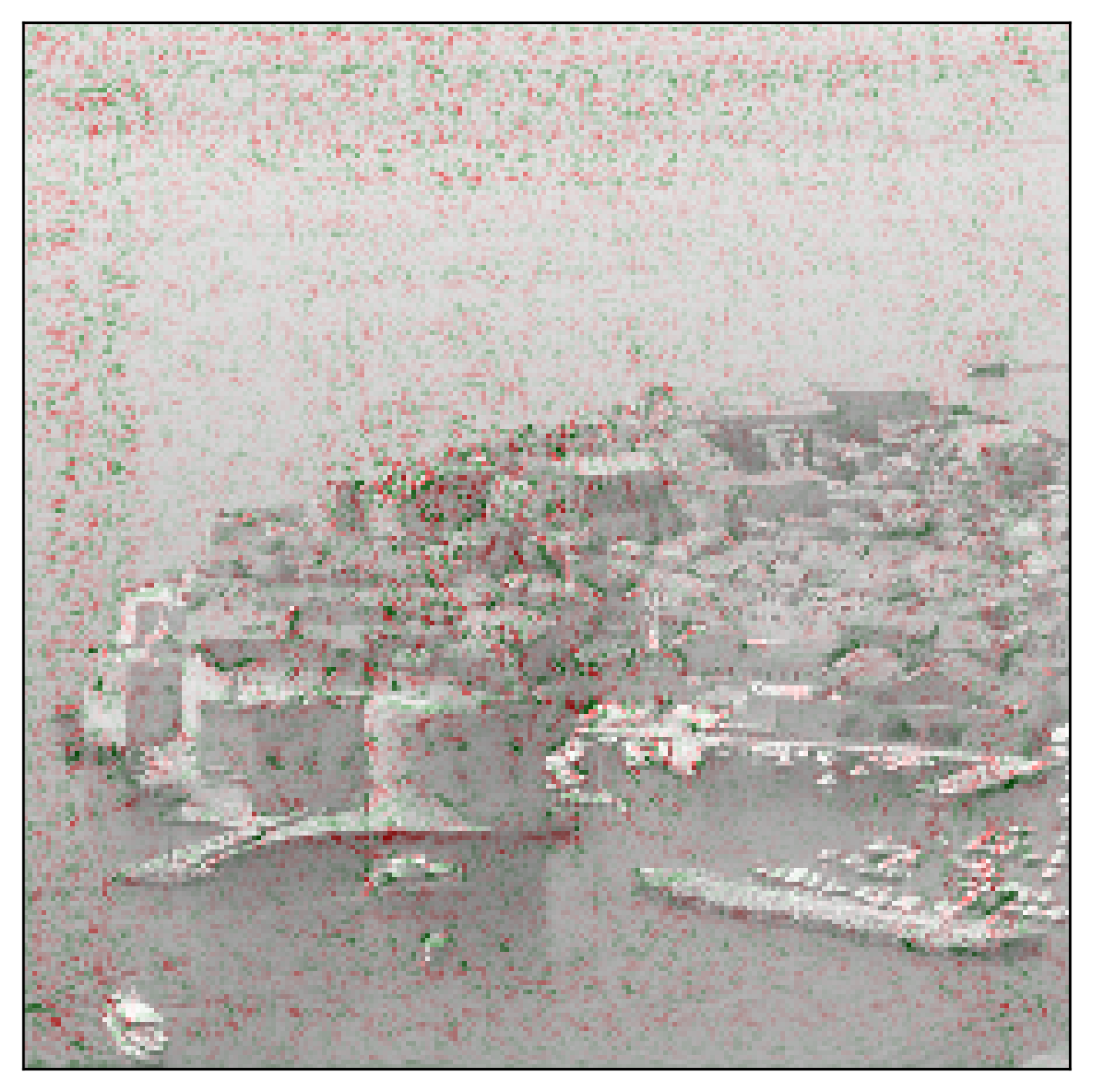} & 
\includegraphics[width=23.9mm]{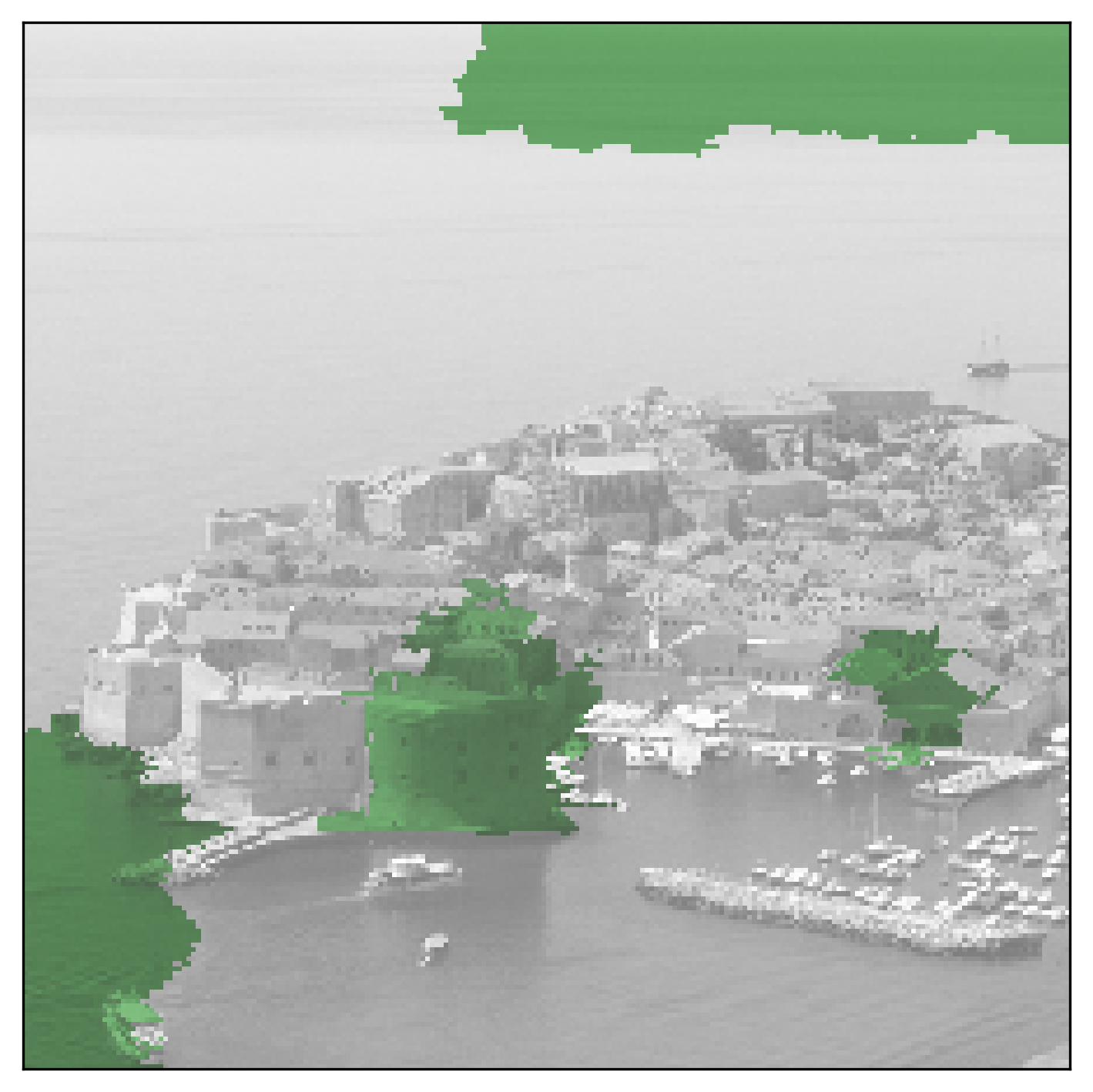} & 
\includegraphics[width=23.9mm]{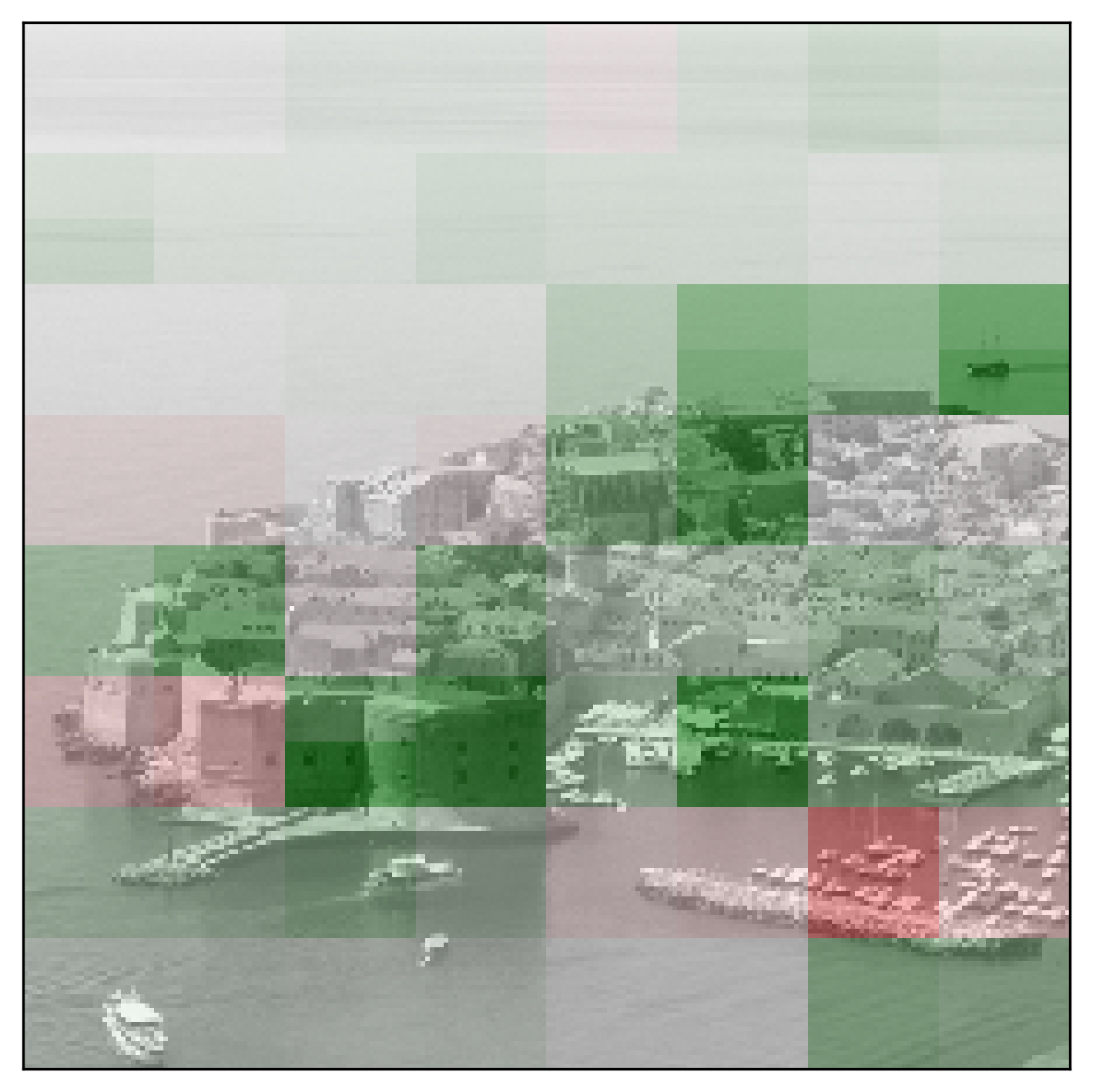} & 
\includegraphics[width=23.9mm]{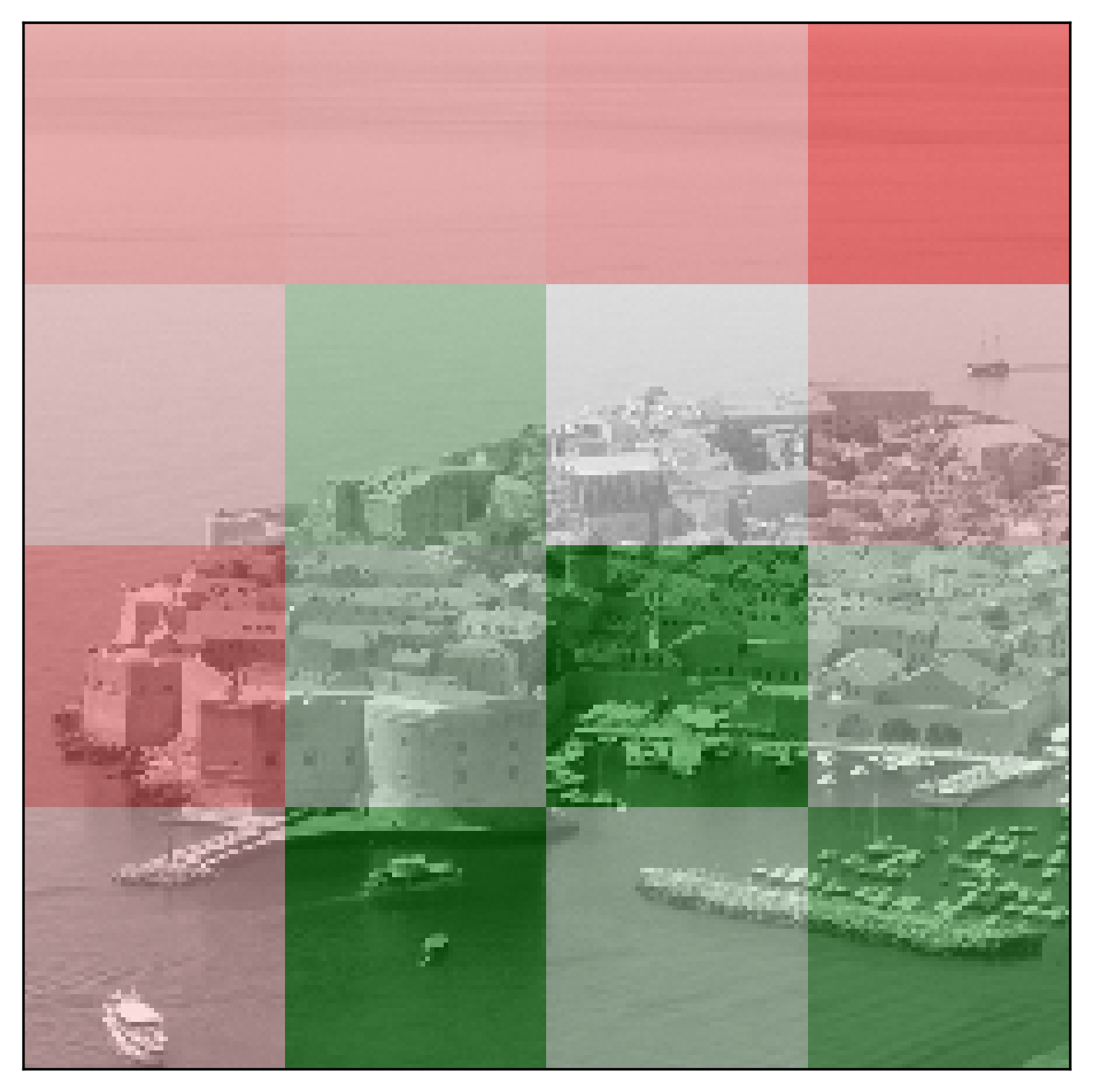} 
\\

\\

\includegraphics[width=23.9mm]{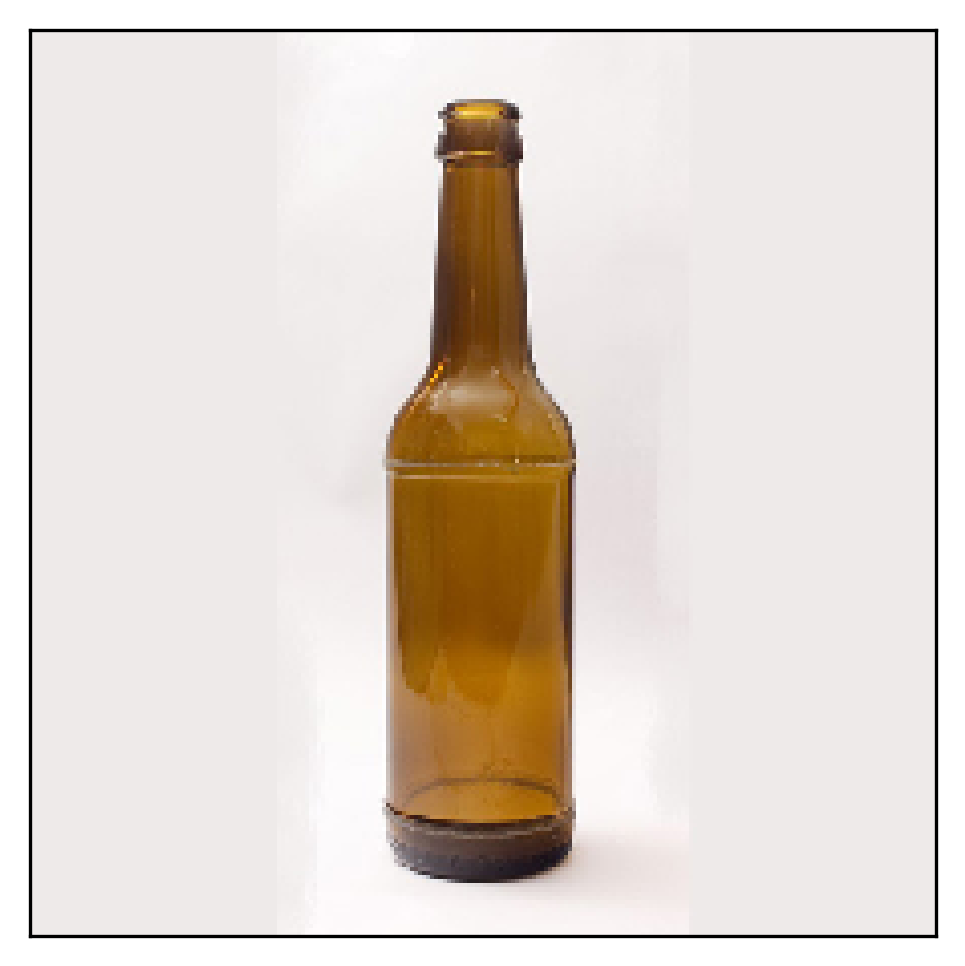} & 
\includegraphics[width=23.9mm]{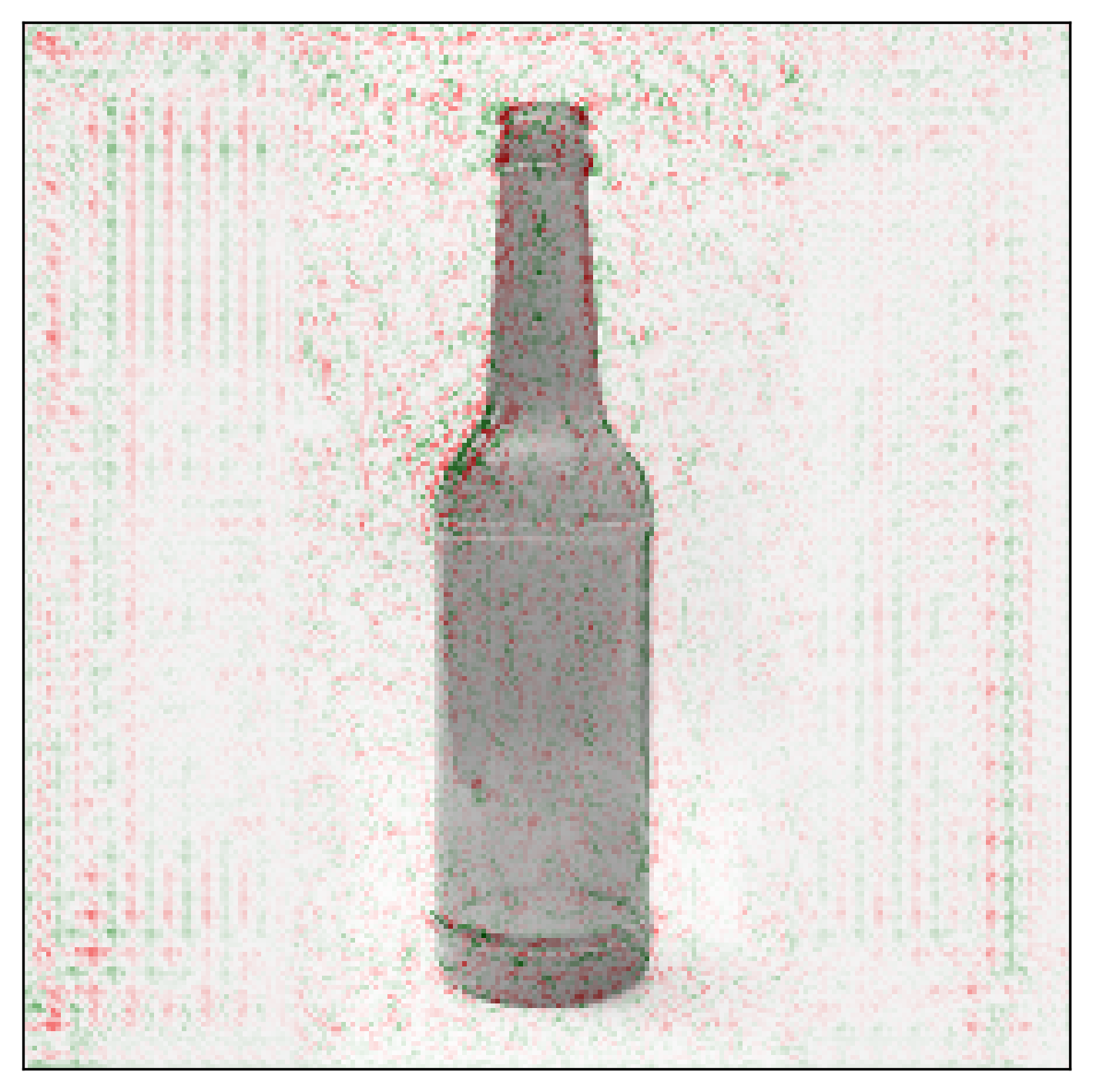} & 
\includegraphics[width=23.9mm]{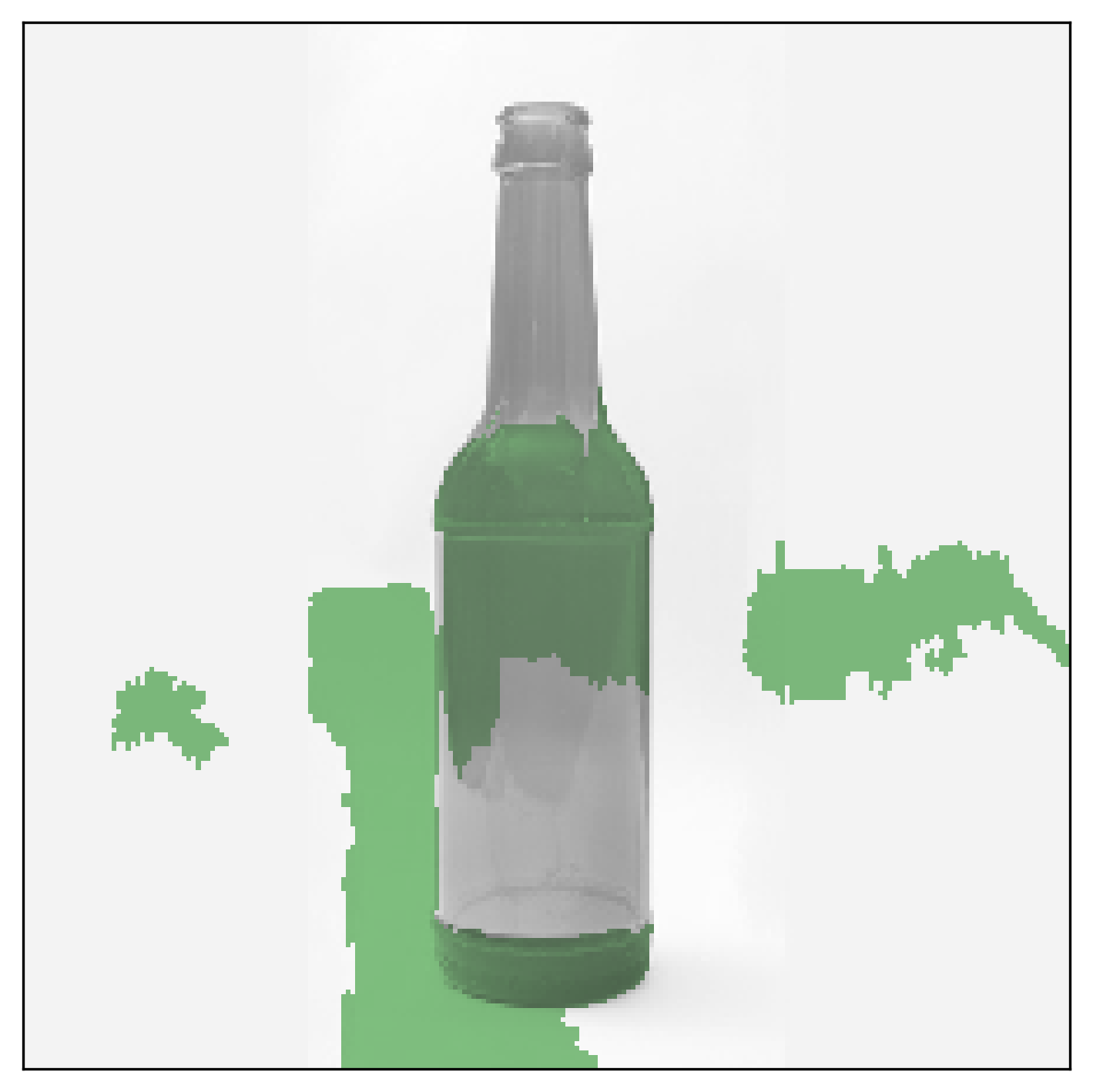} & 
\includegraphics[width=23.9mm]{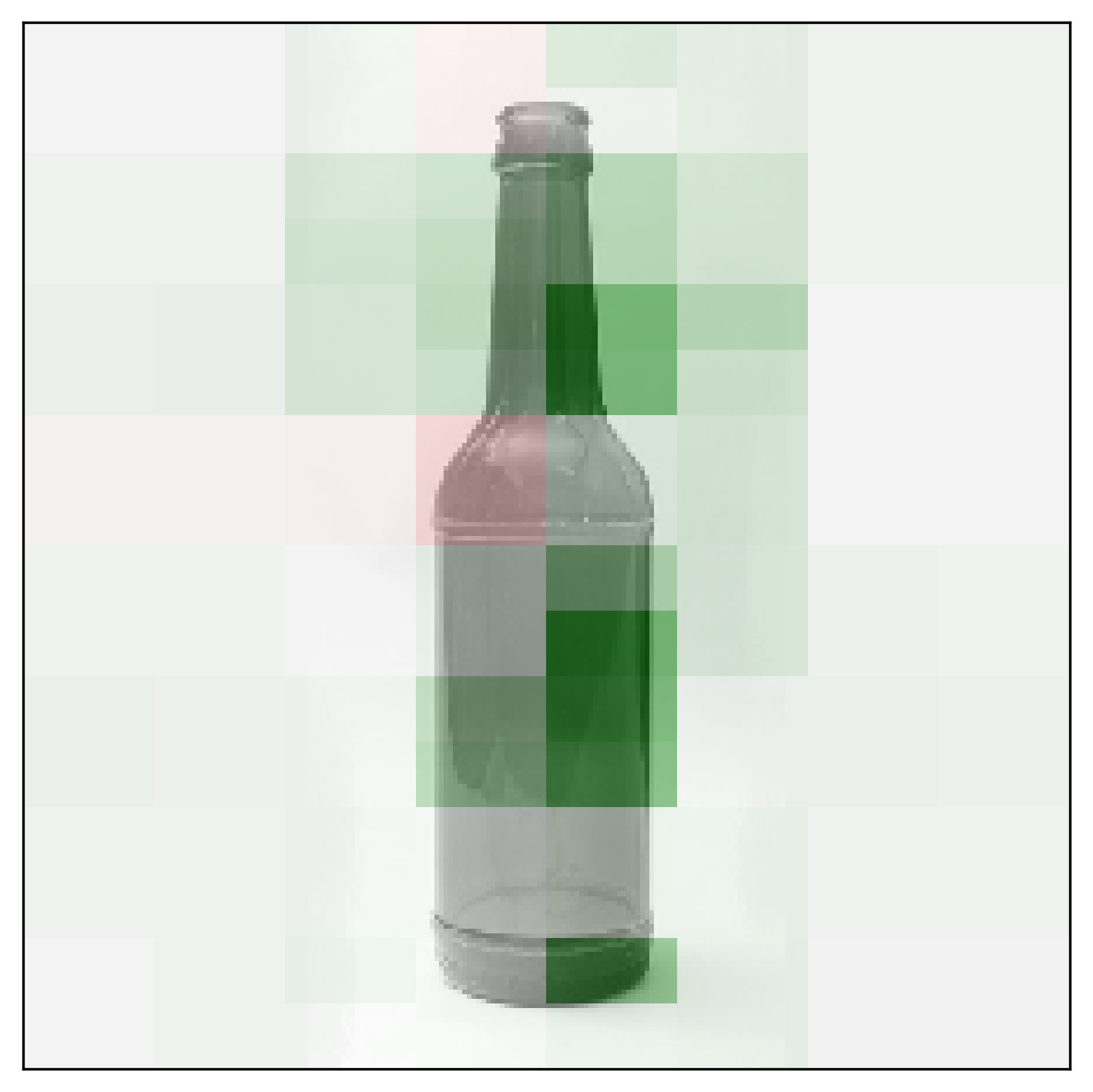} & 
\includegraphics[width=23.9mm]{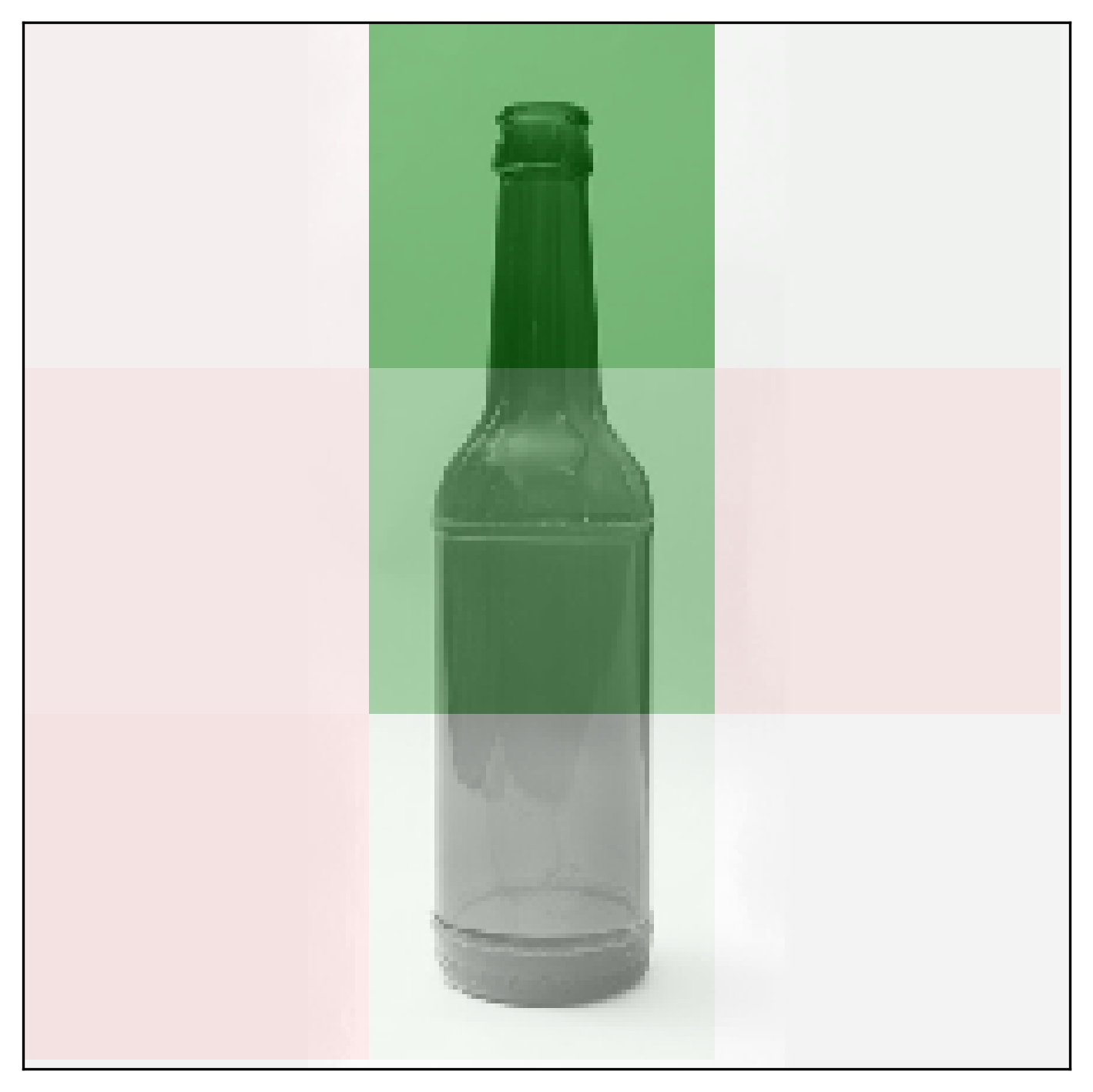} 
\\

\includegraphics[width=23.9mm]{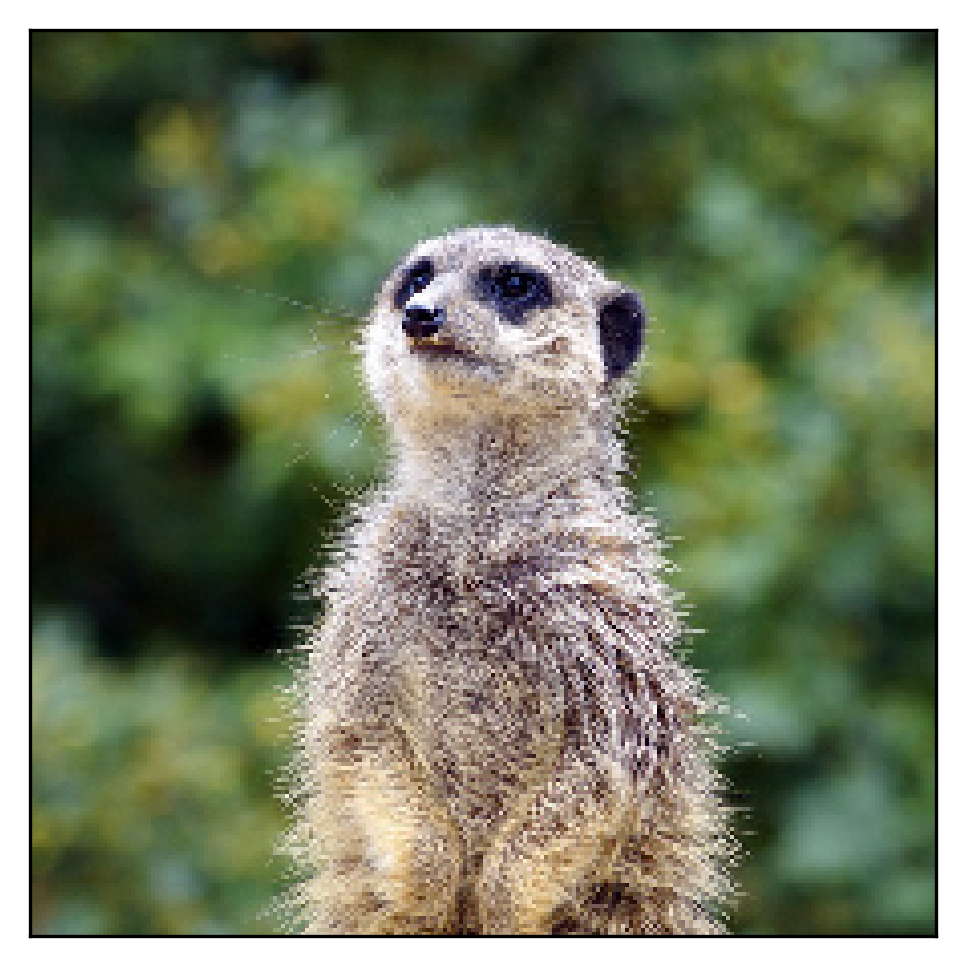} & 
\includegraphics[width=23.9mm]{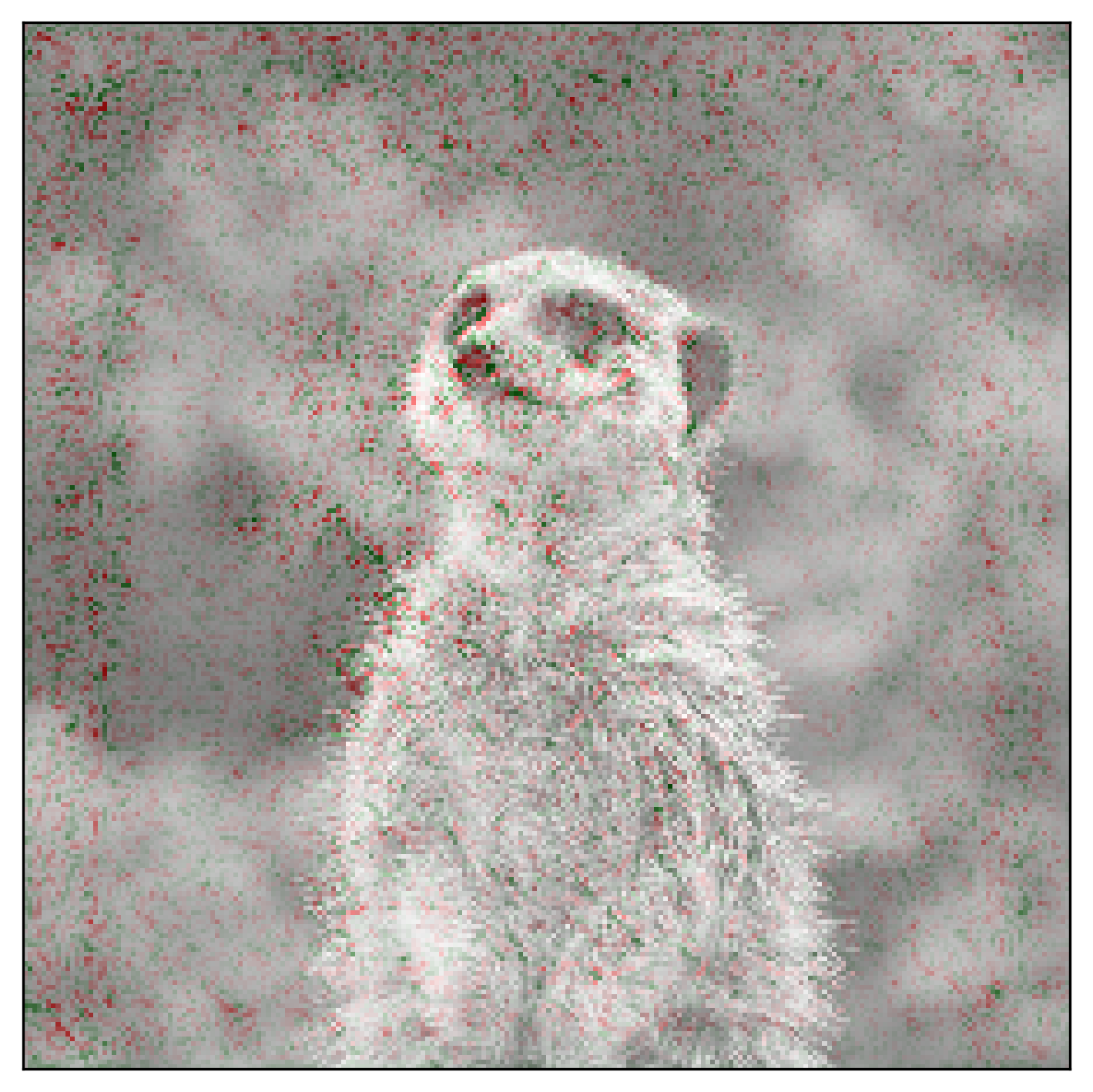} & 
\includegraphics[width=23.9mm]{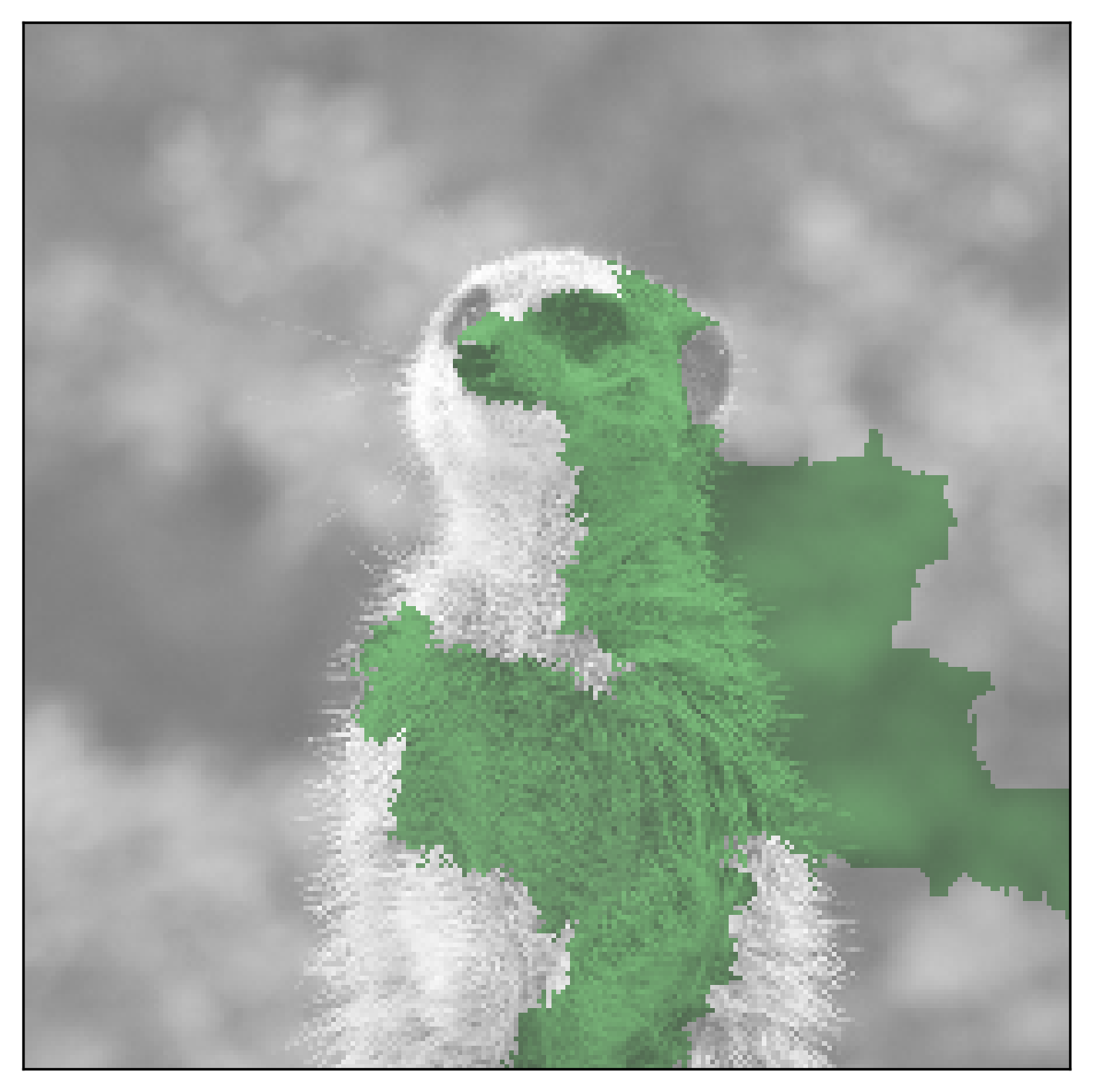} & 
\includegraphics[width=23.9mm]{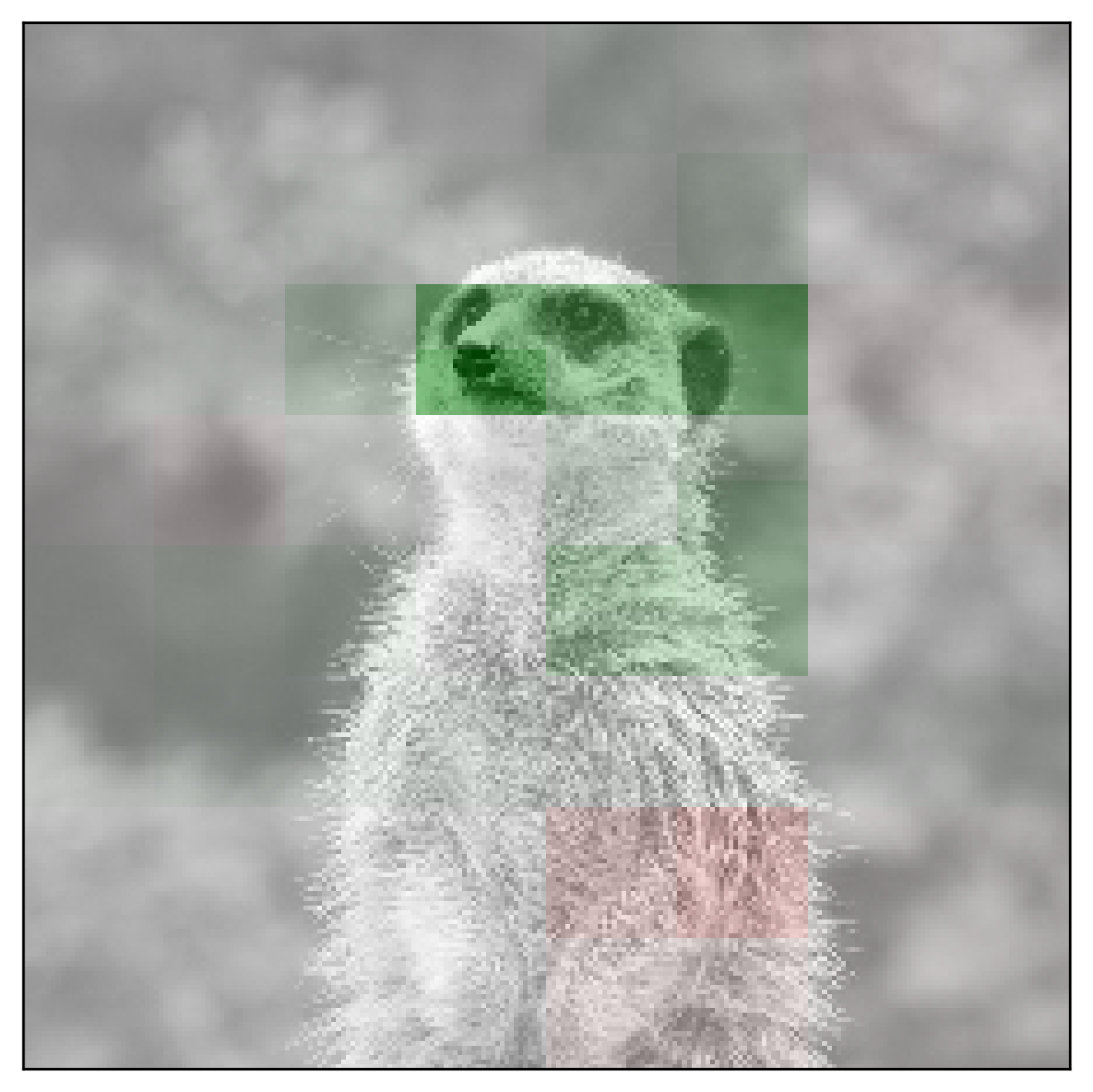} & 
\includegraphics[width=23.9mm]{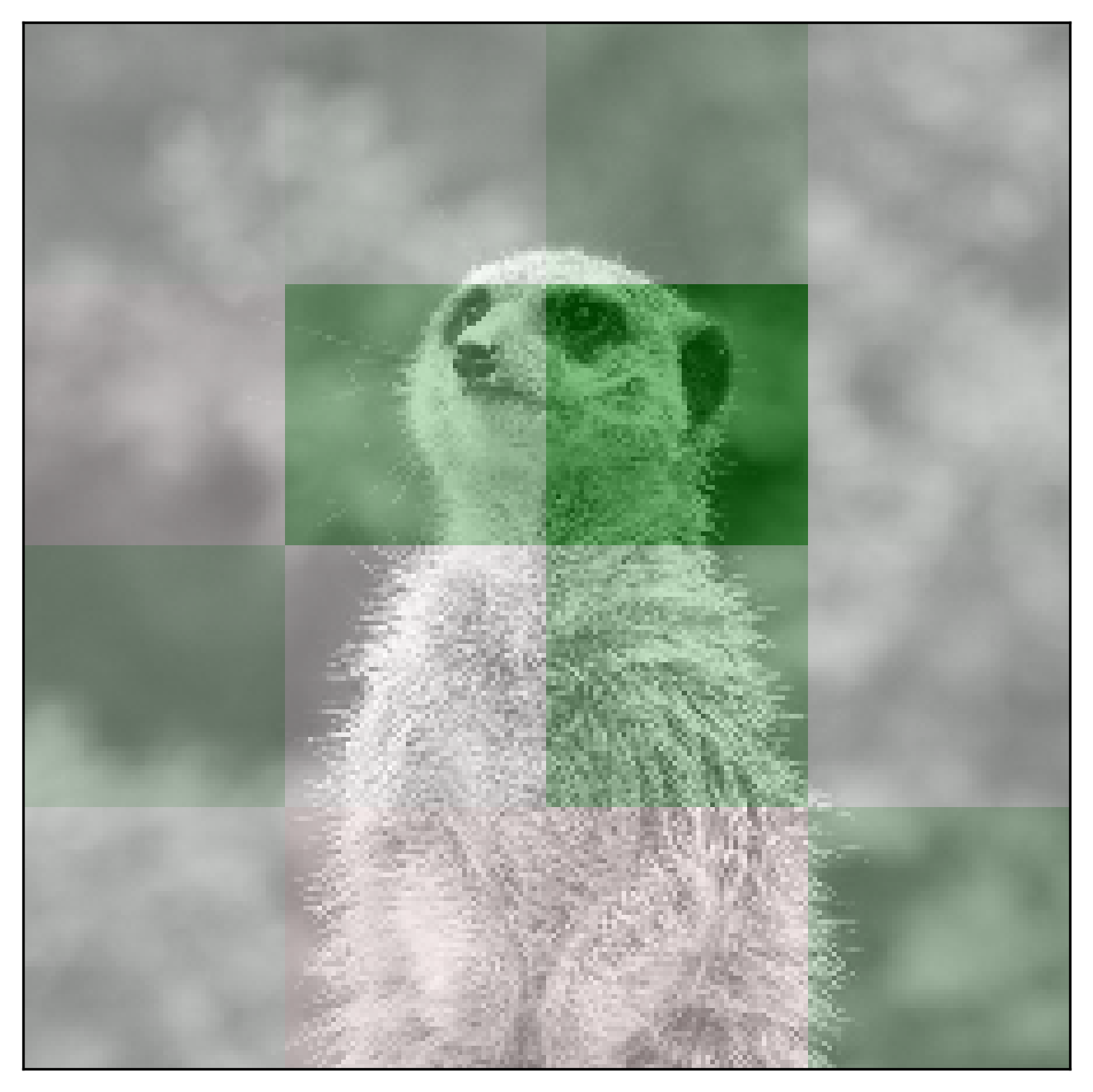} 
\\

\includegraphics[width=23.9mm]{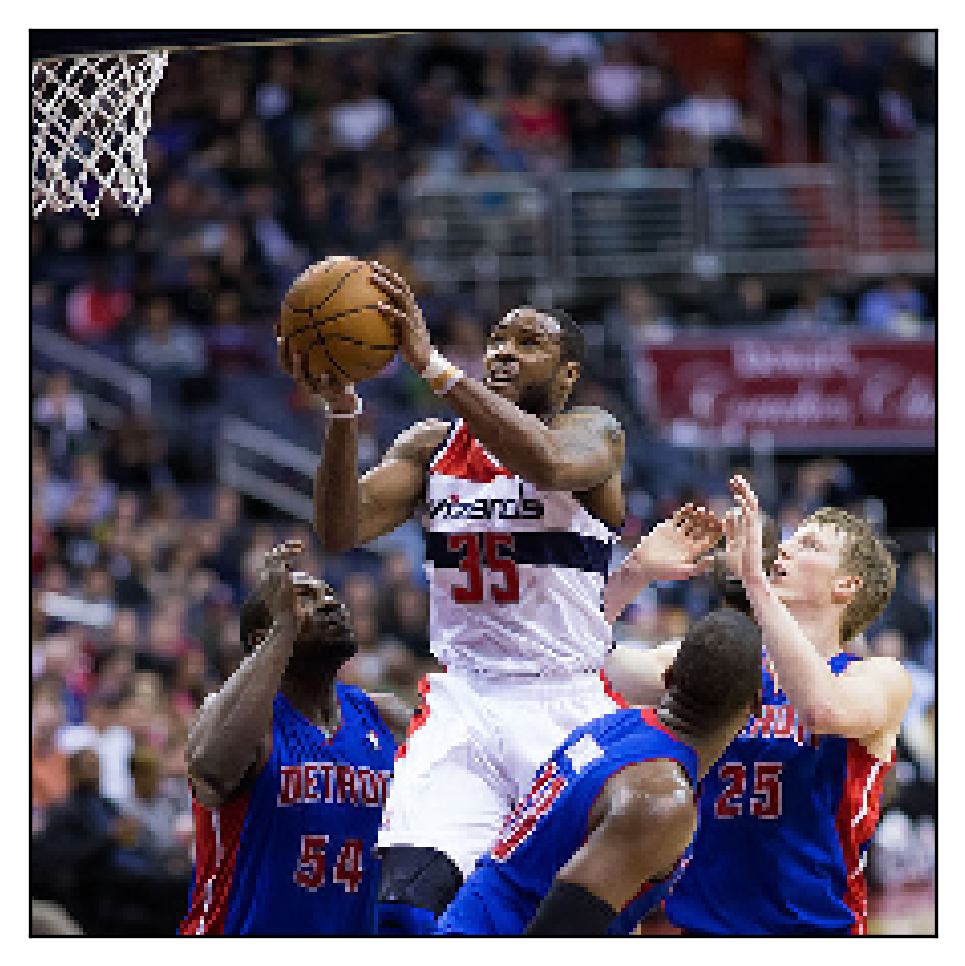} & 
\includegraphics[width=23.9mm]{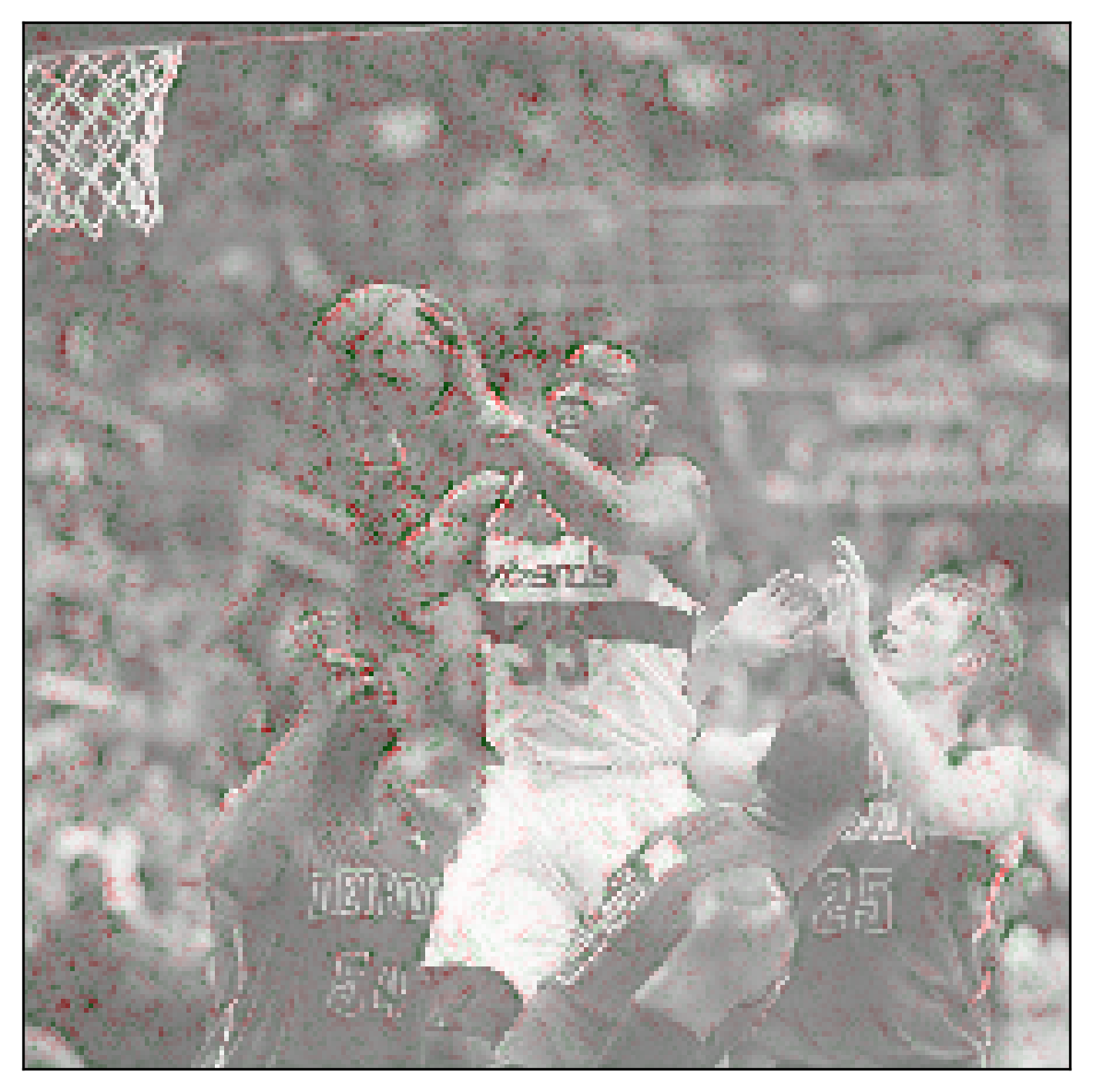} & 
\includegraphics[width=23.9mm]{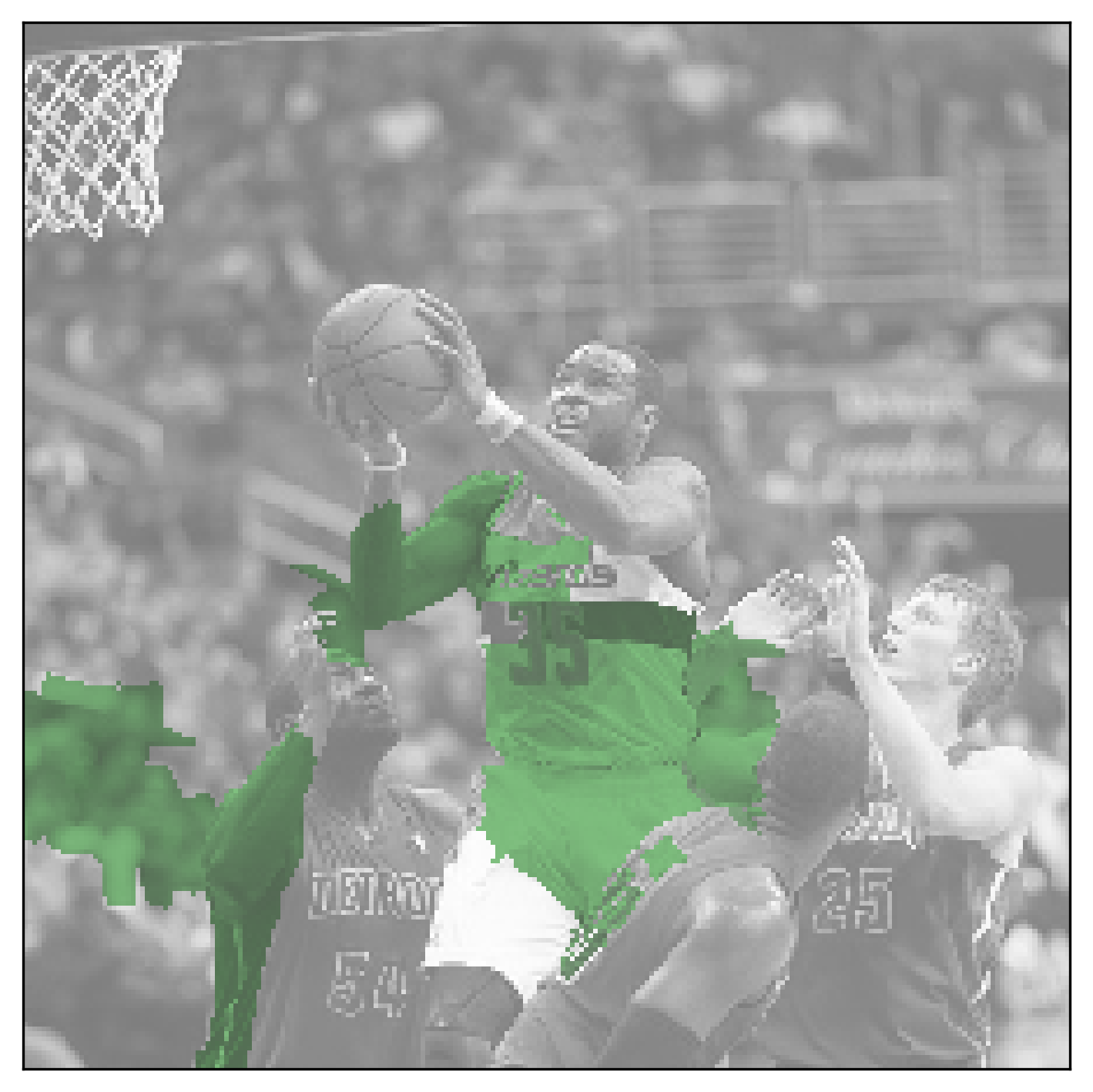} & 
\includegraphics[width=23.9mm]{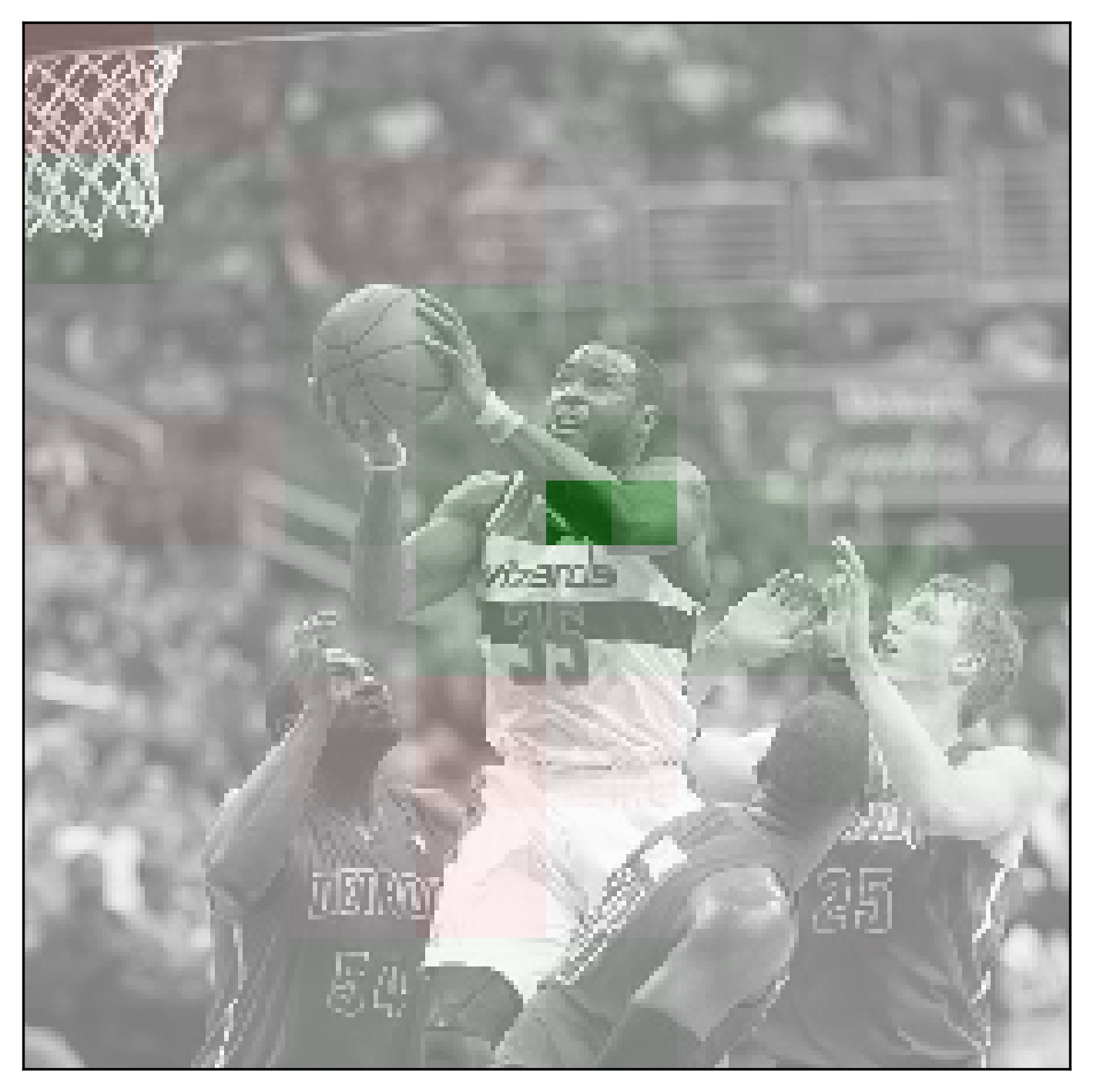} & 
\includegraphics[width=23.9mm]{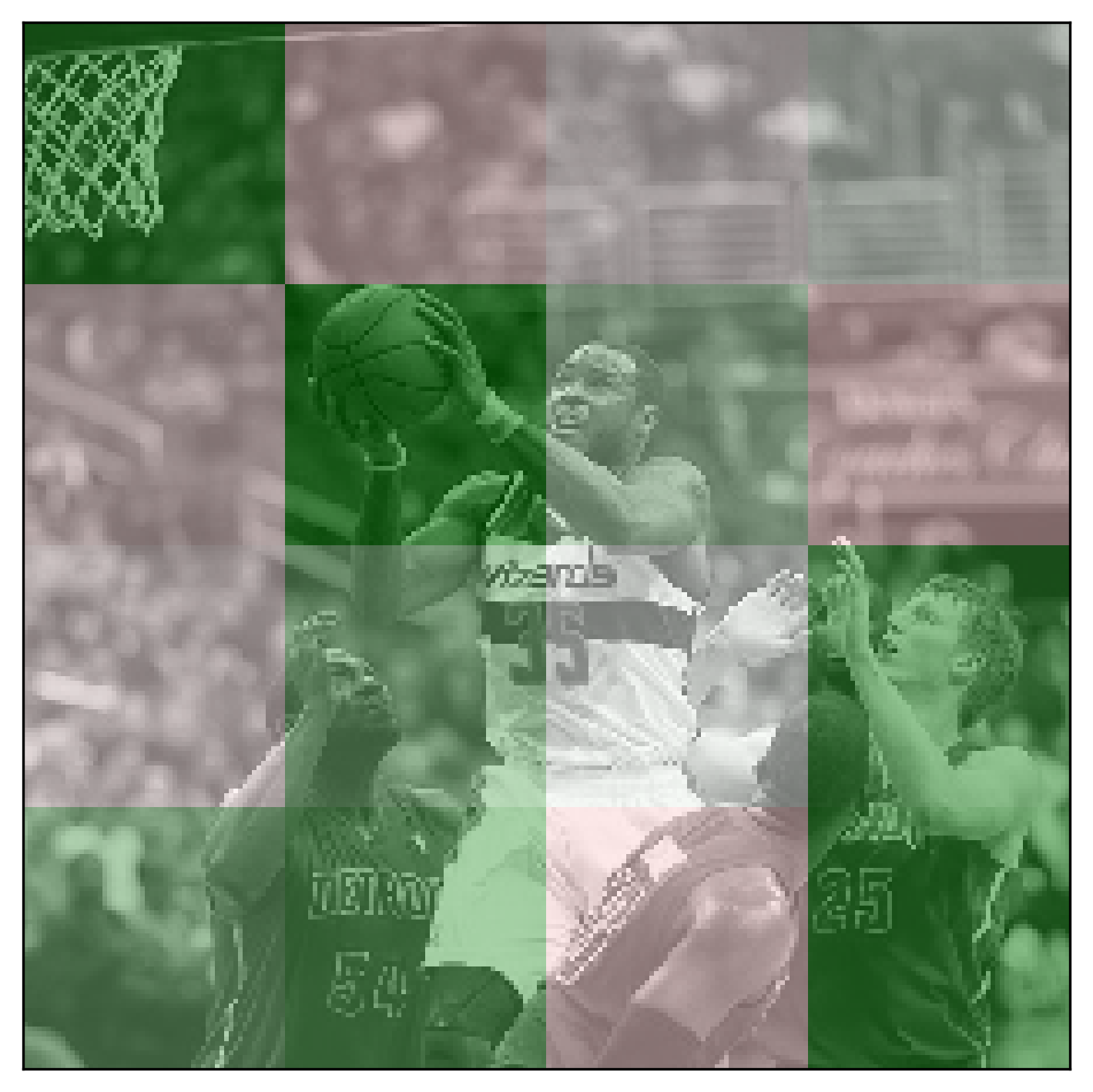} 
\\

\end{tabularx}
\caption{A Comparison of different state-of-the-art local explanation methods for a pre-trained ResNet-50 model~\cite{he2016deep} given random samples from the ImageNet dataset~\cite{deng2009imagenet}. Name of the original classes according to the selected model (from top to bottom): \texttt{American egret}, \texttt{seashore}, \texttt{bottle}, \texttt{marmot}, and \texttt{basketball}. 
}
\label{tab:visual_experiments}
\end{table*}

\section{Experiments}
\label{sec:experiments}

We evaluate the RLE framework on several datasets by visually comparing them with state-of-the-art explanation methods with the goal of ensuring that our method fulfills its purpose. First, in Subsection \ref{sec:exp_visual_data}, we present a qualitative visual comparison of state-of-the-art methods and the proposed framework. Along the same lines, in Subsection \ref{sec:textual_data}, we  compare the local explanations for textual data given a pre-trained DistilBERT model \cite{sanh2019distilbert} and highlight the qualitative superiority of relational local explanations in comparison to state-of-the-art techniques.
Next, in Subsection \ref{sec:empirical}, we presents a quantitative benchmark analysis of the RLE method with a comparison to selected baselines. For more visual experiments and detailed reproducibility details, please refer to the supplemental materials.

\subsection{Visual analysis on image data}
\label{sec:exp_visual_data}

In our first experiment, a qualitative visual evaluation on images from the ImageNet dataset~\cite{deng2009imagenet} is performed for selected baselines: IG~\cite{sundararajan2017axiomatic}, LIME~\cite{ribeiro2016should}, SHAP~\cite{SHAP}, and the proposed RLE framework.  We highlight the explanations from all baselines by appropriate colors to illustrate the attribution they assign to the input variables. A positive attribution value (in green) and a negative attribution value (in red) reflect the contributions of the input values to the CNN model predictions for a given class. The results are summarized in Table \ref{tab:visual_experiments}. Also, an example of a relational local explanation for an image is depicted in Figure \ref{fig:example_image}.

\subsection{Comparison on textual data}
\label{sec:textual_data}

\begin{figure}[t]
  \centering
  \includegraphics[page=1,width=.5\textwidth]{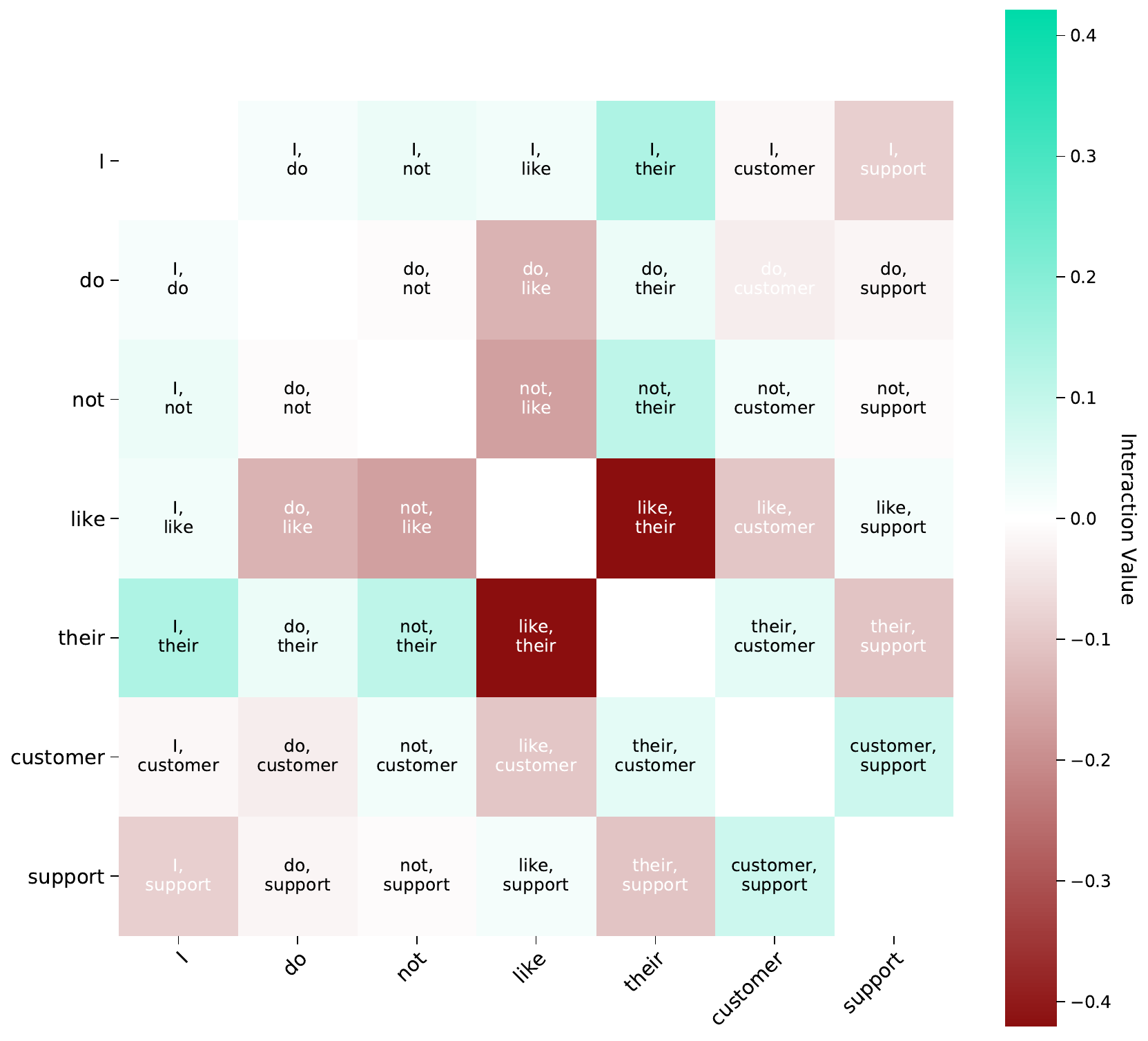}\hspace*{.02\textwidth}%
  \includegraphics[page=1,width=.5\textwidth]{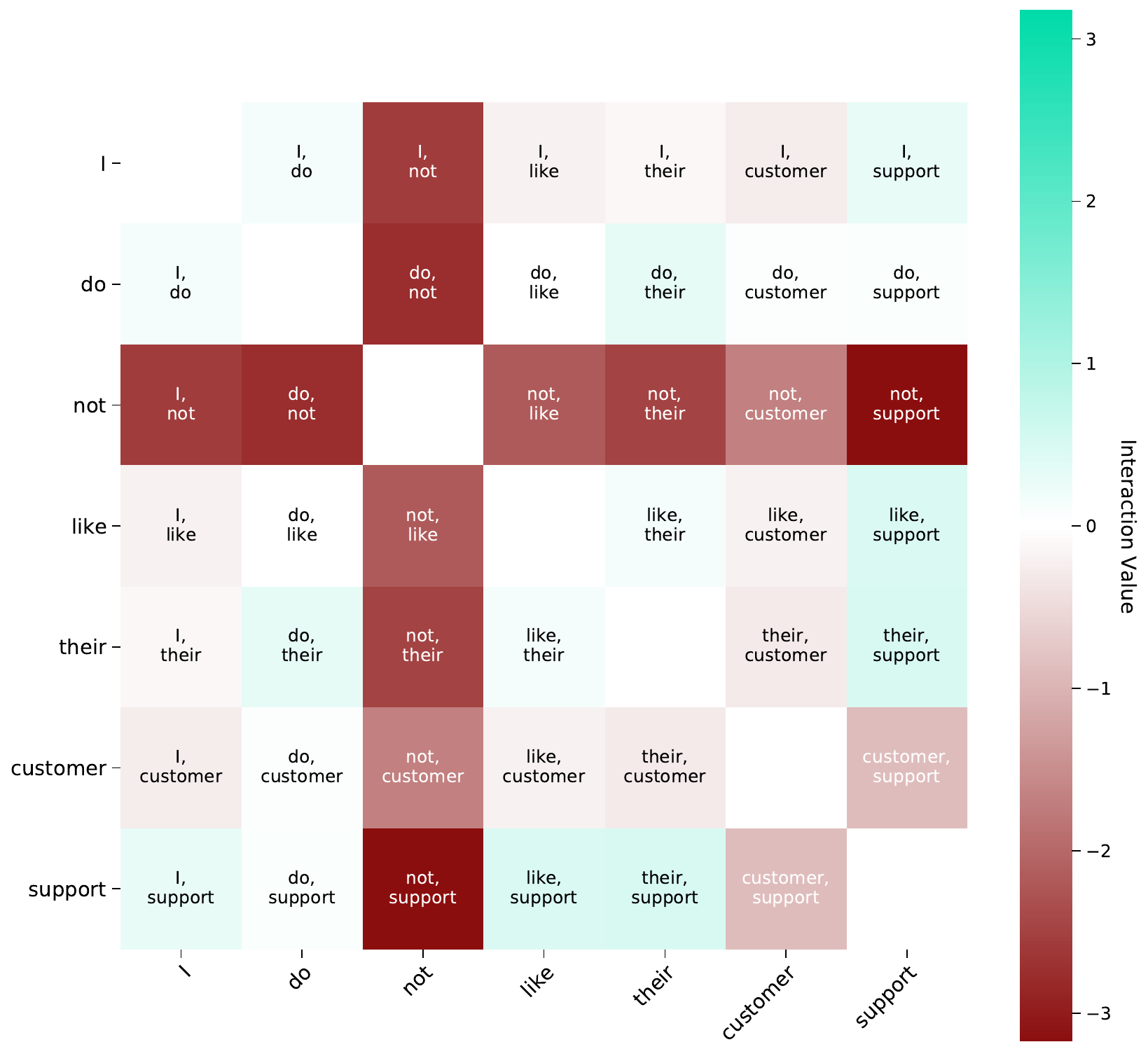}
  \text{ IH \cite{janizek2021explaining} \hspace*{.46\textwidth} RLE (ours)}
  \caption{A comparison of the relational local explanations from IH \cite{janizek2021explaining} and RLE for teh sentence ``\texttt{I do not like their customer support}’’, given a pre-trained \texttt{DistilBERT} model \cite{sanh2019distilbert} for a sentiment analysis task. According to the IH method the most negative pair is ``\texttt{like, their}’’, where our proposed approach shows the most negative word is ``\text{not}’’ with two pairs ``\texttt{not, support}’’ and ``\texttt{do, not}’’. Appendix \ref{sec:app_experiments} presents further results. }
  \label{fig:rle_ih}
  \vspace{-0.5cm}
\end{figure}

For the evaluation of the proposed method on textual data, we utilize a pre-trained self-attention-based DistilBERT model for the sentiment analysis task \cite{sanh2019distilbert} from the open-source Hugging Face library \cite{wolf2020transformers}. The results are summarized in the Table \ref{tab:text_bench1}. Also, we conduct a relational local explanations analysis using the RLE framework (Figure \ref{fig:example_text}) for the same transformer model and compare it to results from the IH method in Figure \ref{fig:rle_ih}.

\begin{table}[t]
    \centering
    \begin{tabular}{l|c}
    \toprule
    Method  &  Local Explanation  \\
    \midrule
    IG \cite{sundararajan2017axiomatic}  & \texttt{You \hlc[red!50]{gonna} \hlc[red!50]{suffer} but you'll be happy \hlc[green!50]{about} \hlc[green!50]{it}}\\ 
    LIME \cite{ribeiro2016should}     &  \texttt{You \hlc[red!50]{gonna} \hlc[red!50]{suffer} \hlc[red!50]{but} \hlc[green!50]{you'll} \hlc[red!50]{be} \hlc[green!50]{happy} \hlc[red!50]{about it}}\\ 
    SHAP \cite{SHAP}                  &   \texttt{\hlc[green!50]{You} gonna \hlc[red!50]{suffer} but \hlc[green!50]{you'll} be \hlc[green!50]{happy} about it}\\
    IH \cite{janizek2021explaining}  &   \texttt{You gonna \hlc[red!50]{suffer} \hlc[green!50]{but} \hlc[green!50]{you'll} be \hlc[green!50]{happy} \hlc[green!50]{about} it}\\
    \textbf{RLE (ours)}          &  \texttt{You \hlc[red!50]{gonna} \hlc[red!50]{suffer} but \hlc[green!50]{you'll} be \hlc[green!50]{happy} about it} \\
    
    \midrule

    IG \cite{sundararajan2017axiomatic}  & \texttt{You \hlc[green!50]{might} be interested \hlc[red!50]{this} product performs \hlc[green!50]{well}}\\ 
   LIME \cite{ribeiro2016should}  &   \texttt{You \hlc[red!50]{might} be \hlc[green!50]{interested} this product \hlc[green!50]{performs} \hlc[green!50]{well}}\\
    SHAP \cite{SHAP} &  \texttt{\hlc[green!50]{You} \hlc[red!50]{might} be \hlc[red!50]{interested} this product performs \hlc[green!50]{well}}\\ 
    IH \cite{janizek2021explaining} &  \texttt{You might \hlc[red!50]{be} \hlc[red!50]{interested} this \hlc[green!50]{product} \hlc[green!50]{performs} \hlc[green!50]{well}}\\ 

    \textbf{RLE (ours)}          &  \texttt{You \hlc[red!50]{might} \hlc[red!50]{be} interested this product \hlc[green!50]{performs} \hlc[green!50]{well}} \\
    
    \midrule
    
    IG \cite{sundararajan2017axiomatic}  & \texttt{\, The idea \hlc[green!50]{is} \hlc[green!50]{nicely} presented, but \hlc[red!50]{it} has some limitations}\\
    LIME \cite{ribeiro2016should}  & \texttt{\, The \hlc[red!50]{idea} is \hlc[green!50]{nicely} presented, \hlc[red!50]{but} it has some \hlc[red!50]{limitations}}\\
    SHAP \cite{SHAP}  & \texttt{\, \hlc[green!50]{The} idea is \hlc[green!50]{nicely} presented, \hlc[red!50]{but} it has \hlc[green!50]{some} limitations}\\
    IH \cite{janizek2021explaining}  & \texttt{\, \hlc[red!50]{The} idea is \hlc[green!50]{nicely} \hlc[green!50]{presented}, but it \hlc[red!50]{has} \hlc[red!50]{some} limitations}\\
    
    \textbf{RLE (ours)}          &  \texttt{\, The \hlc[red!50]{idea} is \hlc[green!50]{nicely} \hlc[green!50]{presented}, but it has some \hlc[red!50]{limitations}} \\
    
    \bottomrule

    \end{tabular}
    \caption{A comparison of state-of-art feature attribution approaches to the presented RLE algorithm. given a pre-trained \texttt{DistilBERT} model \cite{sanh2019distilbert} for the sentiment analysis task. We highlight the most important words according to each feature attribution method. For more results, please refer to the supplementary materials.}
    \label{tab:text_bench1}
    \vspace{-6pt}
\end{table}

\subsection{Quantitative comparison}
\label{sec:empirical}


\begin{wraptable}{r}{5.5cm}
{\small
    \centering
    \begin{tabular}{l|c}
    \toprule
    Method  &  IROF~\cite{rieger2020irof}  \\
    \midrule
    Random &  0.179$\pm$0.18\\ 
    IG \cite{sundararajan2017axiomatic} &  0.211$\pm$0.23\\ 
    LIME \cite{ribeiro2016should}     &   \underline{0.421$\pm$0.19}\\ 
    SHAP \cite{SHAP}                  &   0.368$\pm$0.24 \\
    \textbf{RLE (ours)}          & \textbf{0.434$\pm$0.23} \\
    
    \bottomrule

    \end{tabular}
}
    \caption{Quantitative comparison with the baselines on fifty random images from ImageNet \cite{deng2009imagenet} given a pre-trained ResNet-50 model \cite{sanh2019distilbert}. The top result is bold, the second-best is underlined.}
    \label{tab:irof_main}
    \vspace{-19pt}
\end{wraptable}

In order to quantitatively evaluate our novel explanation framework, we utilize the well-accepted measure in the ML community - Iterative Removal Of Features (IROF)~\cite{rieger2020irof}. The IROF measure is fully described in the supplementary materials. The full definition of the mesure is in the Appendix \ref{app:irof}. 
This technique was featured in multiple studies before \cite{bhatt2020evaluating}. We compare against the baselines selected for this study:  IG \cite{sundararajan2017axiomatic}, LIME \cite{ribeiro2016should}, and SHAP \cite{SHAP}. We also introduce the random baseline as a ``sanity check’’, which assigns variable importance randomly. Notably, the authors of~\cite{hooker2018evaluating, rong2022consistent} show that this primitive baseline can sometimes outperform some of the commonly used explanation approaches based on saliency maps in ablation tests. Results are shown in Table \ref{tab:irof_main}.


\subsection{Reproducibility Details}
\label{sec:repro}

In this subsection, we briefly introduce the main frameworks used in this study; further details about all experiments, such as hyperparameters for each baseline and experiment, are provided in the supplementary materials. We selected official implementation for the LIME, SHAP, and IH baselines, and for the IG baseline, we employed the Captum library \cite{kokhlikyan2020captum}. The graph structure was analyzed using the NetworkX library \cite{hagberg2008exploring}. For all experiments we use a single NVIDIA 2080TI GPU with 12 GB of memory. For future comparison, we also open-source the code for the RLE framework for PyTorch \cite{NEURIPS2019_9015} models and publish it online.

\section{Discussion and Future Work}
\label{sec:discussion}

\textbf{Experimental results.} In challenging experiments with multiple data modalities, local explanations obtained by the RLE framework show competitive performance against selected feature attribution baselines. Overall, our quantitative experimental results highlight similar and yet succinctly more relevant image areas or words as other state-of-the-art non relational explanation methods. In quantitative experiments, the proposed approach shows the best results on the IROF measure \cite{rieger2020irof}. For more results, please refer to the supplementary material.

\textbf{Permutation step.} One of the core steps of the proposed algorithm is the random permutation with replacement - we refer to it as \textit{weak perturbation}. In comparison to other perturbation-based feature attribution methods which use \textit{strong perturbation}, e.g., perturb a data sample by adding random noise \cite{ribeiro2016should, sundararajan2017axiomatic} or removing parts of information \cite{SHAP}, the RLE framework preserves the local structure unchanged, only shuffling the global structure, which is based on the insight that local details are more important than a global structure for deep neural network models, as shown in various vision and text-related tasks \cite{chen2019destruction, yang2019xlnet, huang2021shuffle}. Another known issue related to strong-perturbation approaches for local approximations is that this type of perturbation leads to the out-of-the-distribution problem \cite{hase2021out}, which creates the vulnerability to adversarial attacks \cite{slack2020fooling}.

\textbf{Evaluation measure for relational local explanations.} Another future research aspect about relational local explanations is the absence of a transparent evaluation technique. For future work, a trustworthy and plausibility measure is needed; this is challenging since there is no access to a ground truth. On the flip side, with an unambiguous measure, a possible strategy would involve direct optimization over it and possibly lead to much simpler techniques than the one presented in this paper.

\textbf{Ensembling of feature attributions.} To improve the robustness of local relational explanations, unsupervised ensemble techniques can be applied to the outputs of multiple runs of the explanation algorithm. By aggregating the outputs of multiple runs of the algorithm, we can effectively reduce the impact of any individual run that may have produced biased or inaccurate explanations. This approach has been shown to be effective at improving the robustness of explanations in a variety of contexts \cite{borisov2021robust}. 

\textbf{RLE limitations.}  
The proposed approach does not currently support the quantification of higher-order interactions between features for relational local explanations. A more complex graph-based representation can be utilized for this task in future work. 
Furthermore, patches of image data should have adequate local information, and the adequacy depends on the resolution of the images. In our experiments using the ImageNet dataset - an image is usually cropped into a $224\times224\times3$ format, we observe that we can operate with up to 36-49 patches (depending on the content of an image).   
Lastly, the current implementation of the RLE framework cannot be utilized on tabular data modalities yet. Further research in this direction could focus on better leveraging hidden, and yet non-local, relationships between variables \cite{borisov2021deep}.


\section{Conclusion}
\label{sec:conclusion}
We introduced RLE (Relational Local Explanations), a novel, model-agnostic approach for generating local explanations with the aim to address a common challenge in post-hoc explanations: interpretability based on inter-variable relationships. RLE is capable of quantifying the interaction between different feature attributions, which is useful for various data types and problems where knowledge of the relationships between features is necessary but difficult to obtain. In addition, the RLE framework also offers standard feature attributions as local explanations, which provide insight into the specific contributions of each feature towards the final prediction made by a machine learning model. Through extensive qualitative and quantitative experiments, we demonstrated that the proposed method often outperforms state-of-the-art methods for generating comprehensive (relational) local explanations. These results suggest that RLE may be a valuable tool for practitioners seeking to understand and improve the performance of their machine learning models.


\bibliographystyle{unsrt}
\bibliography{ref}

\appendix

\section{Additional Experiments}
\label{sec:app_experiments}

\begin{figure}[t]
  \centering
  \includegraphics[page=1,width=.5\textwidth]{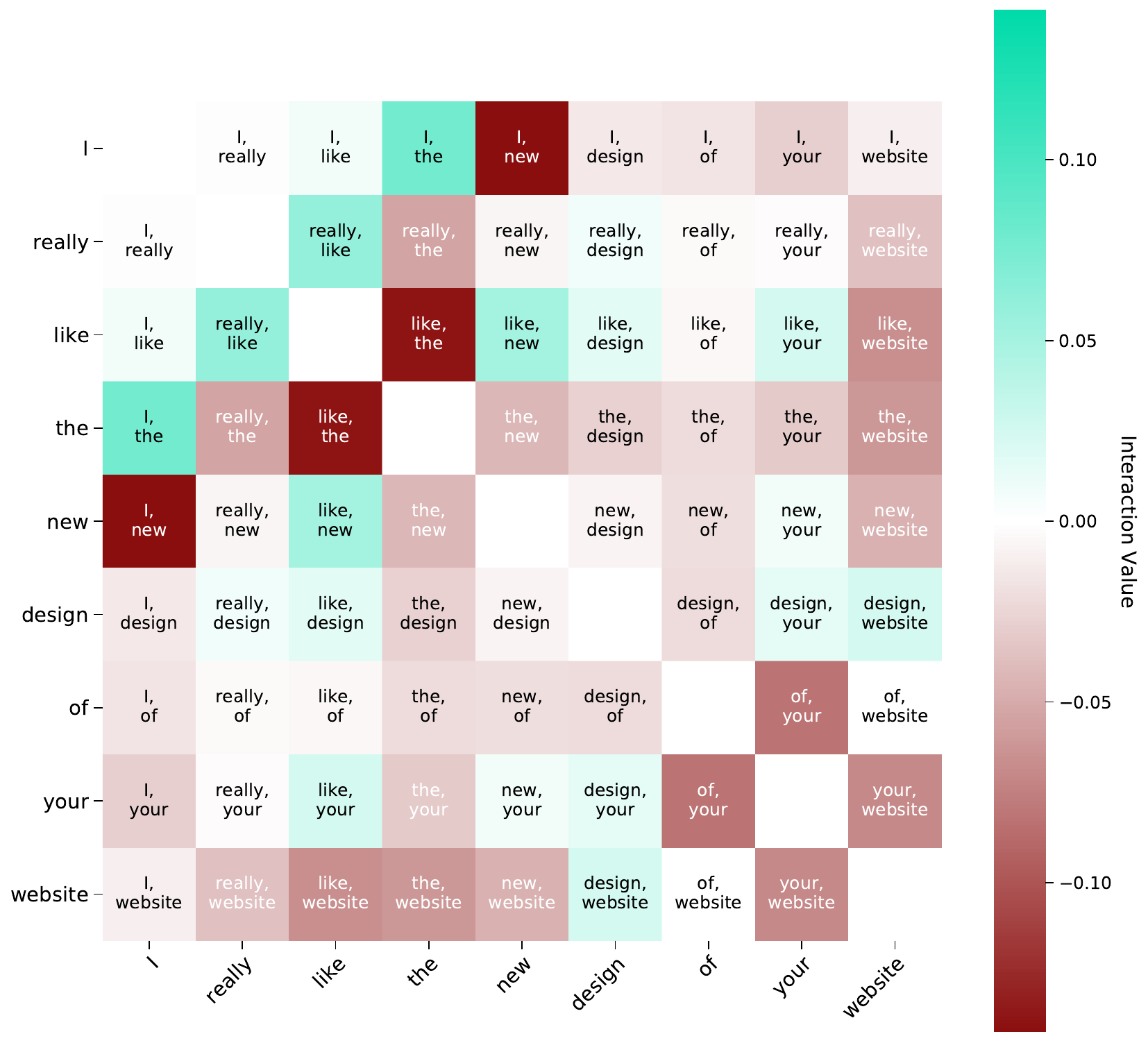}\hspace*{.05\textwidth}%
  \includegraphics[page=1,width=.5\textwidth]{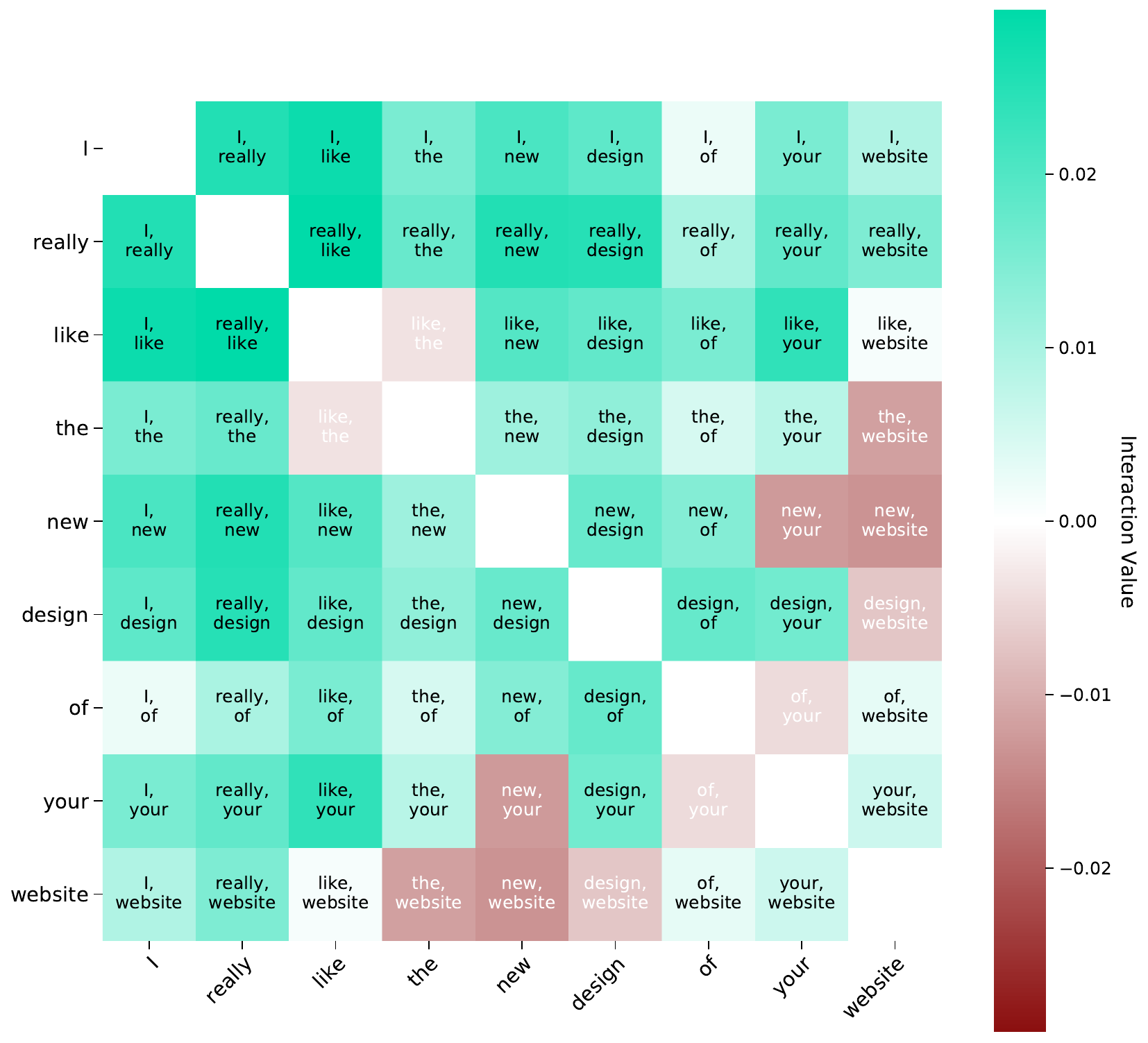}
  \text{ IH \cite{janizek2021explaining} \hspace*{.46\textwidth} RLE (ours)}
  \caption{A comparison of the relational local explanations from IH \cite{janizek2021explaining} and RLE methods for a sentence ``\texttt{I really like the new design of your website}’’, given a pre-trained DistilBERT model \cite{sanh2019distilbert} for the sentiment analysis task.}
  \label{fig:rle_ih_1}
\end{figure}

\begin{figure}[t]
  \centering
  \includegraphics[page=1,width=.5\textwidth]{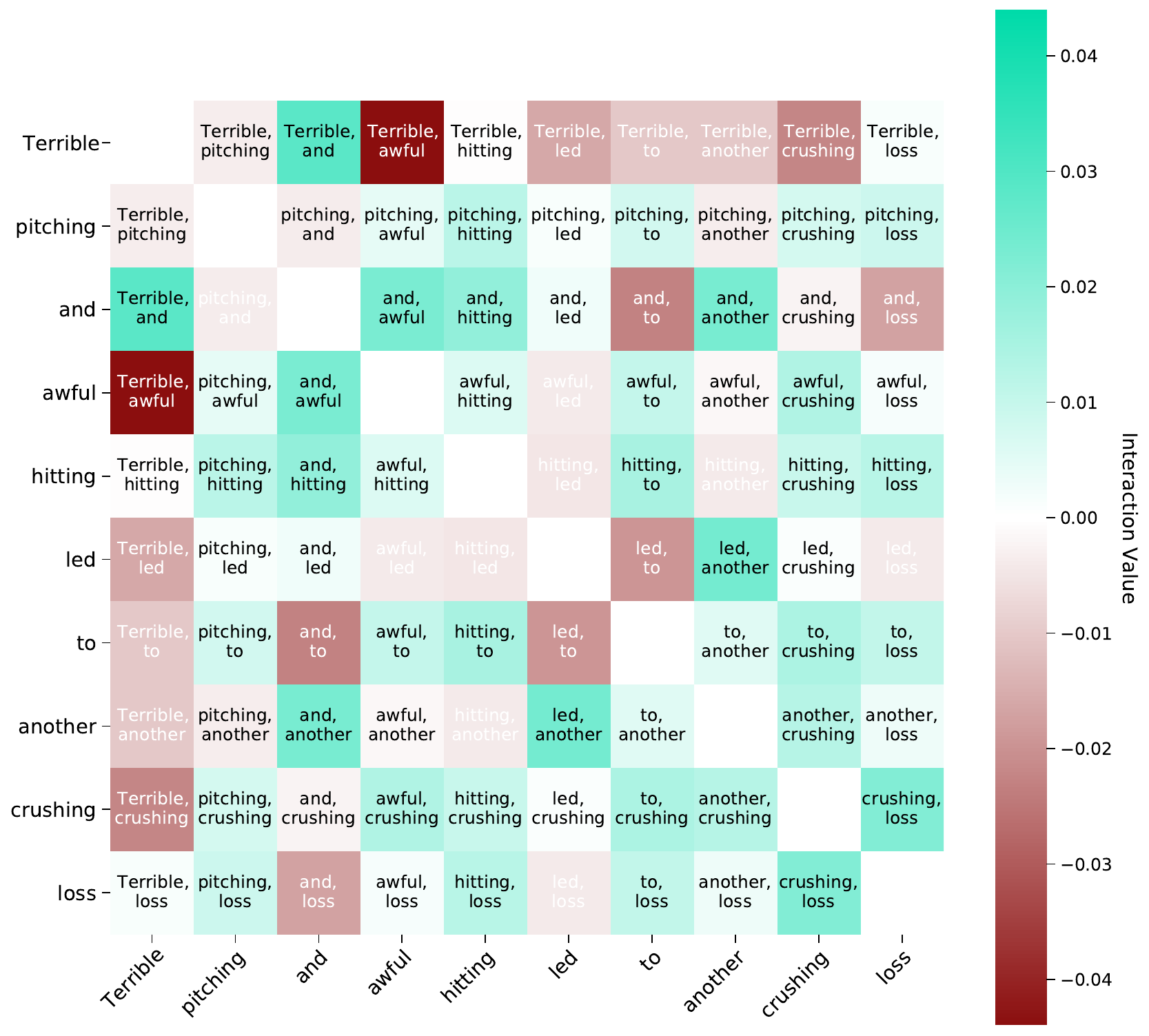}\hspace*{.05\textwidth}%
  \includegraphics[page=1,width=.5\textwidth]{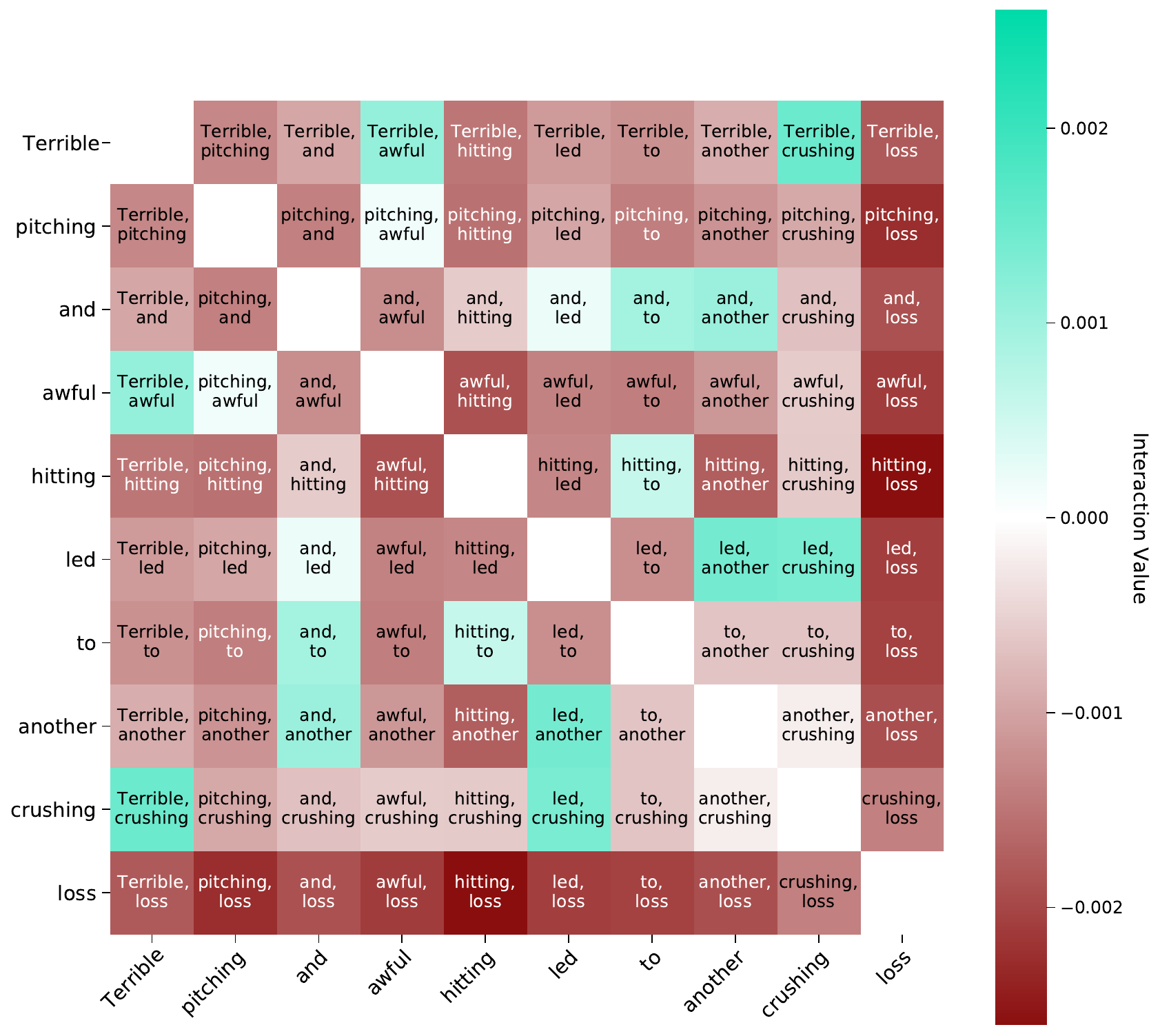}
  \text{ IH \cite{janizek2021explaining} \hspace*{.46\textwidth} RLE (ours)}
  \caption{A comparison of the relational local explanations from IH \cite{janizek2021explaining} and RLE methods for a sentence ``\texttt{Terrible pitching and awful hitting led to another crushing loss}’’, given a pre-trained DistilBERT model \cite{sanh2019distilbert} for the sentiment analysis task.}
  \label{fig:rle_ih_2}
\end{figure}

\begin{figure}[t]
  \centering
  \includegraphics[page=1,width=.5\textwidth]{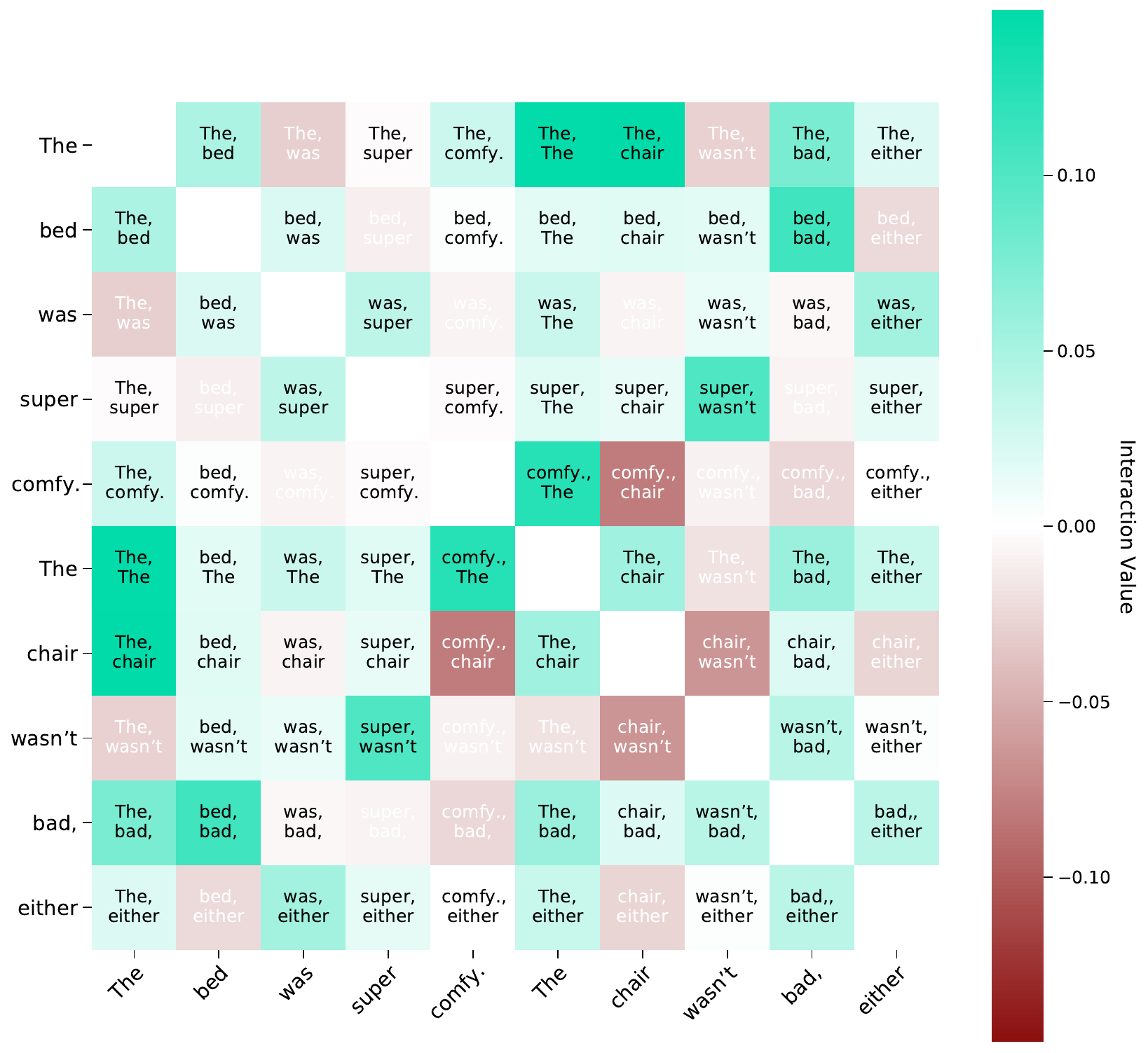}\hspace*{.05\textwidth}%
  \includegraphics[page=1,width=.5\textwidth]{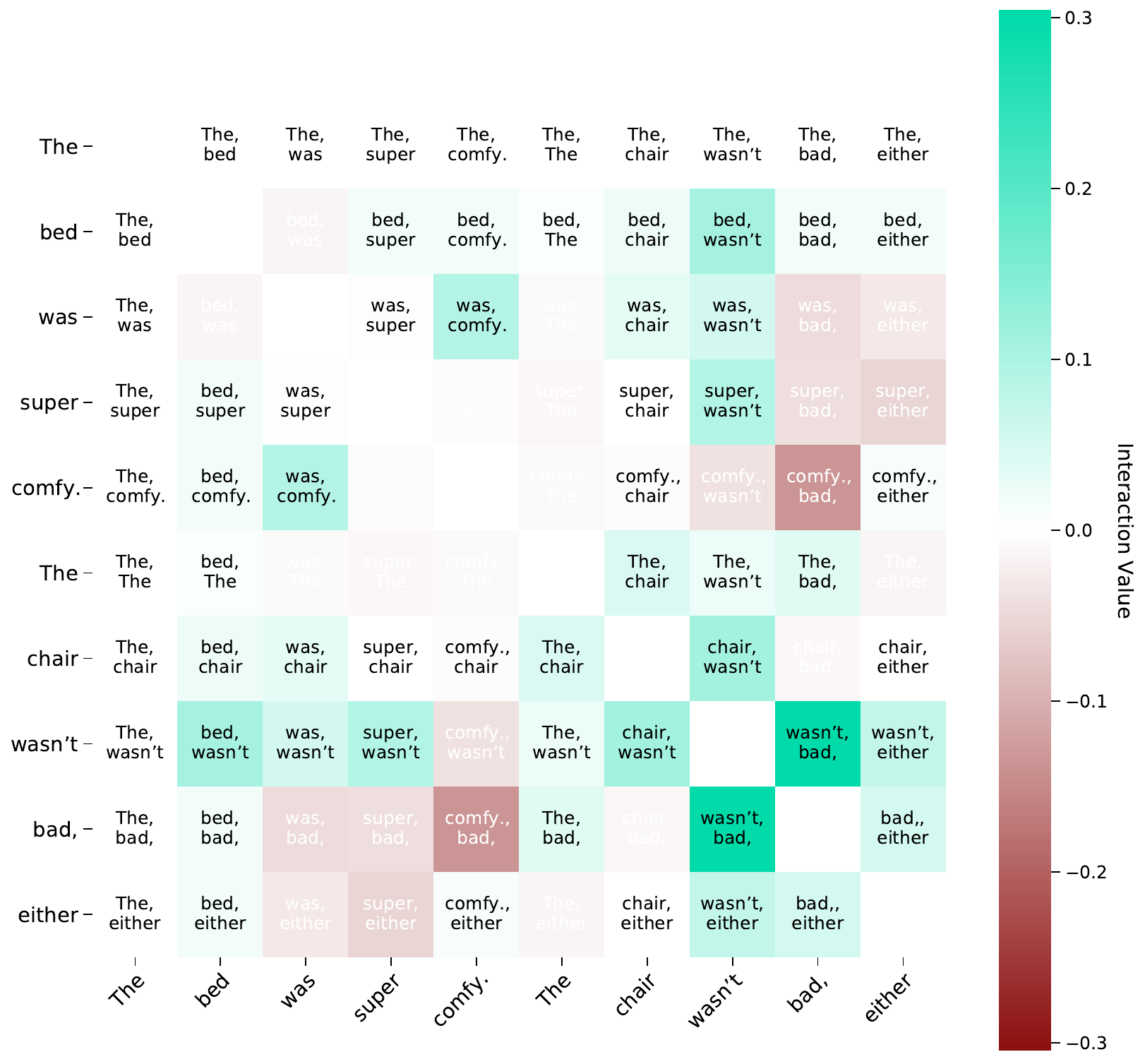}
  \text{ IH \cite{janizek2021explaining} \hspace*{.46\textwidth} RLE (ours)}
  \caption{A comparison of the relational local explanations from IH \cite{janizek2021explaining} and RLE methods for a sentence ``\texttt{The bed was super comfy. The chair wasn’t bad, either}’’, given a pre-trained DistilBERT model \cite{sanh2019distilbert} for the sentiment analysis task.}
  \label{fig:rle_ih_3}
\end{figure}

\begin{figure}[t]
  \centering
  \includegraphics[page=1,width=.5\textwidth]{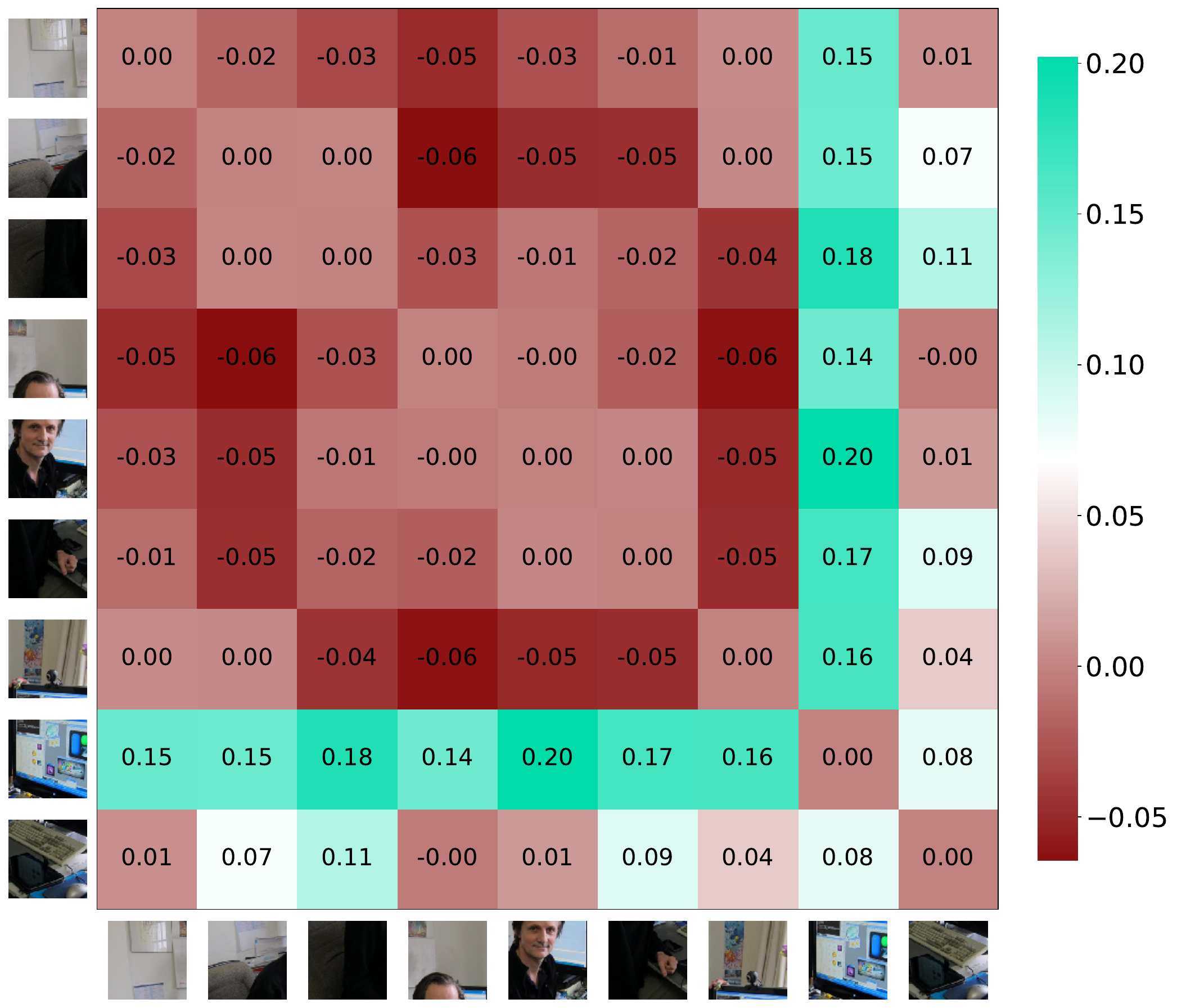}\hspace*{.05\textwidth}
  {\raisebox{15mm}{\includegraphics[width=.25\textwidth]{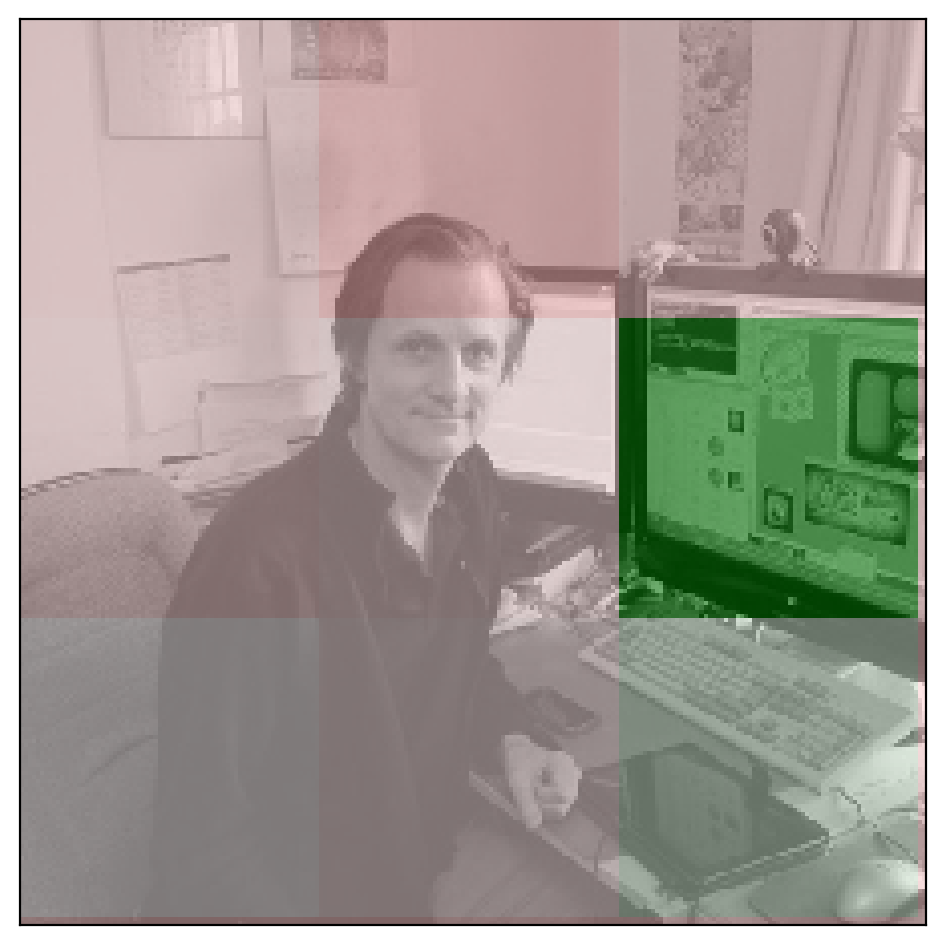}}}
  \caption{An example of the relational local explanation (\textit{left}) and standard local explanation (\textit{right}) for visual data from the proposed RLE framework where green color indicates positive influence, and red negative.  For the task we select a pre-trained ResNet-50 model~\cite{he2016deep} and an image with a class \texttt{desktop computer} from the ImageNet dataset~\cite{deng2009imagenet}, e.g., to uncover a combination of patches that is the most important to a model.}
  \label{fig:example_image_1}
\end{figure}

\begin{figure}[t]
  \centering
  \includegraphics[page=1,width=.5\textwidth]{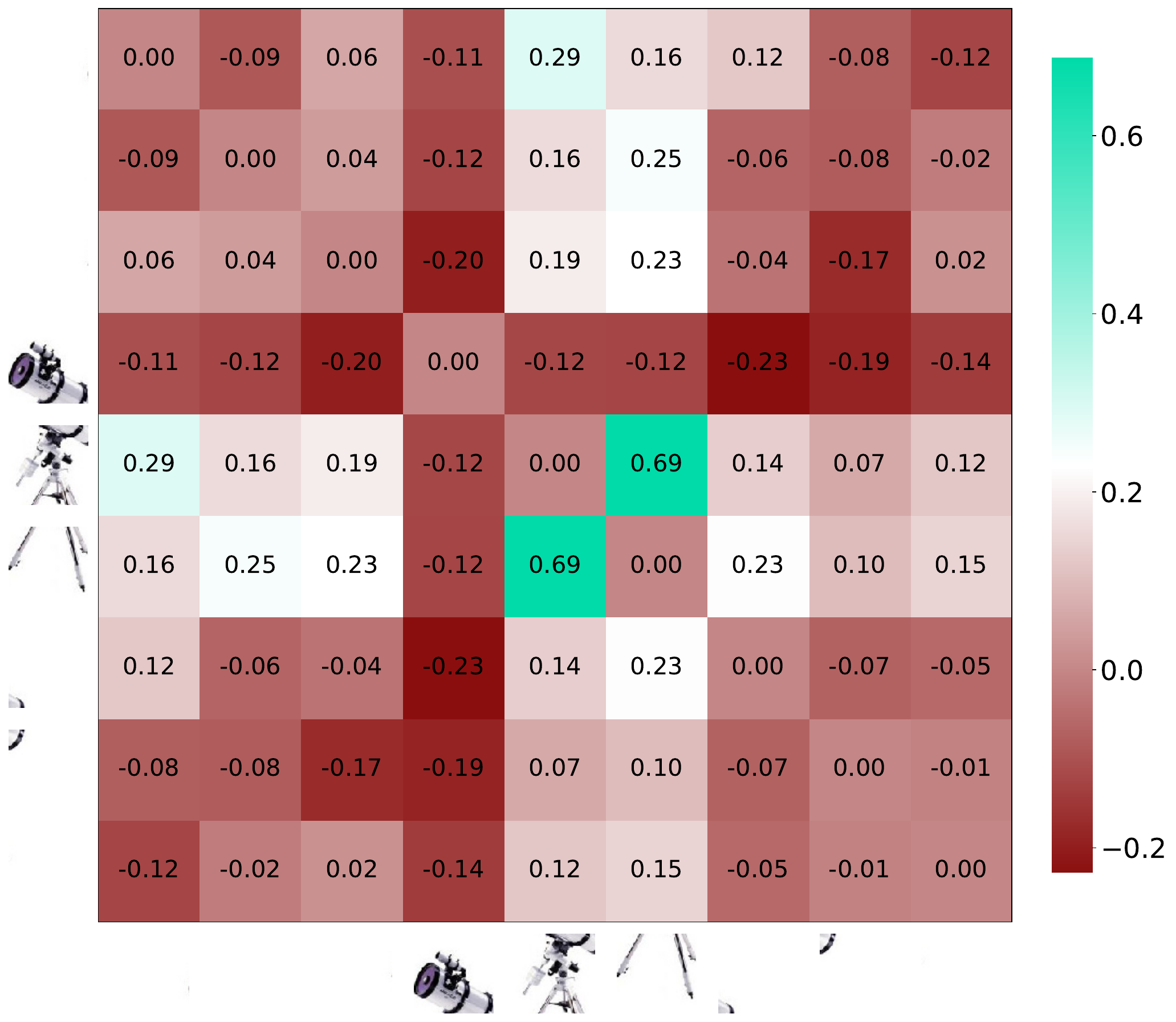}\hspace*{.05\textwidth}
  {\raisebox{15mm}{\includegraphics[width=.25\textwidth]{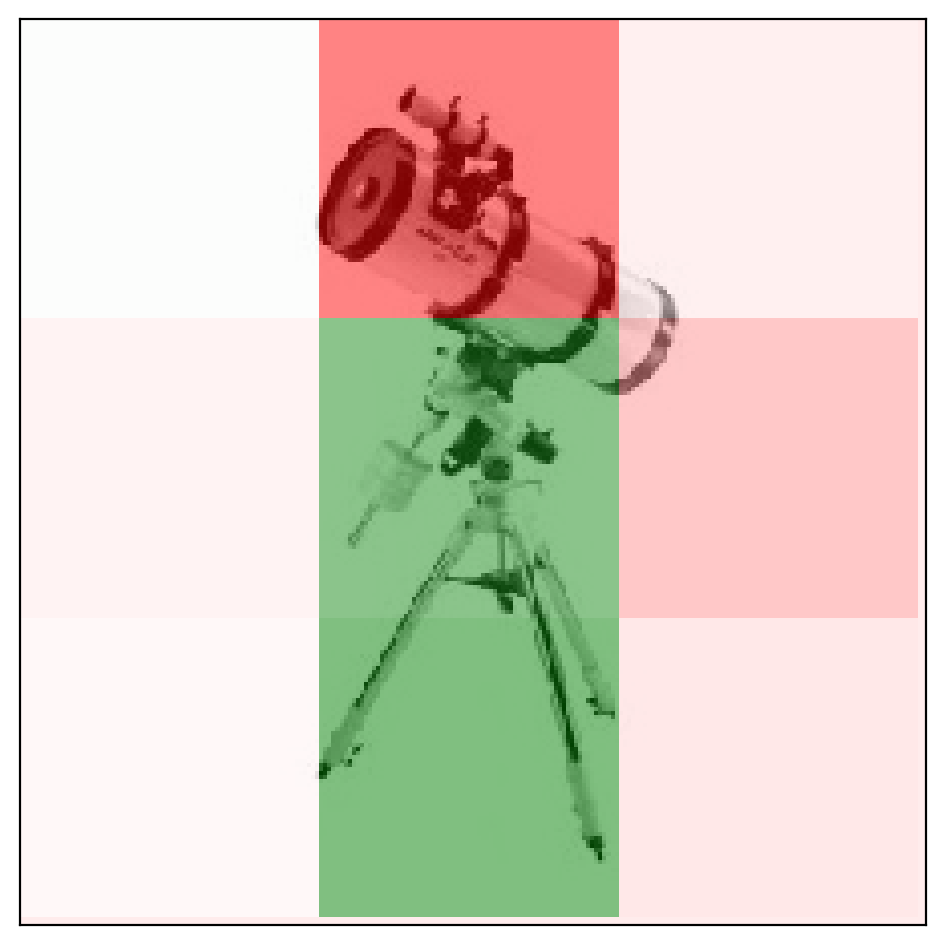}}}
  \caption{An example of the relational local explanation (\textit{left}) and standard local explanation (\textit{right}) for visual data from the proposed RLE framework where green color indicates positive influence, and red negative.  For the task we select a pre-trained ResNet-50 model~\cite{he2016deep} and an image with a class \texttt{tripod} from the ImageNet dataset~\cite{deng2009imagenet}, e.g., to uncover a combination of patches that is the most important to a model.}
  \label{fig:example_image_2}
\end{figure}

\begin{figure}[t]
  \centering
  \includegraphics[page=1,width=.5\textwidth]{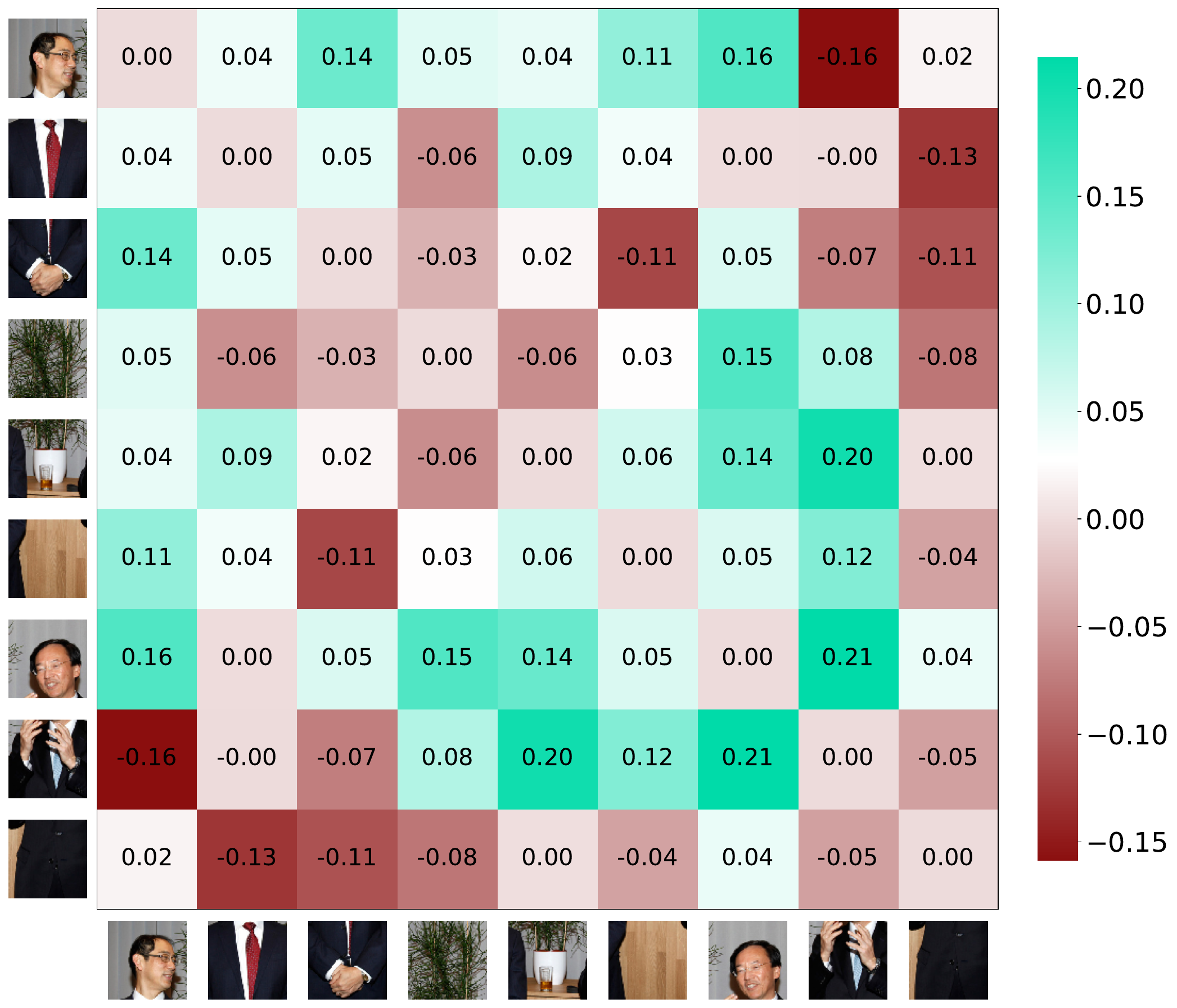}\hspace*{.05\textwidth}
  {\raisebox{15mm}{\includegraphics[width=.25\textwidth]{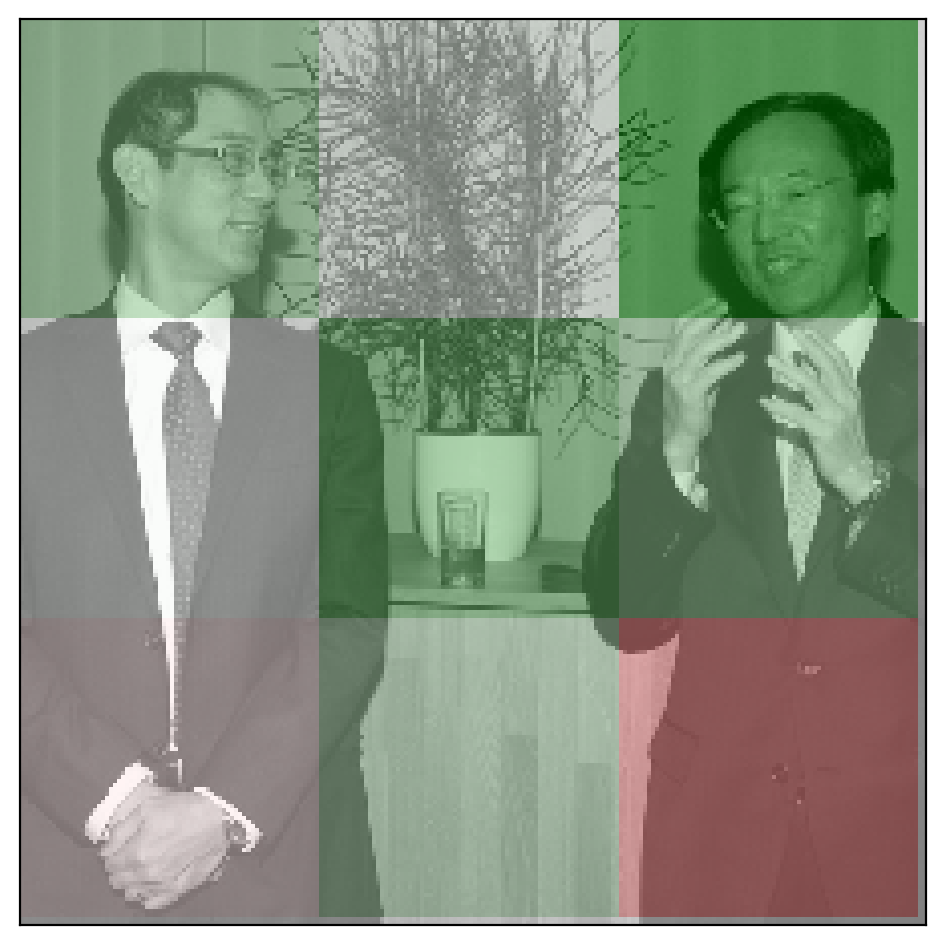}}}
  \caption{An example of the relational local explanation (\textit{left}) and standard local explanation (\textit{right}) for visual data from the proposed RLE framework where green color indicates positive influence, and red negative.  For the task we select a pre-trained ResNet-50 model~\cite{he2016deep} and an image with a class \texttt{groom} from the ImageNet dataset~\cite{deng2009imagenet}, e.g., to uncover a combination of patches that is the most important to a model.}
  \label{fig:example_image_3}
\end{figure}

\begin{figure}[t]
  \centering
  \includegraphics[page=1,width=.5\textwidth]{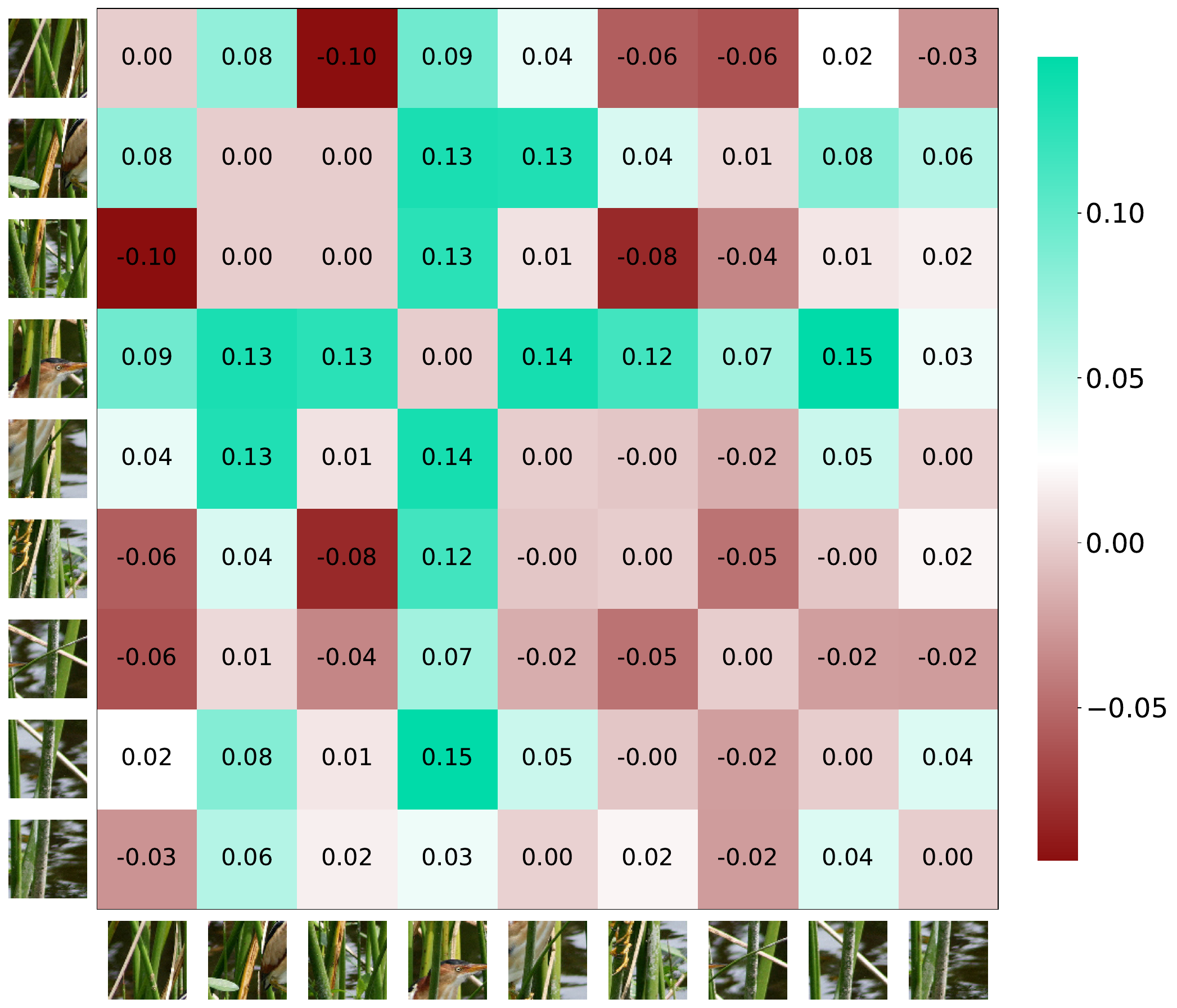}\hspace*{.05\textwidth}
  {\raisebox{15mm}{\includegraphics[width=.25\textwidth]{imgs_sup_mat/image_RLE11_size_3_local.png}}}
  \caption{An example of the relational local explanation (\textit{left}) and standard local explanation (\textit{right}) for visual data from the proposed RLE framework where green color indicates positive influence, and red negative.  For the task we select a pre-trained ResNet-50 model~\cite{he2016deep} and an image with a class \texttt{bittern} from the ImageNet dataset~\cite{deng2009imagenet}, e.g., to uncover a combination of patches that is the most important to a model.}
  \label{fig:example_image_4}
\end{figure}

This section present experimental results on visual (Table \ref{tab:app_visual_experiments}) and textual (Table \ref{tab:text_bench3}) data. Additionally, we show relational local explanations for several visual and text samples in Figures \ref{fig:rle_ih_1}, \ref{fig:rle_ih_2}, \ref{fig:rle_ih_3}, \ref{fig:example_image_1}, \ref{fig:example_image_2}, \ref{fig:example_image_3},  \ref{fig:example_image_4}. We compare results from the proposed algorithm to the baselines of this study: IG~\cite{sundararajan2017axiomatic}, LIME~\cite{ribeiro2016should}, SHAP~\cite{SHAP}, and IH \cite{janizek2021explaining}.

\section{Further Reproducibility Details}
\label{sec:app_repro}

\textbf{Hyperparameters.} We select similar hyperparameters for each baseline to have a fair evaluation; for image data, the number of perturbations (auxiliary samples) $n$ is set to $5000$, and for textual, we set $n$ to $2000$. In our experiments, we observe that a higher number of perturbation steps leads to better quality local explanations. This was also observed in \cite{bansal2020sam}. The rest of the hyperparameters default to a selected package. 

\textbf{RLE plotting function.} For the relational local explanation visualization we apply a plotting function from the IH \cite{janizek2021explaining} official implementation.\footnote{\href{https://github.com/suinleelab/path_explain}{https://github.com/suinleelab/path_explain}} 
From the open-source library Captum \cite{kokhlikyan2020captum}, we utilize plotting function for visualization feature attribution maps on visual data.\footnote{\href{https://captum.ai/api/utilities.html}{https://captum.ai/api/utilities.html}}

\textbf{datasets.} For image data, we utilize samples from ImageNet dataset \cite{deng2009imagenet} provided by the official python package of the SHAP algorithm \cite{SHAP}.\footnote{\href{https://shap.readthedocs.io/en/latest/generated/shap.datasets.imagenet50.html}{https://shap.readthedocs.io/en/latest/generated/shap.datasets.imagenet50.html}} 

\textbf{Pre-trained models.} In this work, we employ the pre-trained ResNet-50 model \cite{he2016deep} from the \textit{torchvision} package \cite{NEURIPS2019_9015}.\footnote{\href{https://pytorch.org/vision/stable/models.html}{https://pytorch.org/vision/stable/models.html}}. We utilize the pre-trained DistilBERT model \cite{sanh2019distilbert} from the \textit{HugginFace} library \cite{wolf2020transformers}.\footnote{\href{https://huggingface.co/distilbert-base-uncased-finetuned-sst-2-english}{https://huggingface.co/distilbert-base-uncased-finetuned-sst-2-english}}

For even better reproducibility, we also report the used package versions in \texttt{requirements.txt} file. It can be found in the corresponding code repository of the RLE algorithm.

\section{The IROF Measure}
\label{app:irof}

\textbf{Choice of the quantitative measure.} In our work, we select the IROF framework \cite{rieger2020irof}, since it allows for an efficient and fairly evaluation of feature attribution methods for the visual data. In comparison to popular approaches for single-pixel-based evaluation of local explanations (e.g., DAUC, IAUC), the chosen evaluation framework uses the super-pixel approach. Since the influence of a single pixel is minimal, the unsupervised grouping of a pixel into local regions allows us for a more fair comparison of the feature attribution methods.  

The IROF approach has several steps: First, the image is divided into coherent segments and bypasses the input features' inter-dependency. According to the creators of the IROF measure, we use the SLIC method for unsupervised image segmentation \cite{achanta2012slic}. 

Formally, the IROF measure is defined as follows: 

\begin{equation}
	\text{IROF}(e_j) = \frac{1}{N} \sum_{n =1}^{N} \text{AOC} \left(\frac{f(x_n^{0})_y}{f(x_n^{0})_y} \right)_{l=0}^{L}
\end{equation}

where $e_j$ is a local feature attribution map, $N$ is the number of super-pixels, $x^0$ is an image to explain, $f$ is a black-box model, $y$ is a target, and $L$  represents the class score based on how many segments of the image were removed. Also, the proposed measure utilized the area over the curve (AOC) function. The higher the IROF score, the more plausible the local explanations, i.e., the more information was collected.

\begin{table*}[t]
\begin{tabularx}{\textwidth}{c @{\extracolsep{\fill}} c @{\extracolsep{\fill}} c@{\extracolsep{\fill}}c@{\extracolsep{\fill}} c@{\extracolsep{\fill}}}
\toprule
$\,\,$Original & IG \cite{sundararajan2017axiomatic} & LIME \cite{ribeiro2016should} &  SHAP  \cite{SHAP} & RLE (Ours) \\
\midrule

\includegraphics[width=25mm]{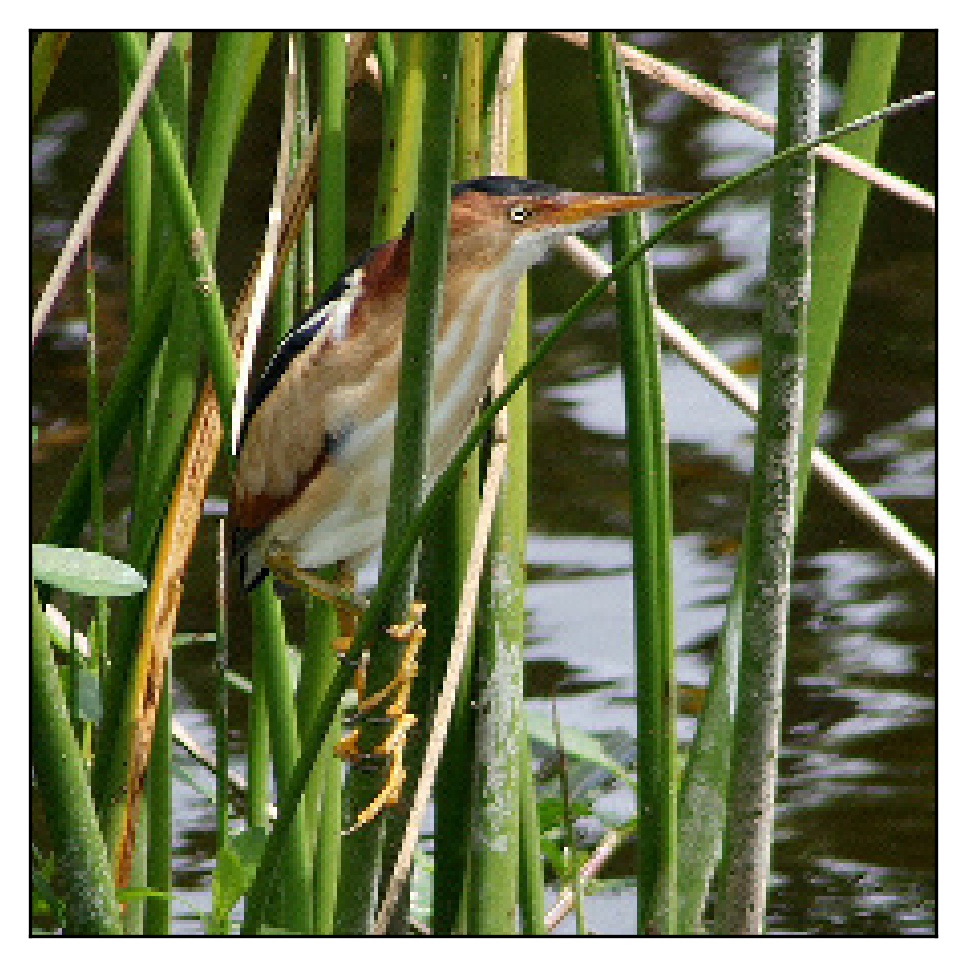} & 
\includegraphics[width=25mm]{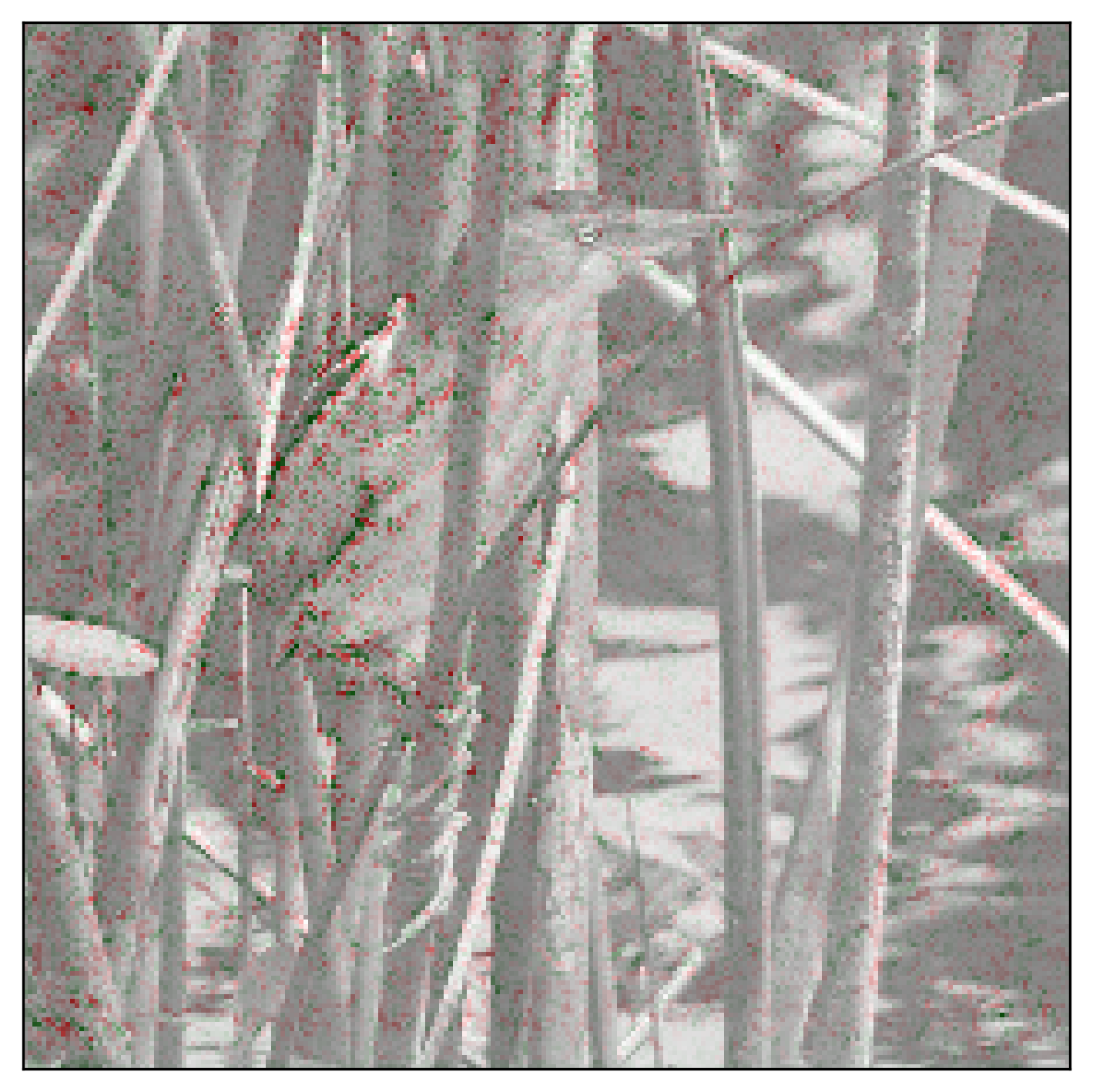} & 
\includegraphics[width=25mm]{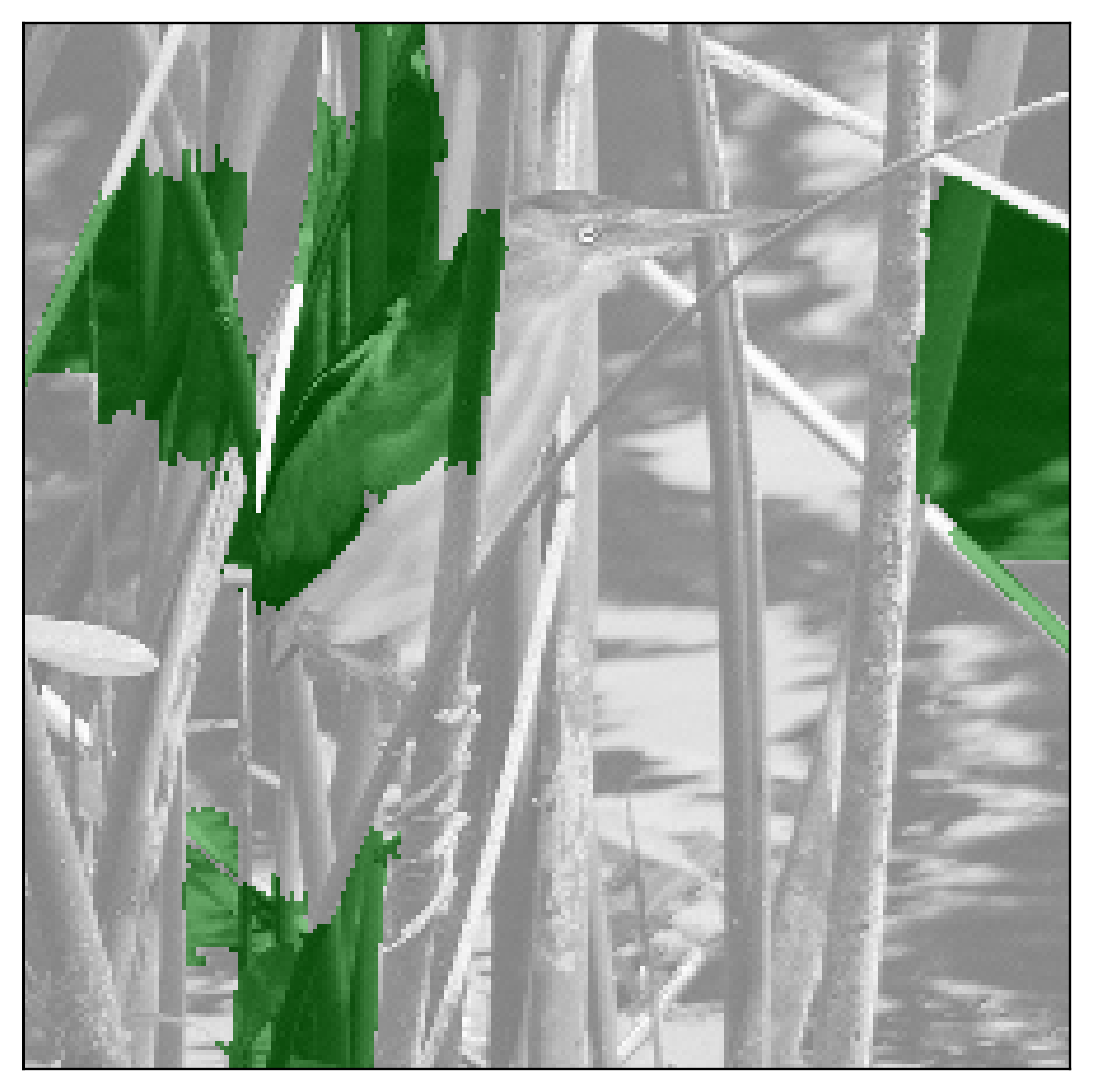} & 
\includegraphics[width=25mm]{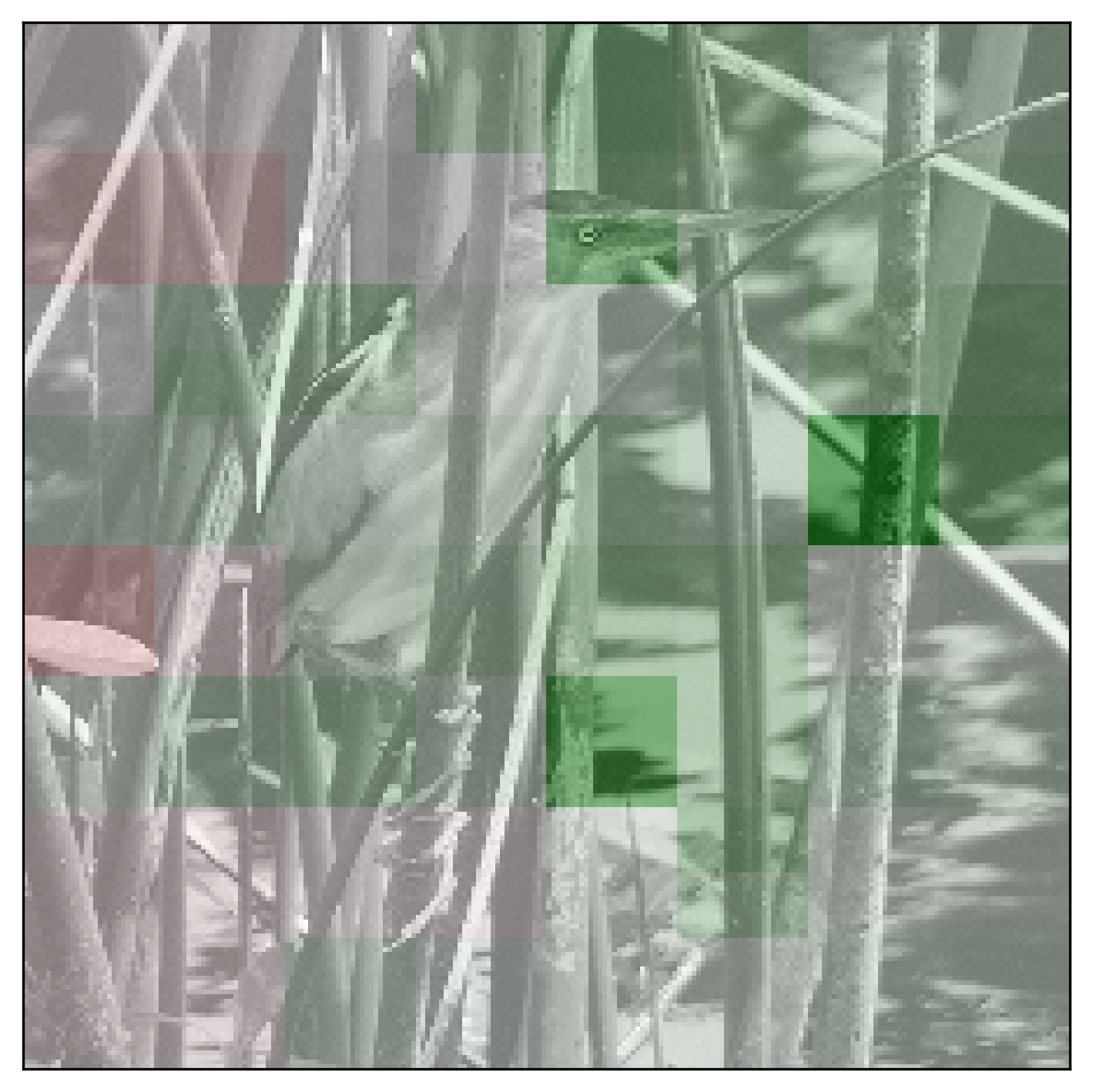} & 
\includegraphics[width=25mm]{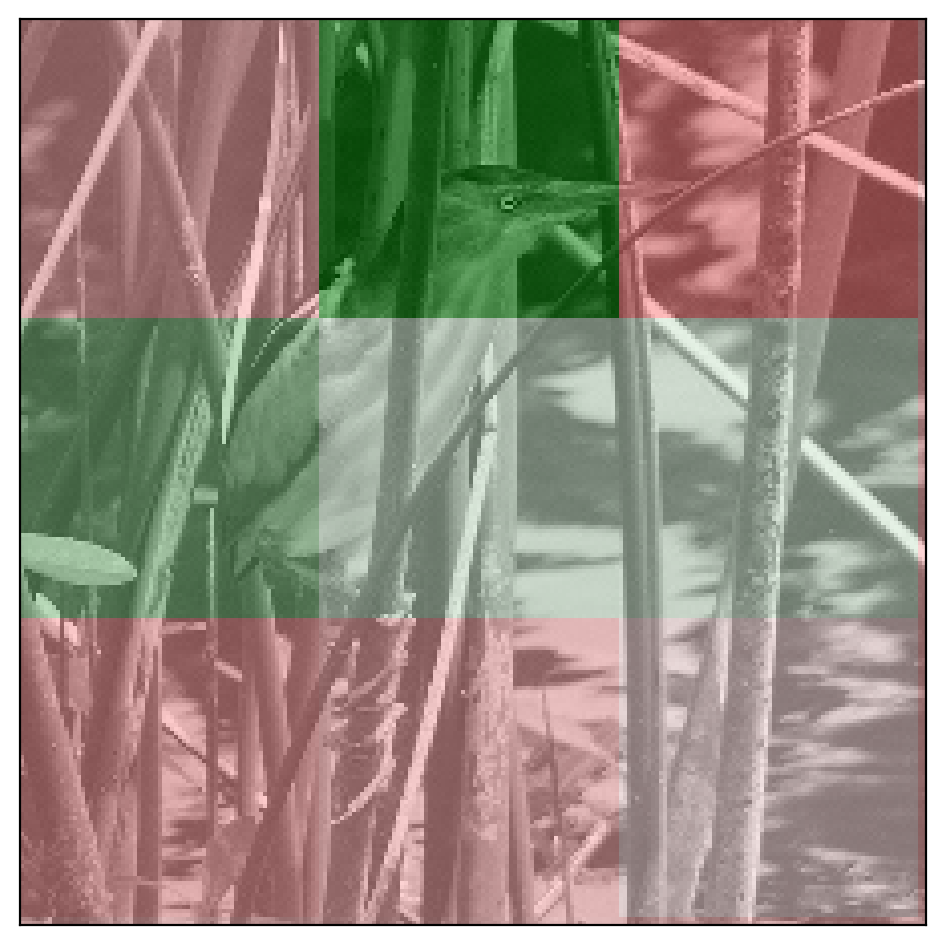} 
\\

\includegraphics[width=25mm]{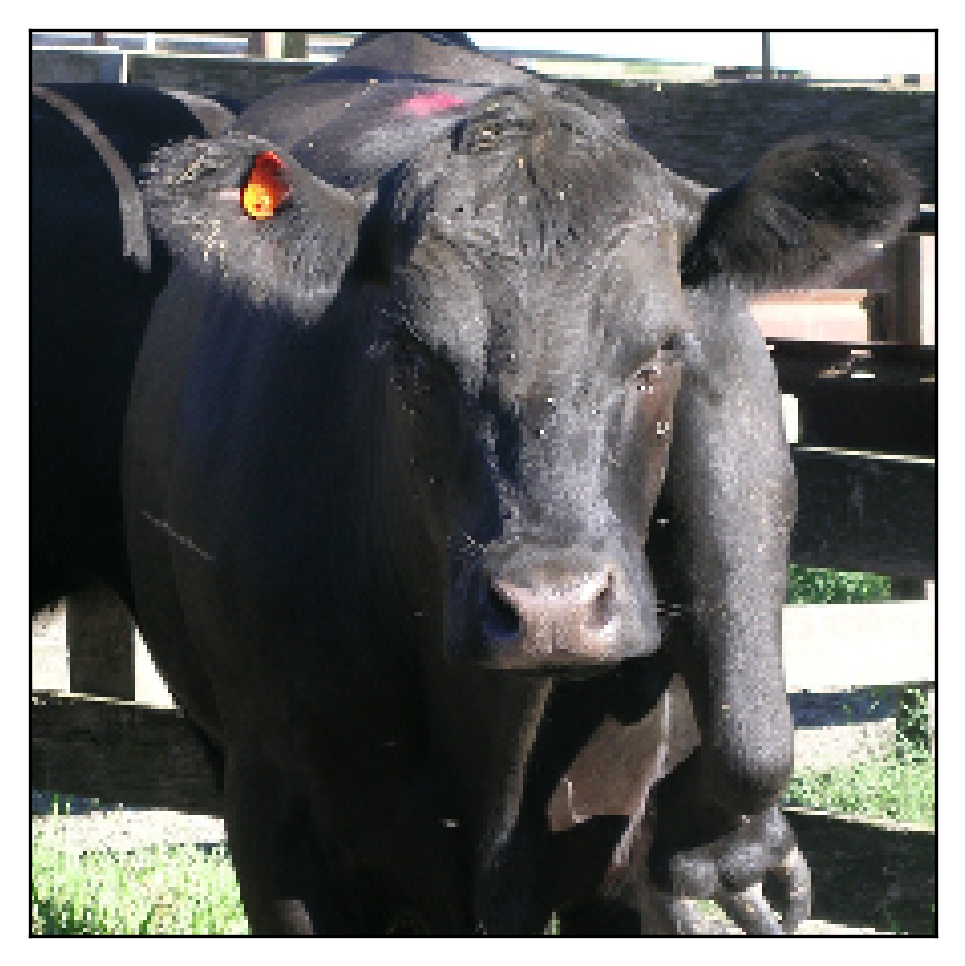} & 
\includegraphics[width=25mm]{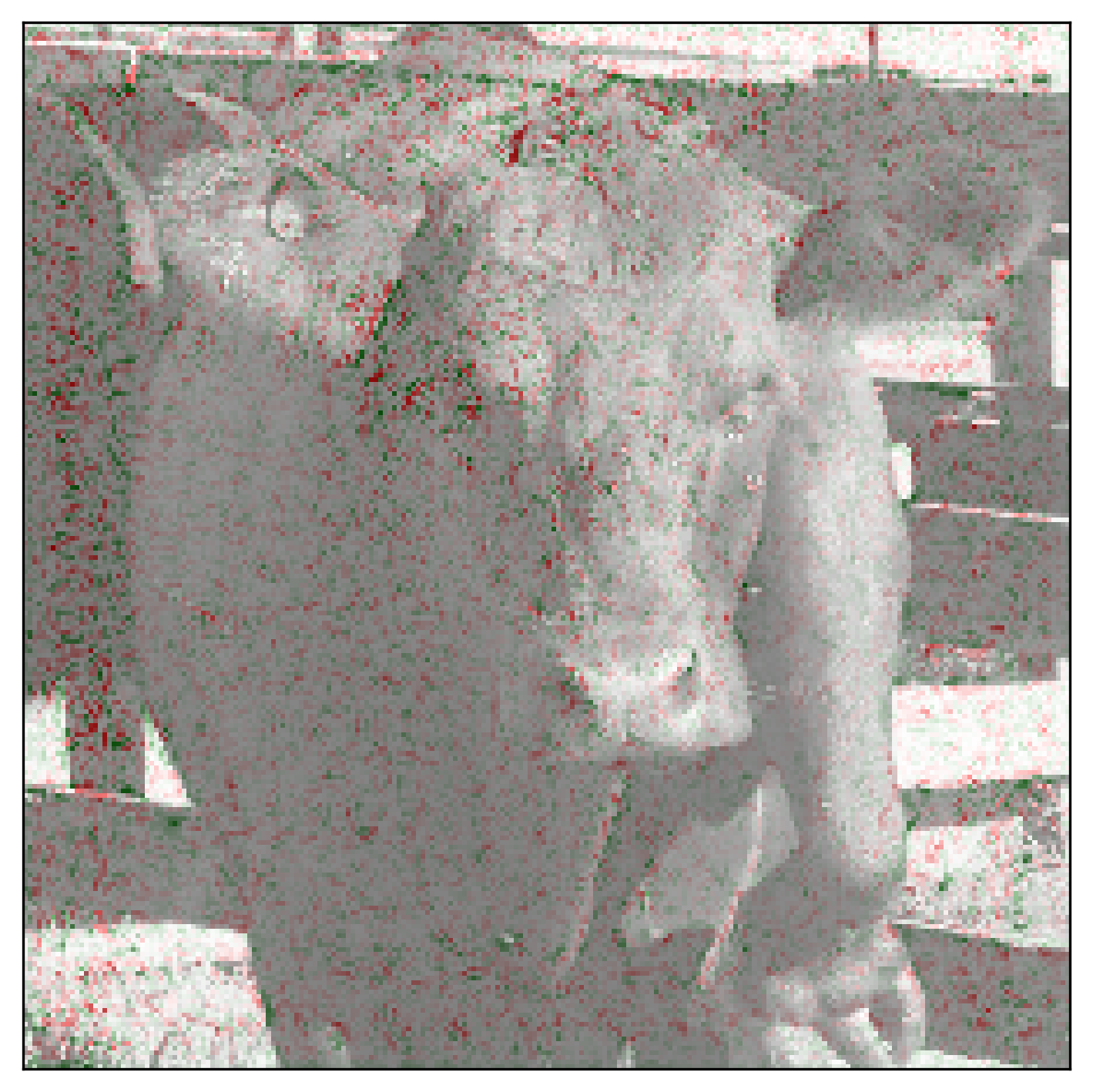} & 
\includegraphics[width=25mm]{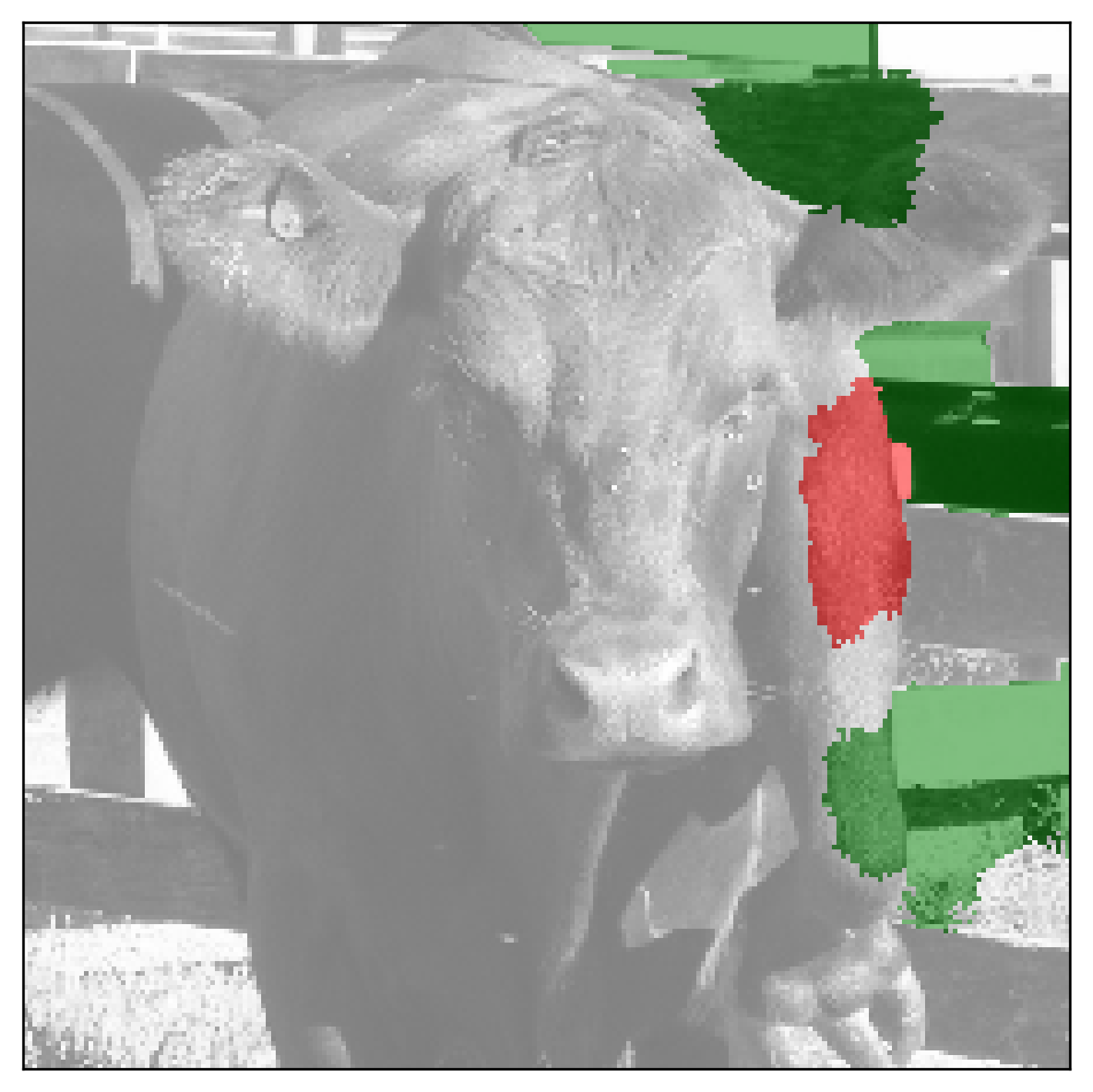} & 
\includegraphics[width=25mm]{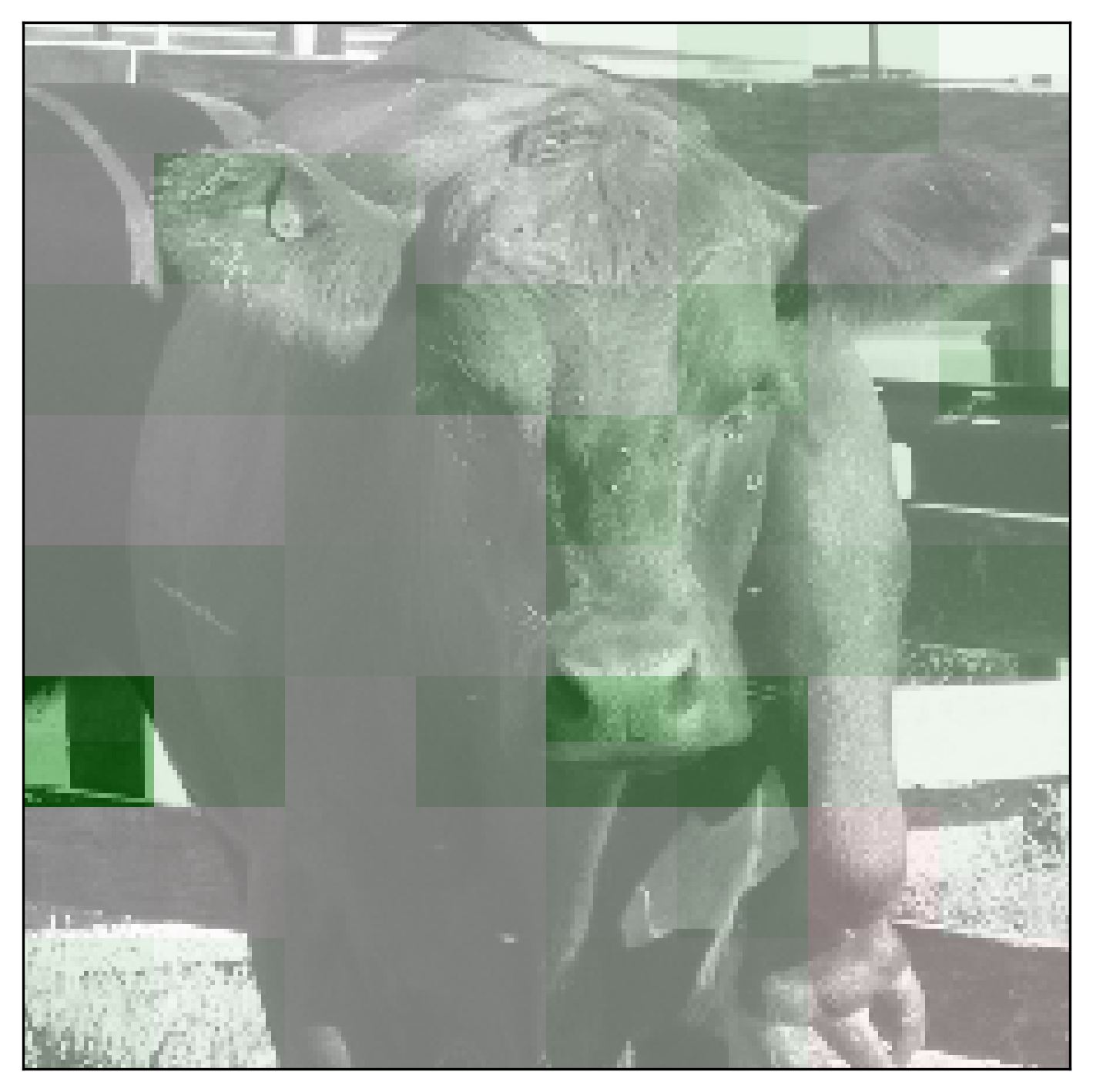} & 
\includegraphics[width=25mm]{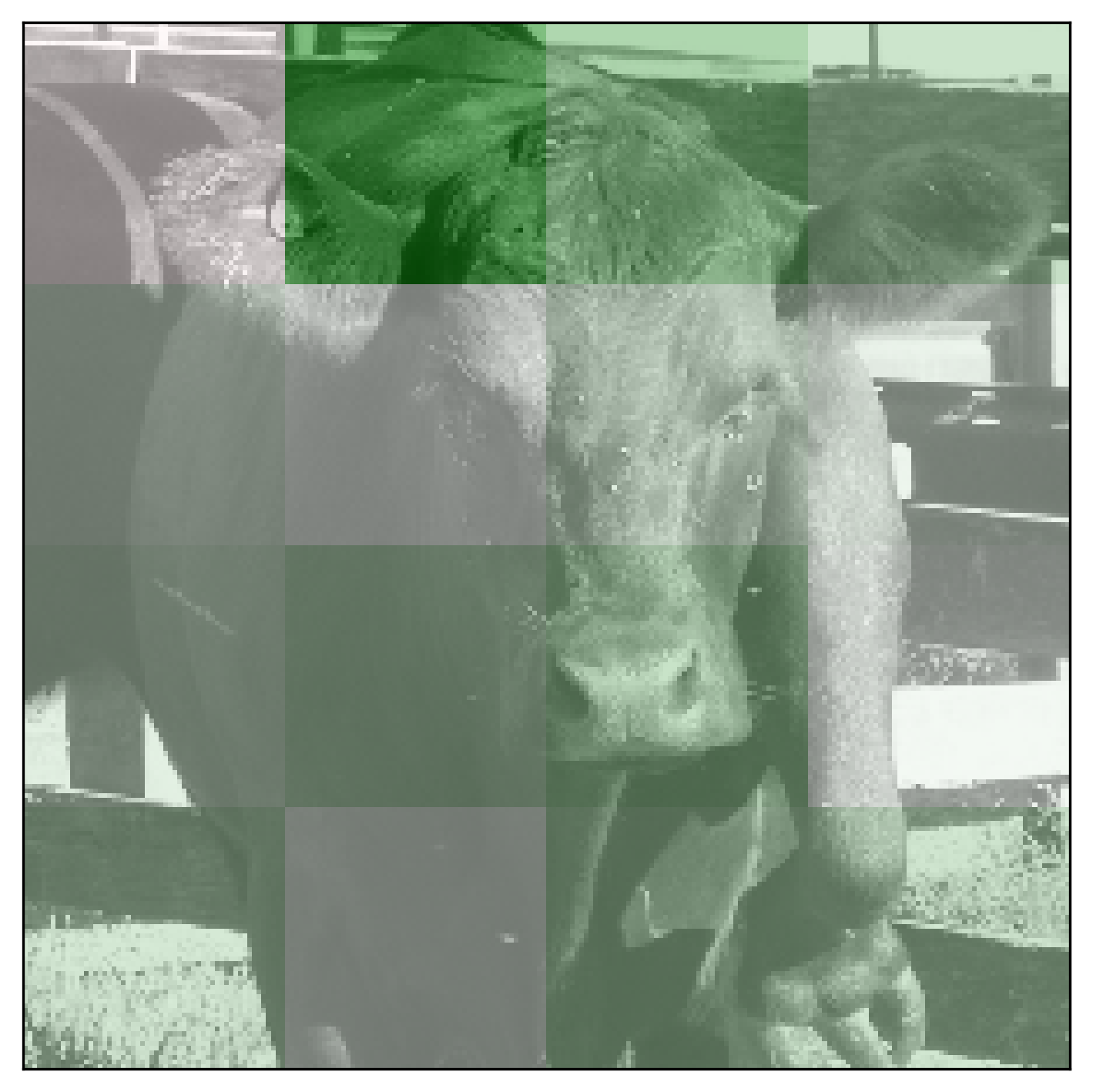} 

\\

\includegraphics[width=25mm]{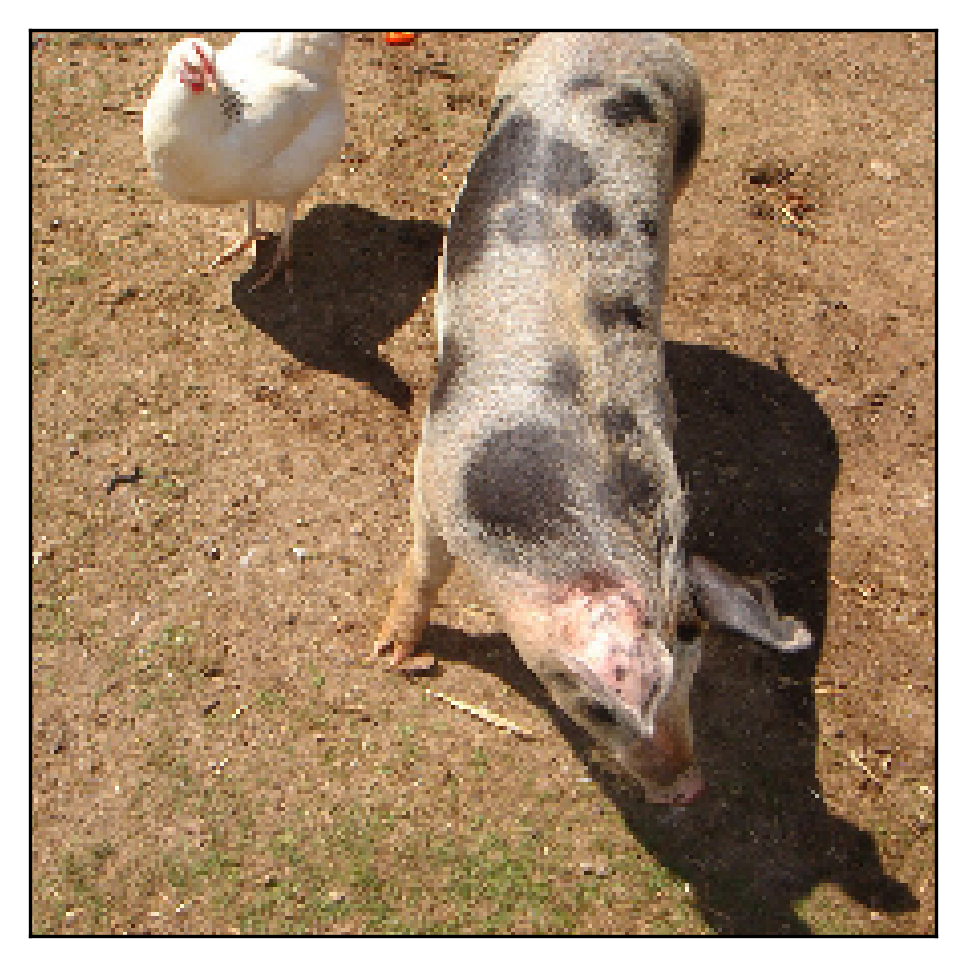} & 
\includegraphics[width=25mm]{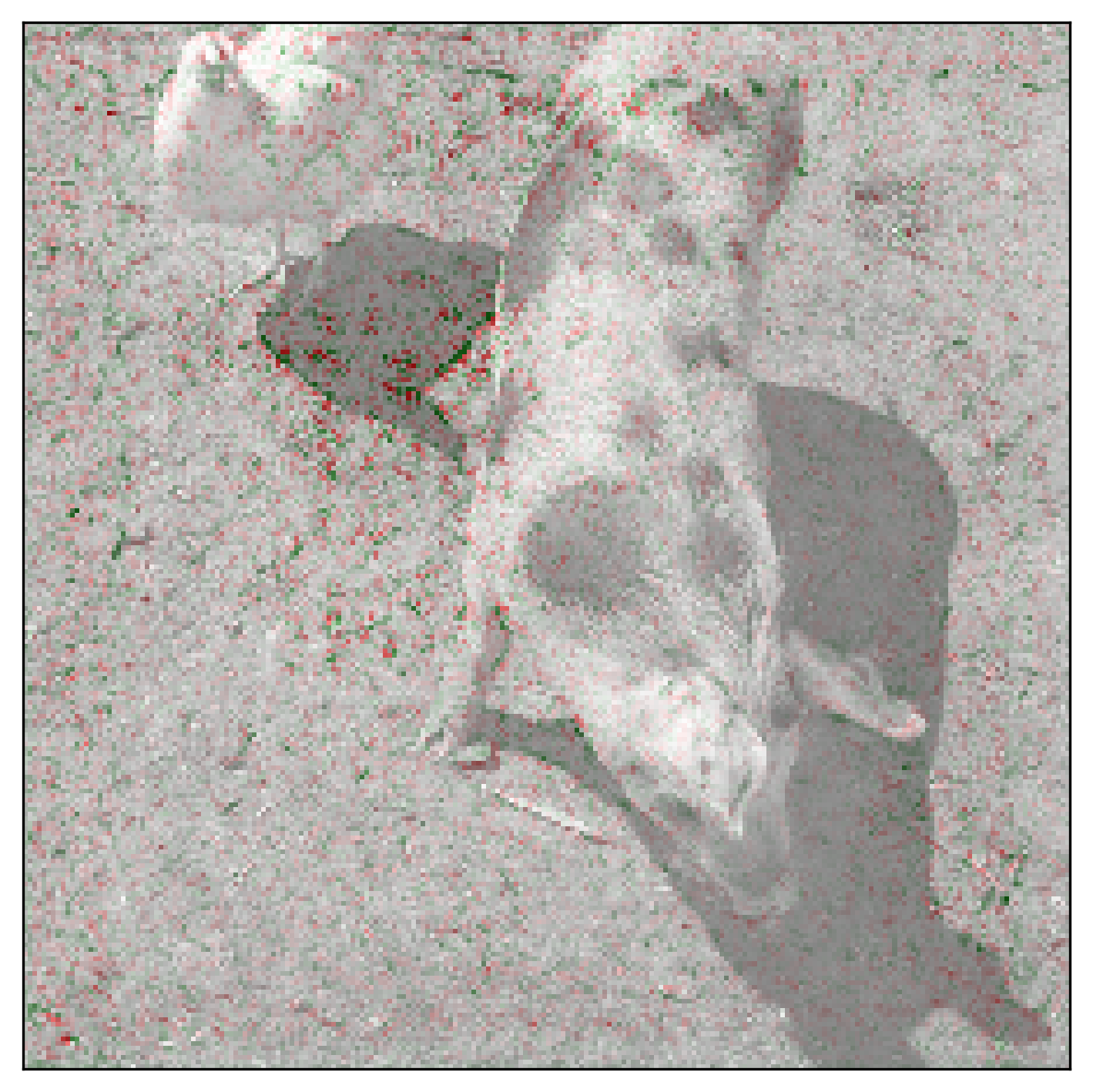} & 
\includegraphics[width=25mm]{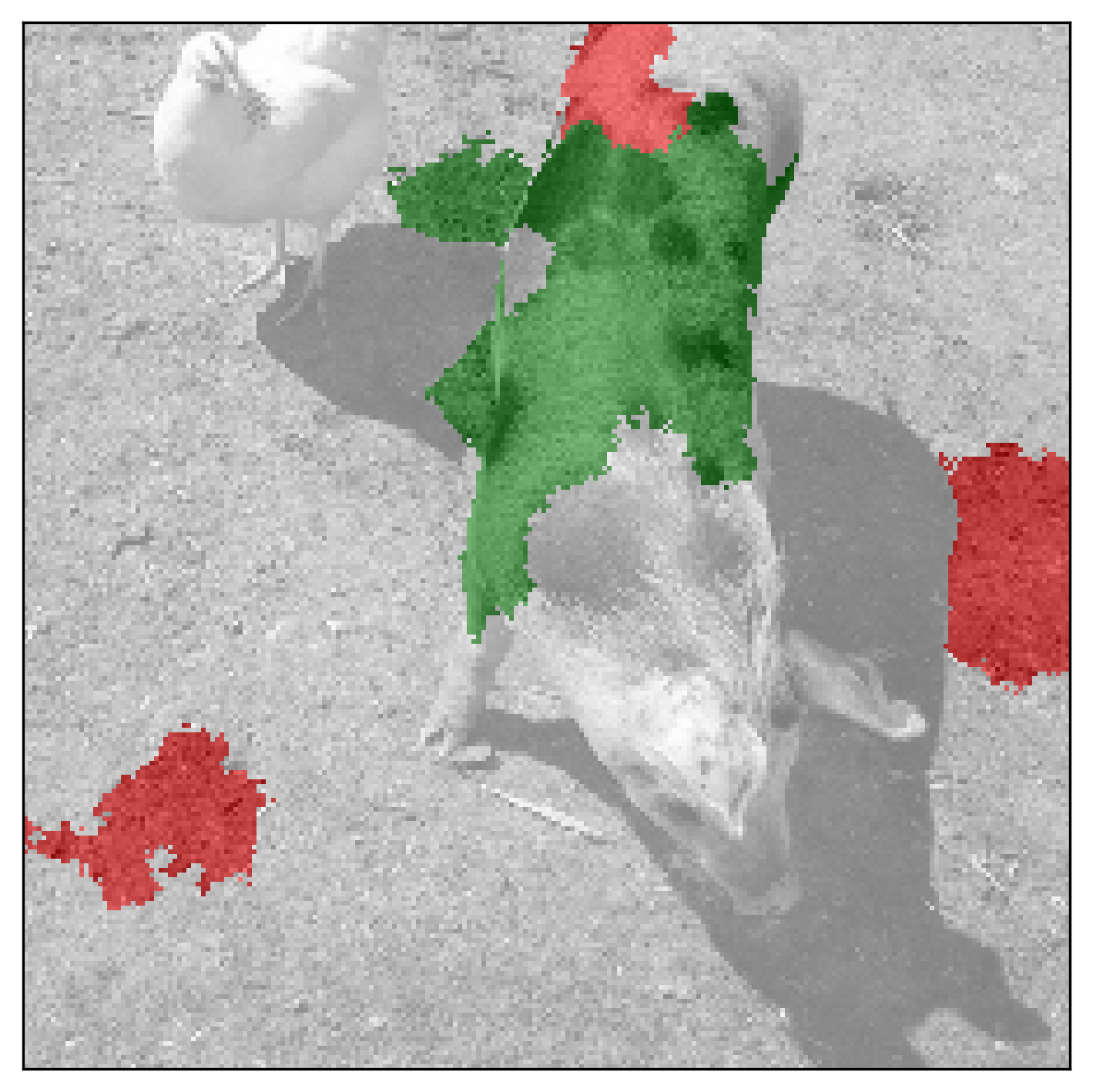} & 
\includegraphics[width=25mm]{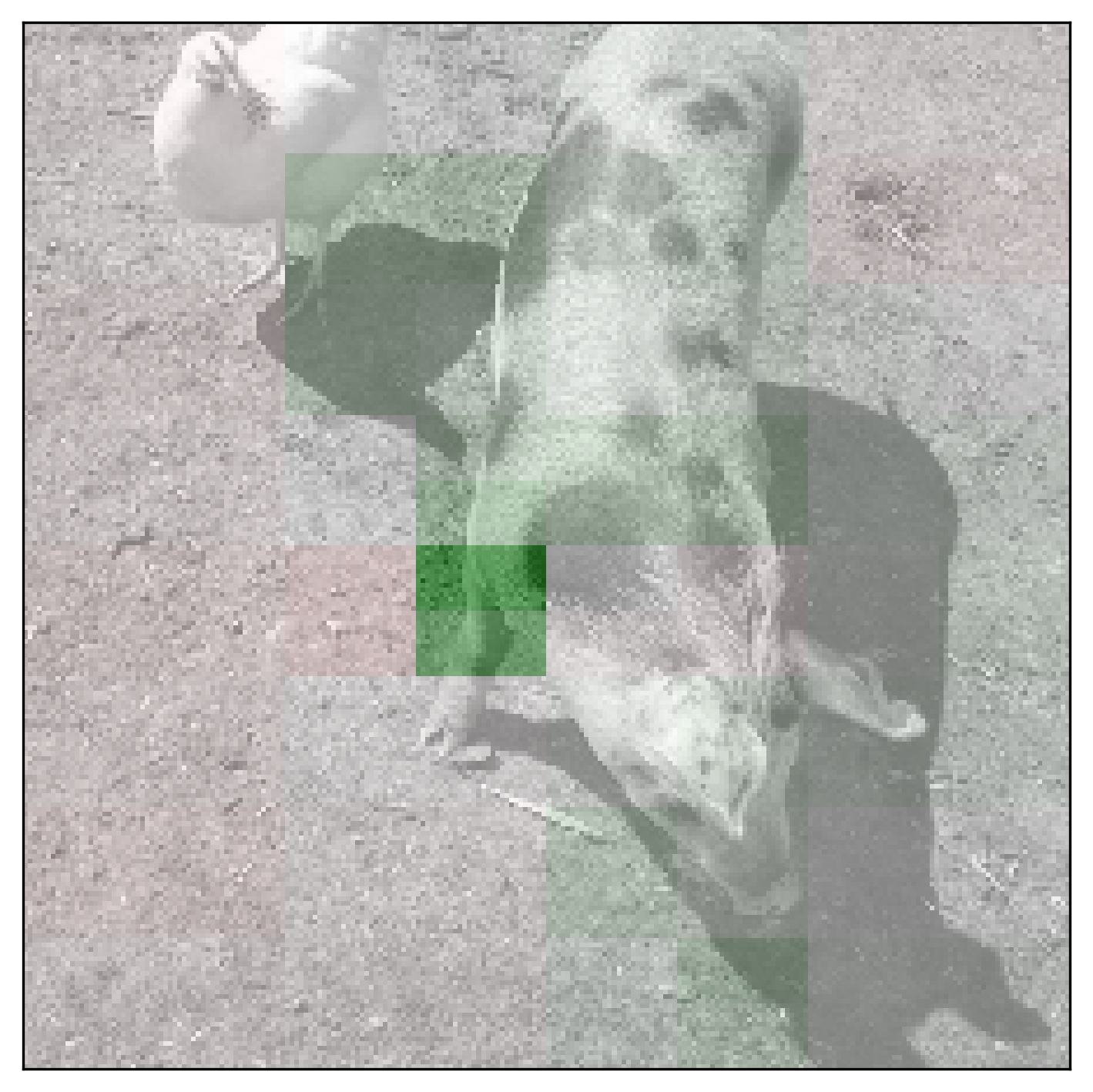} & 
\includegraphics[width=25mm]{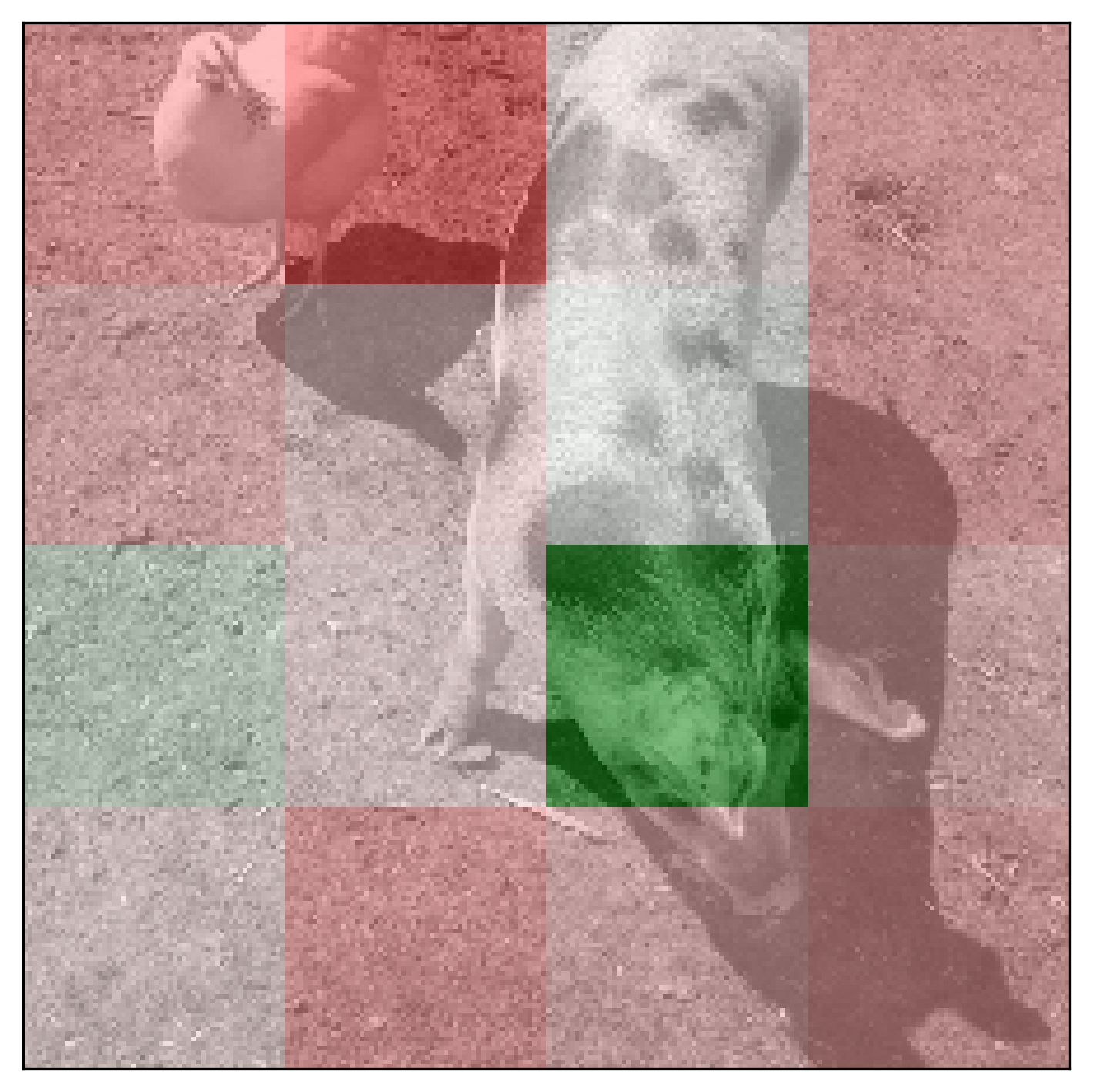} 
\\

\includegraphics[width=25mm]{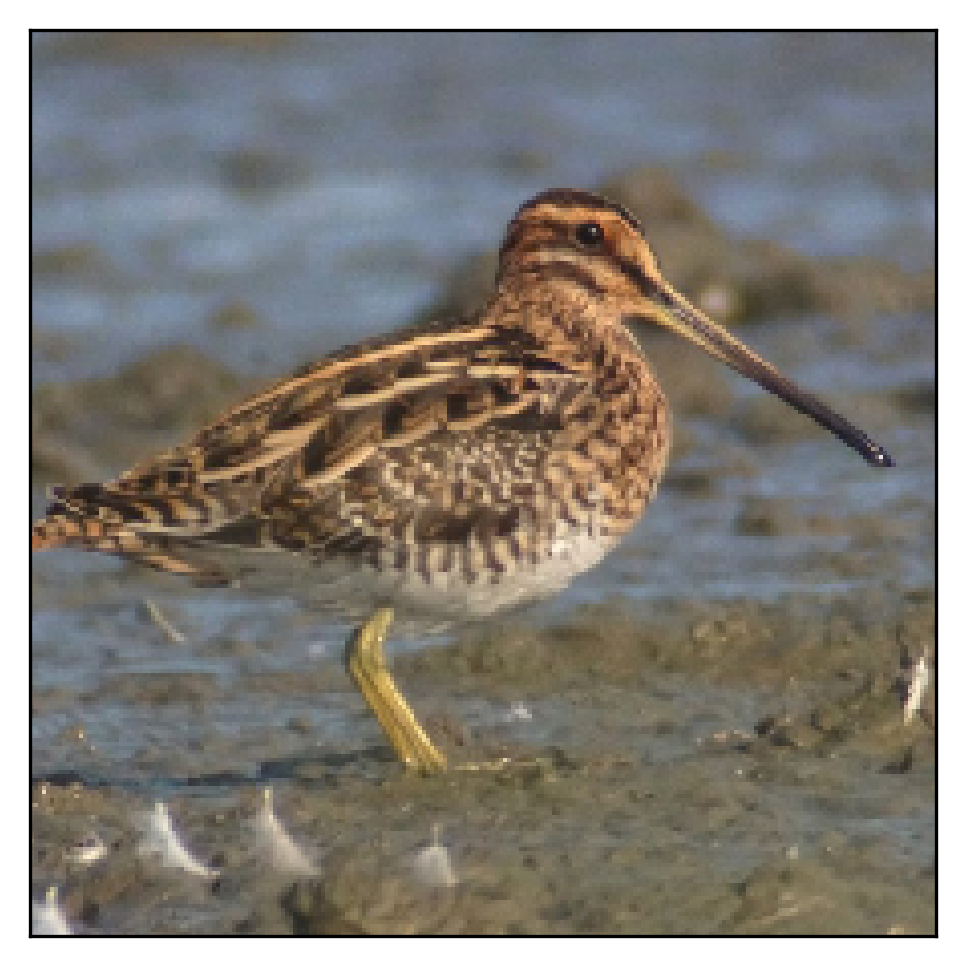} & 
\includegraphics[width=25mm]{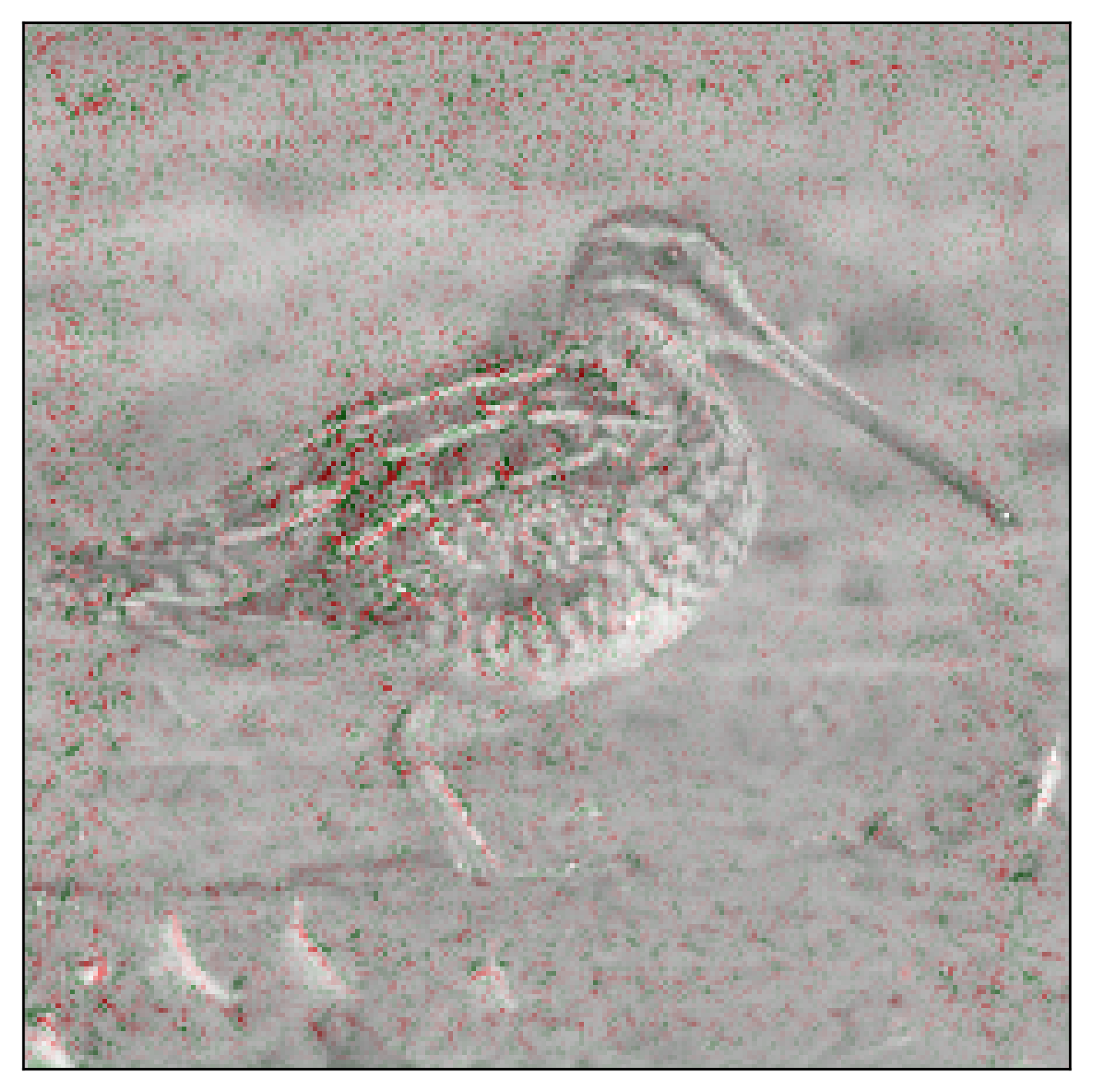} & 
\includegraphics[width=25mm]{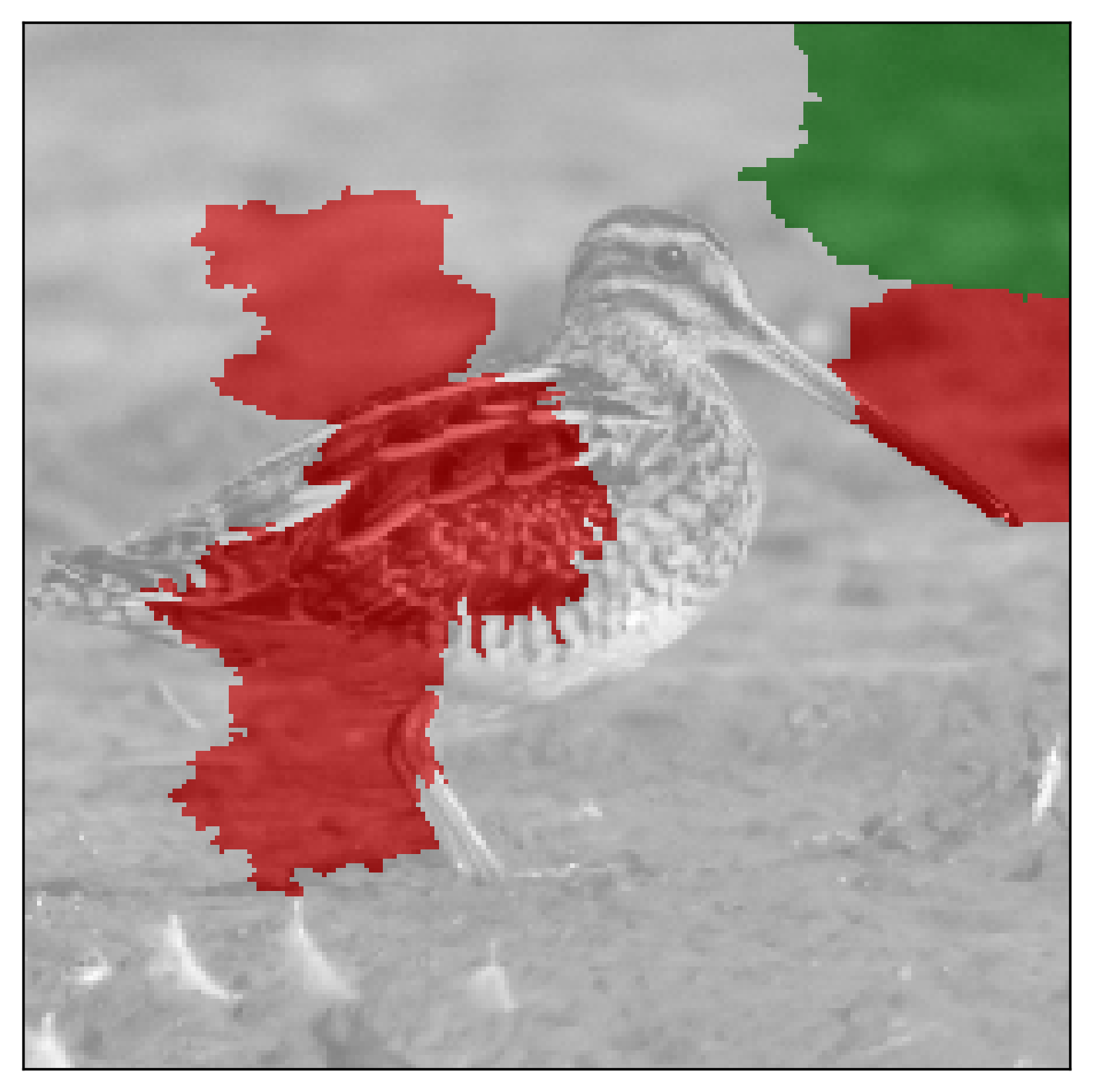} & 
\includegraphics[width=25mm]{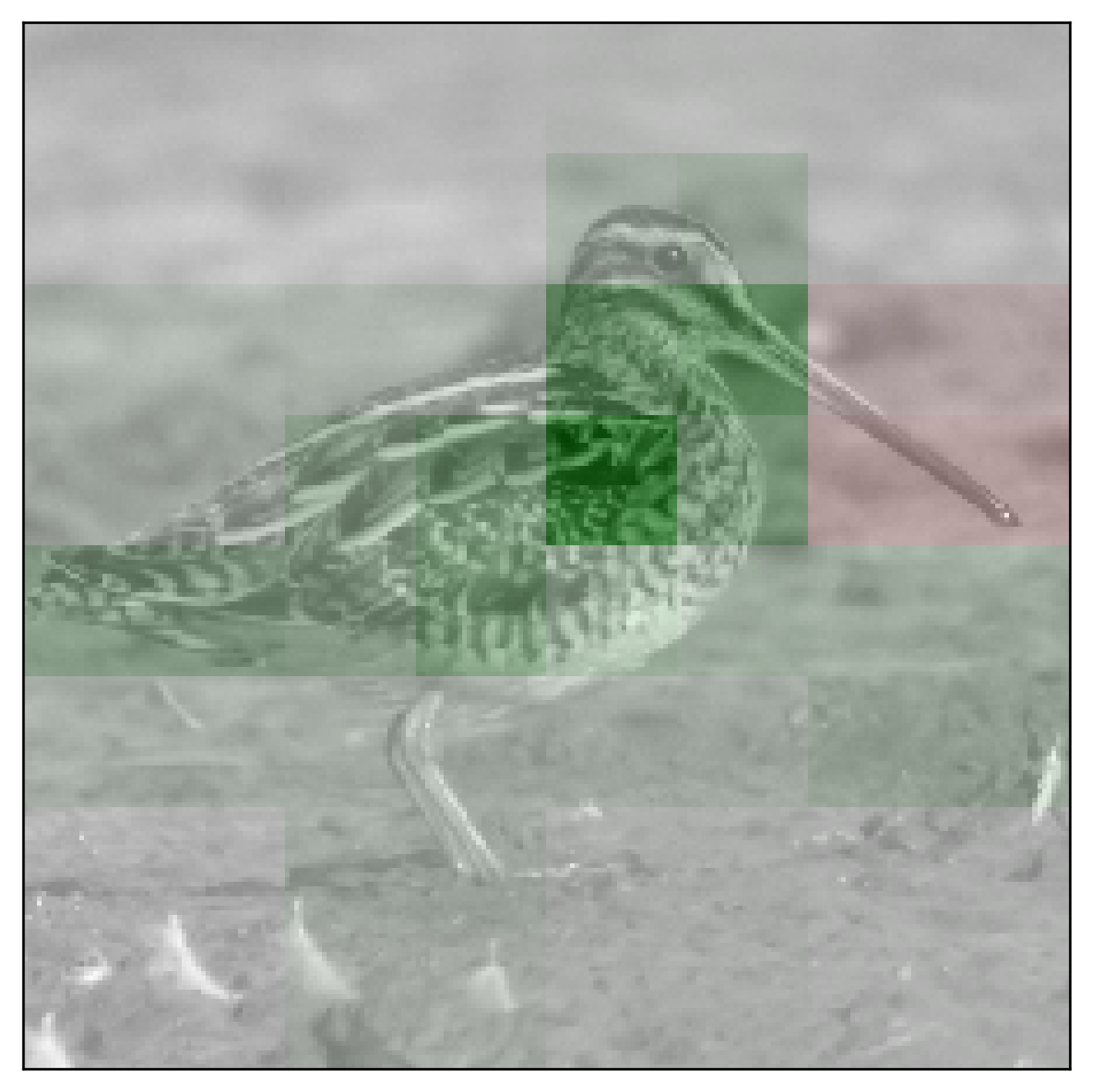} & 
\includegraphics[width=25mm]{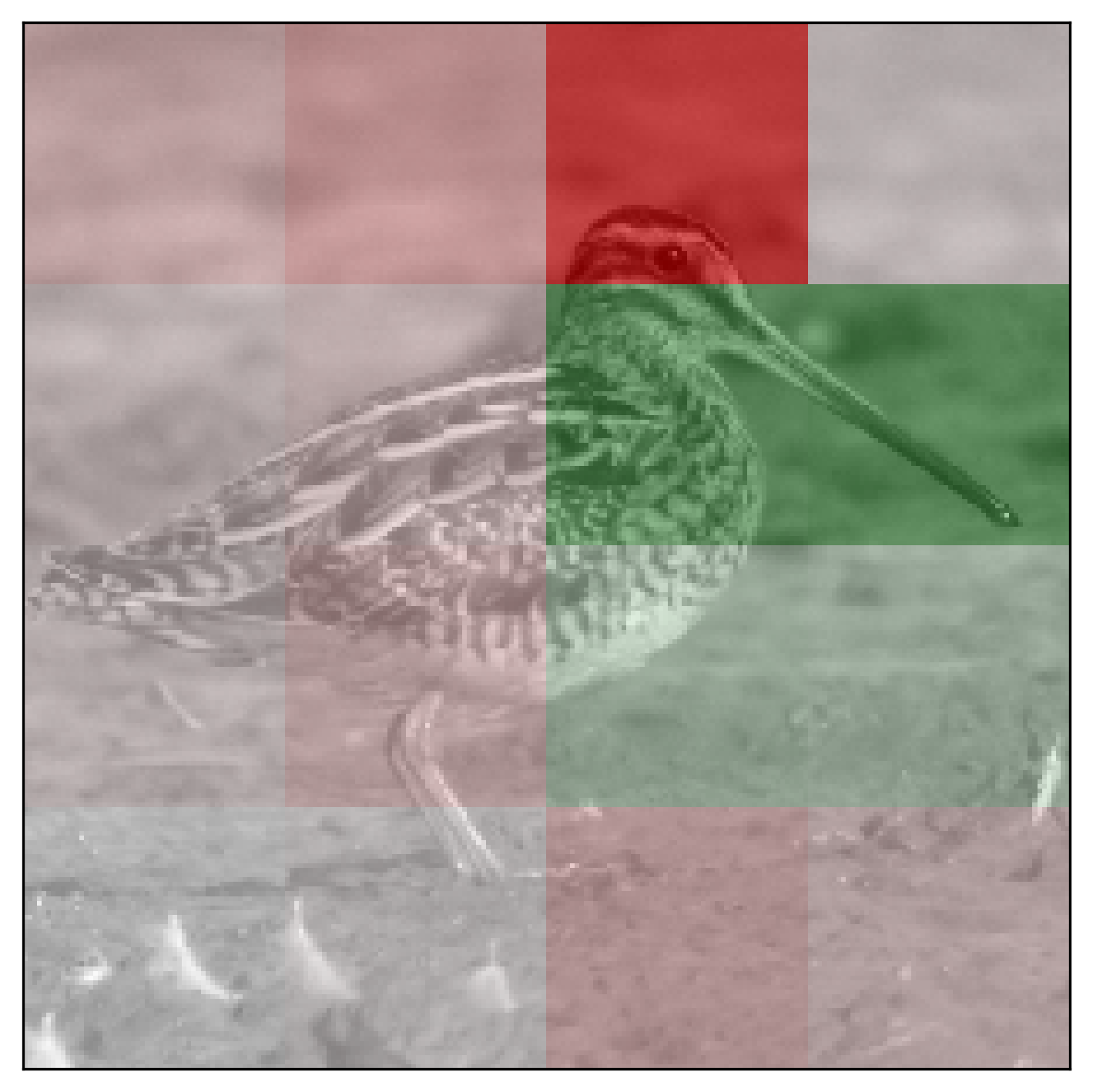} 
\\

\includegraphics[width=25mm]{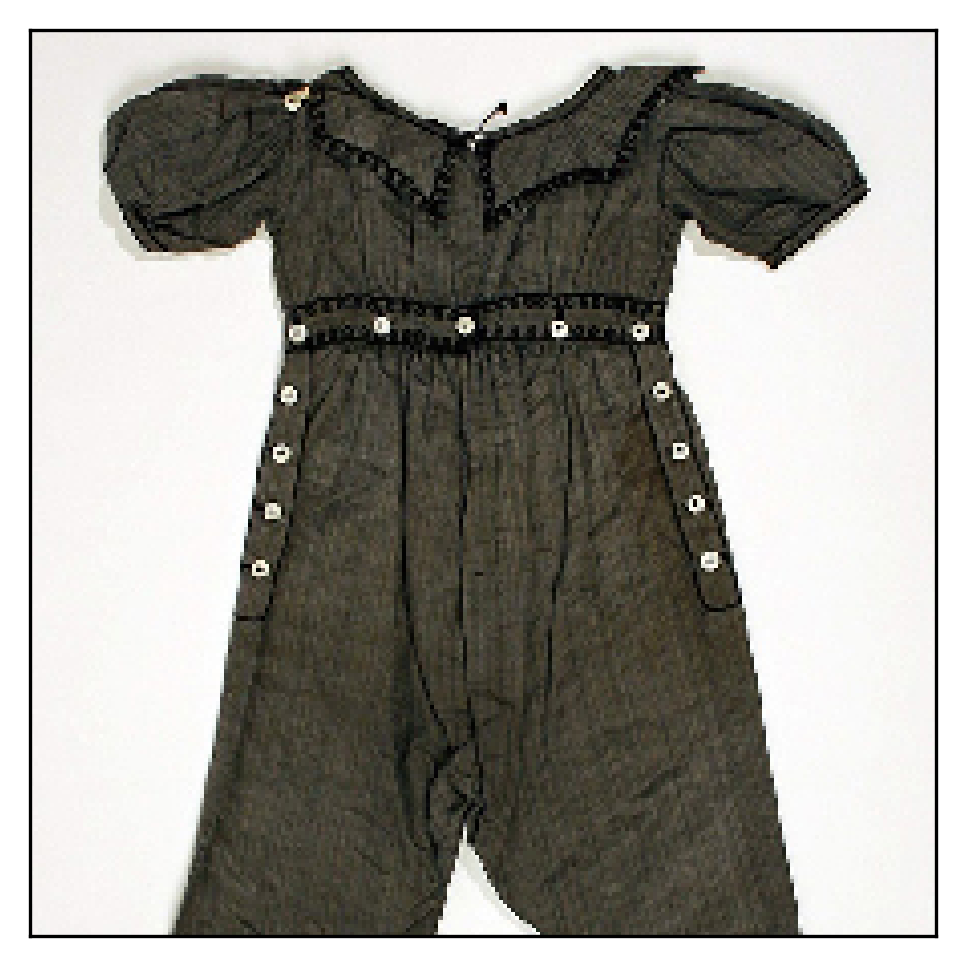} & 
\includegraphics[width=25mm]{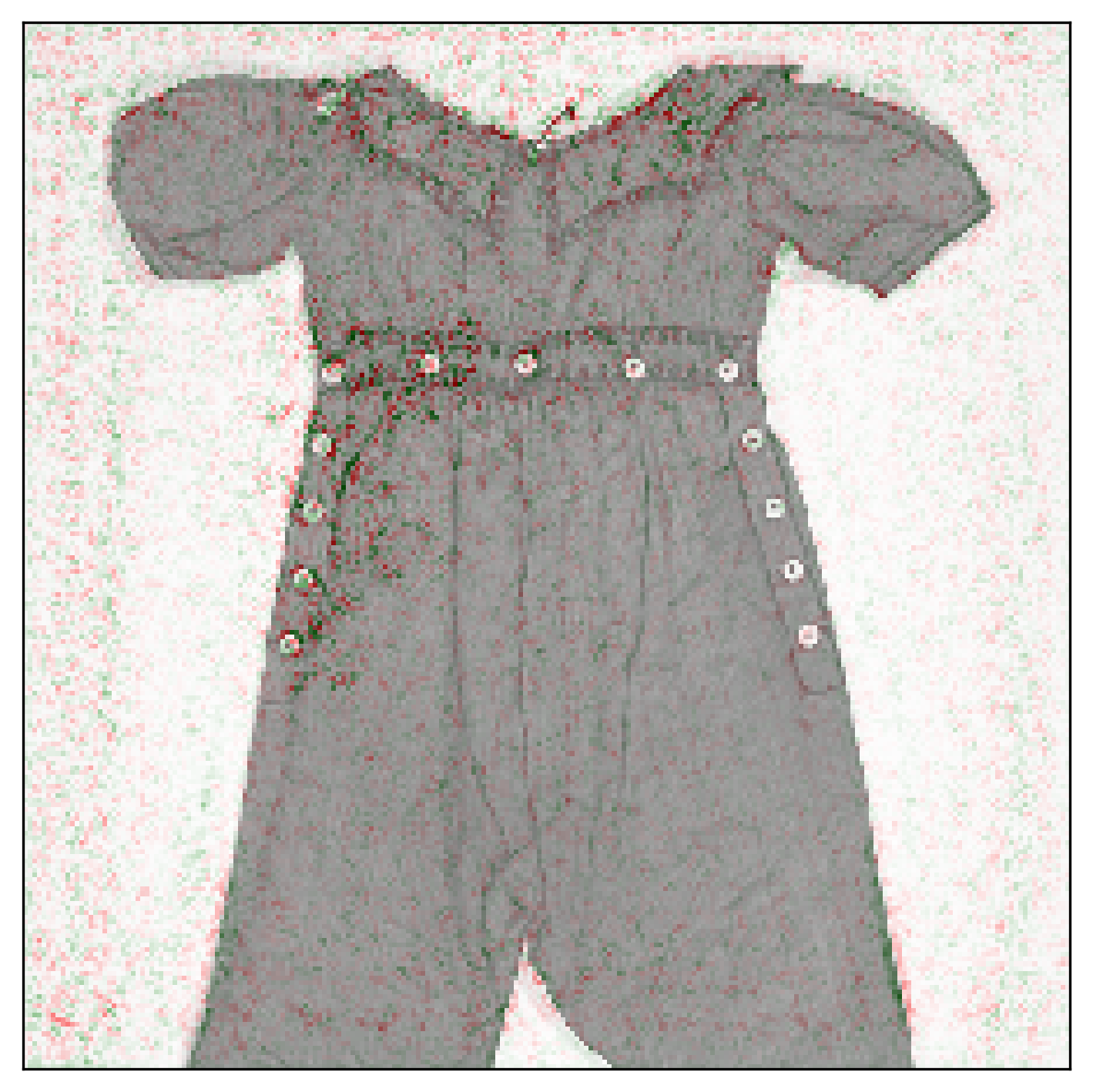} & 
\includegraphics[width=25mm]{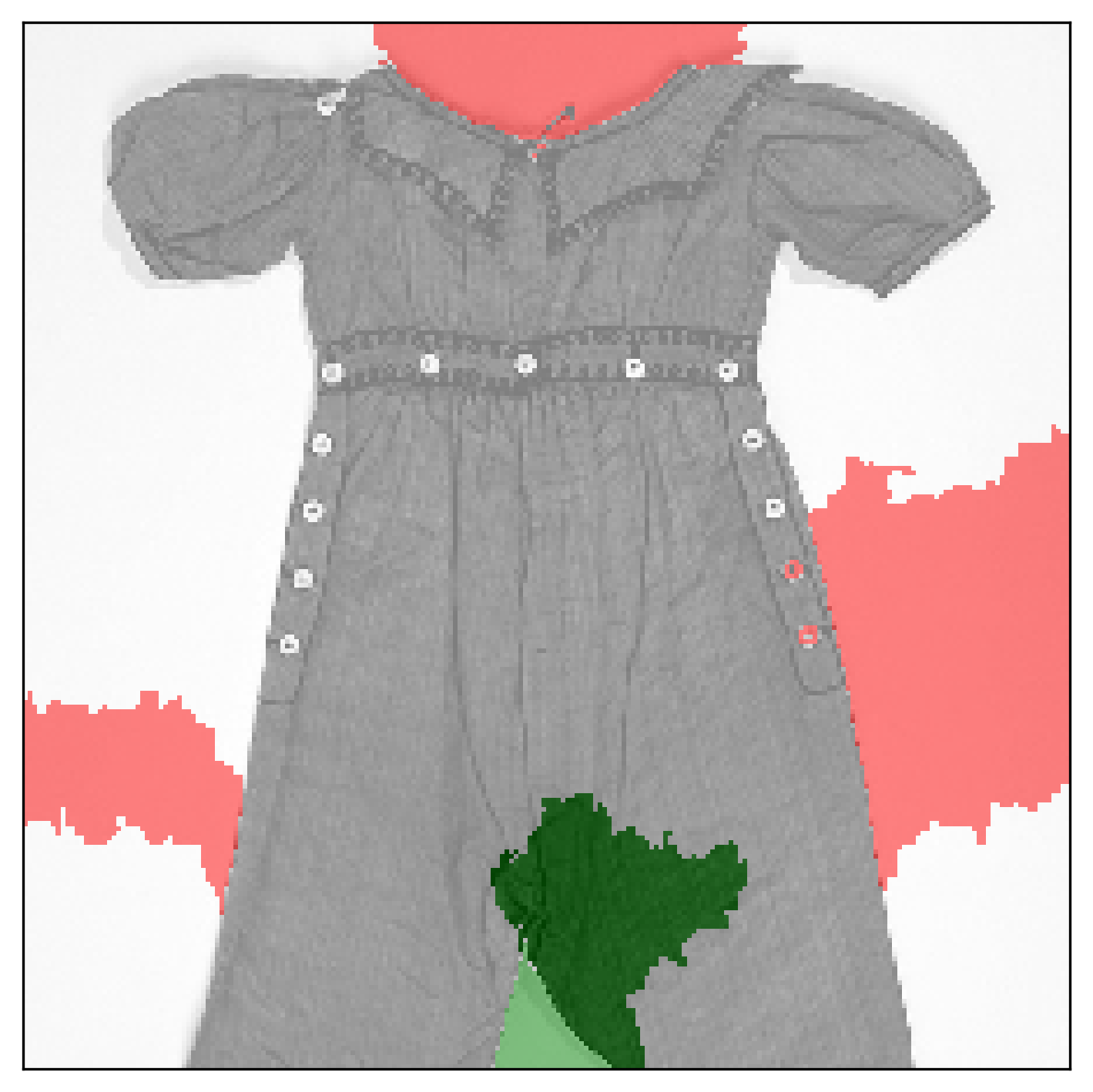} & 
\includegraphics[width=25mm]{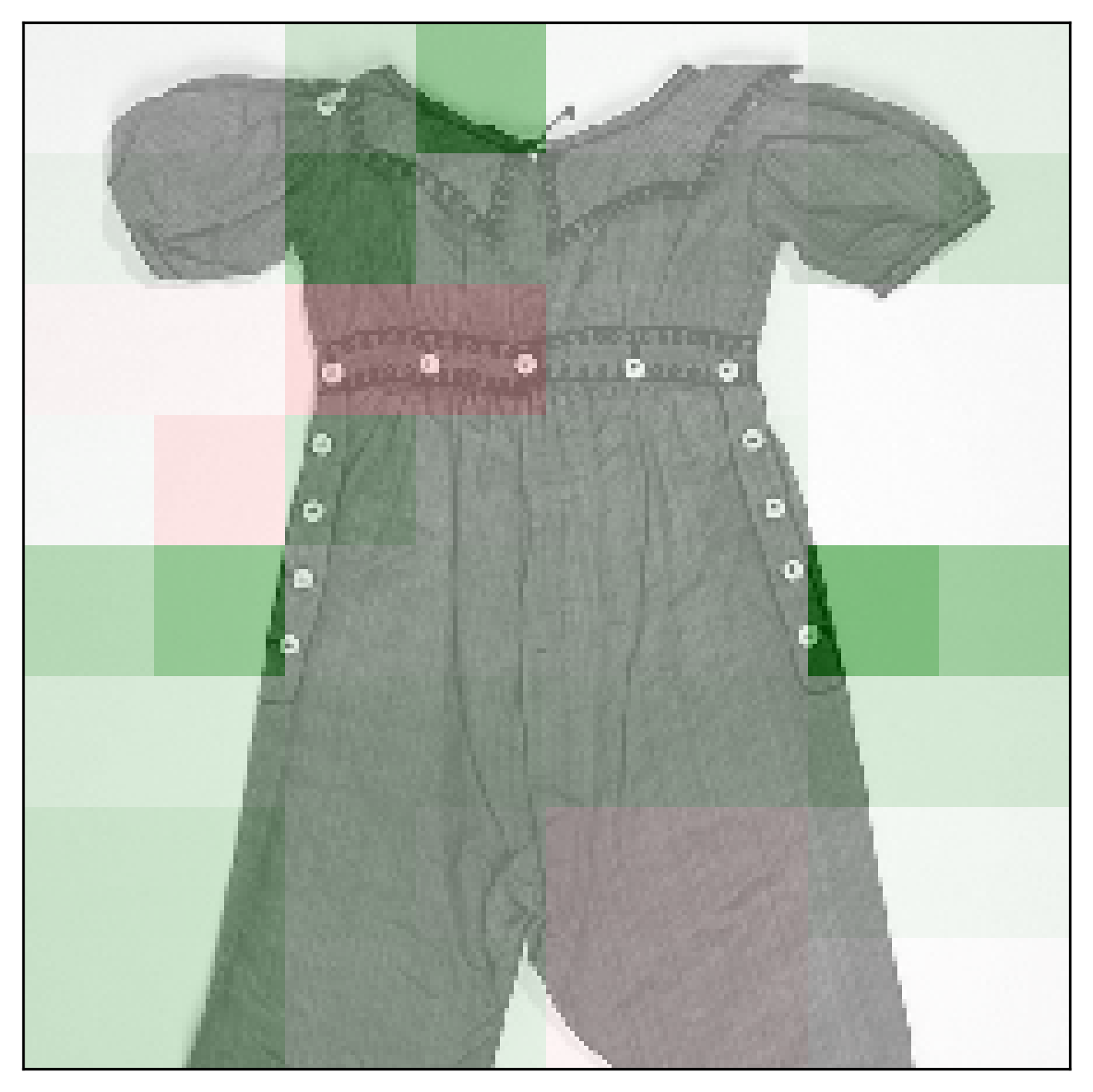} & 
\includegraphics[width=25mm]{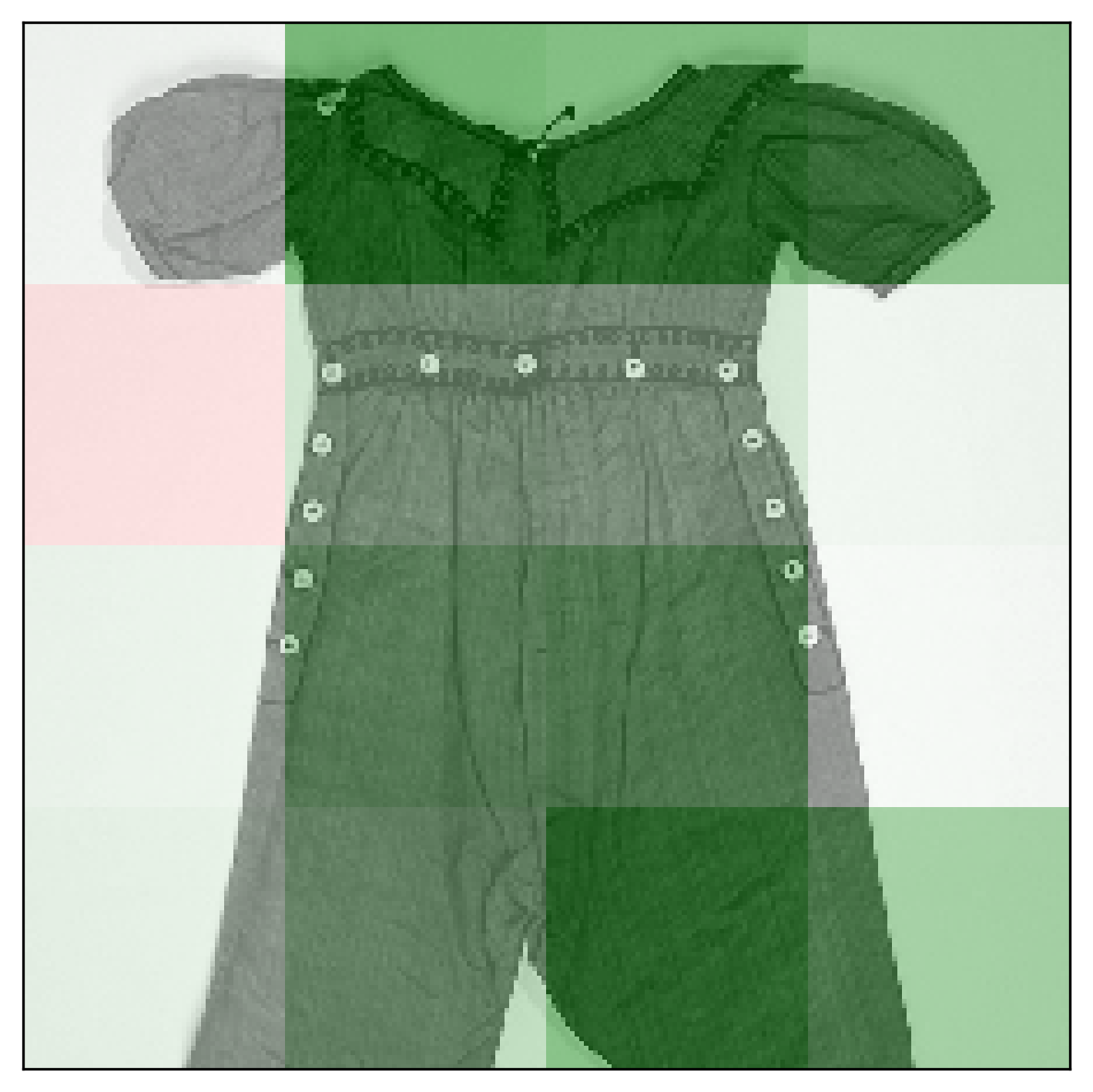} 
\\

\includegraphics[width=25mm]{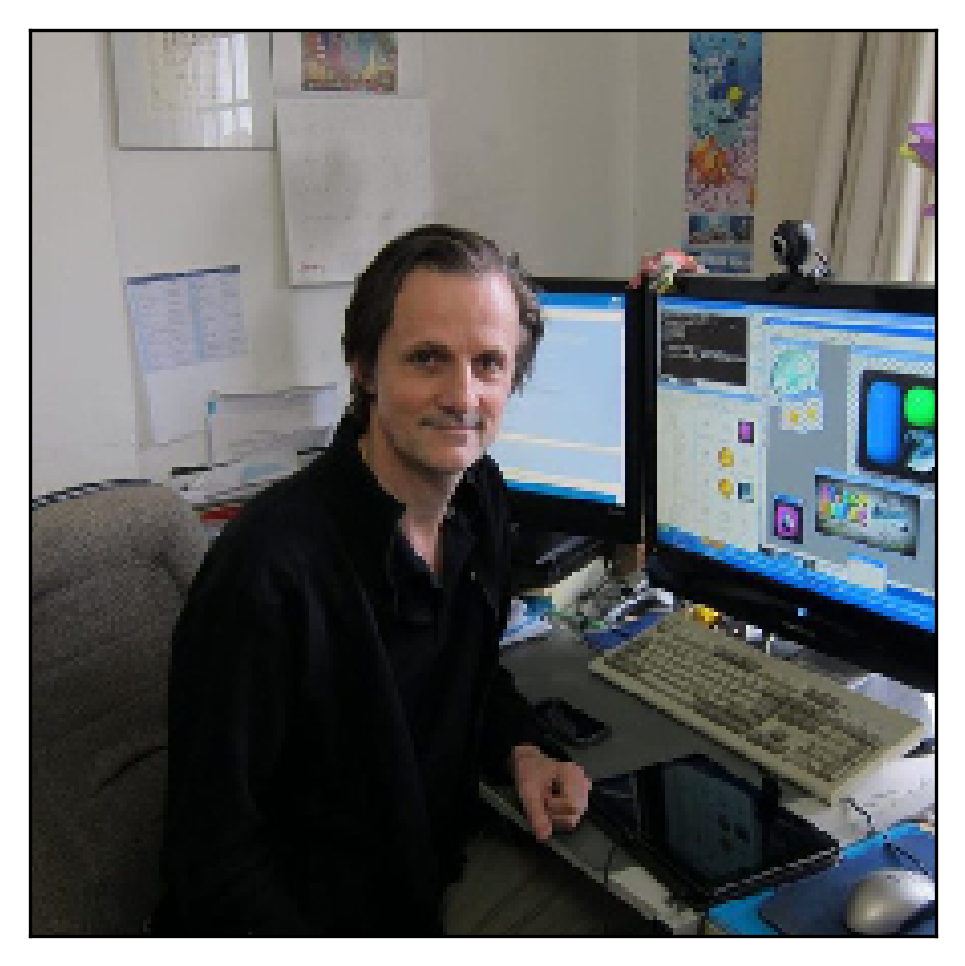} & 
\includegraphics[width=25mm]{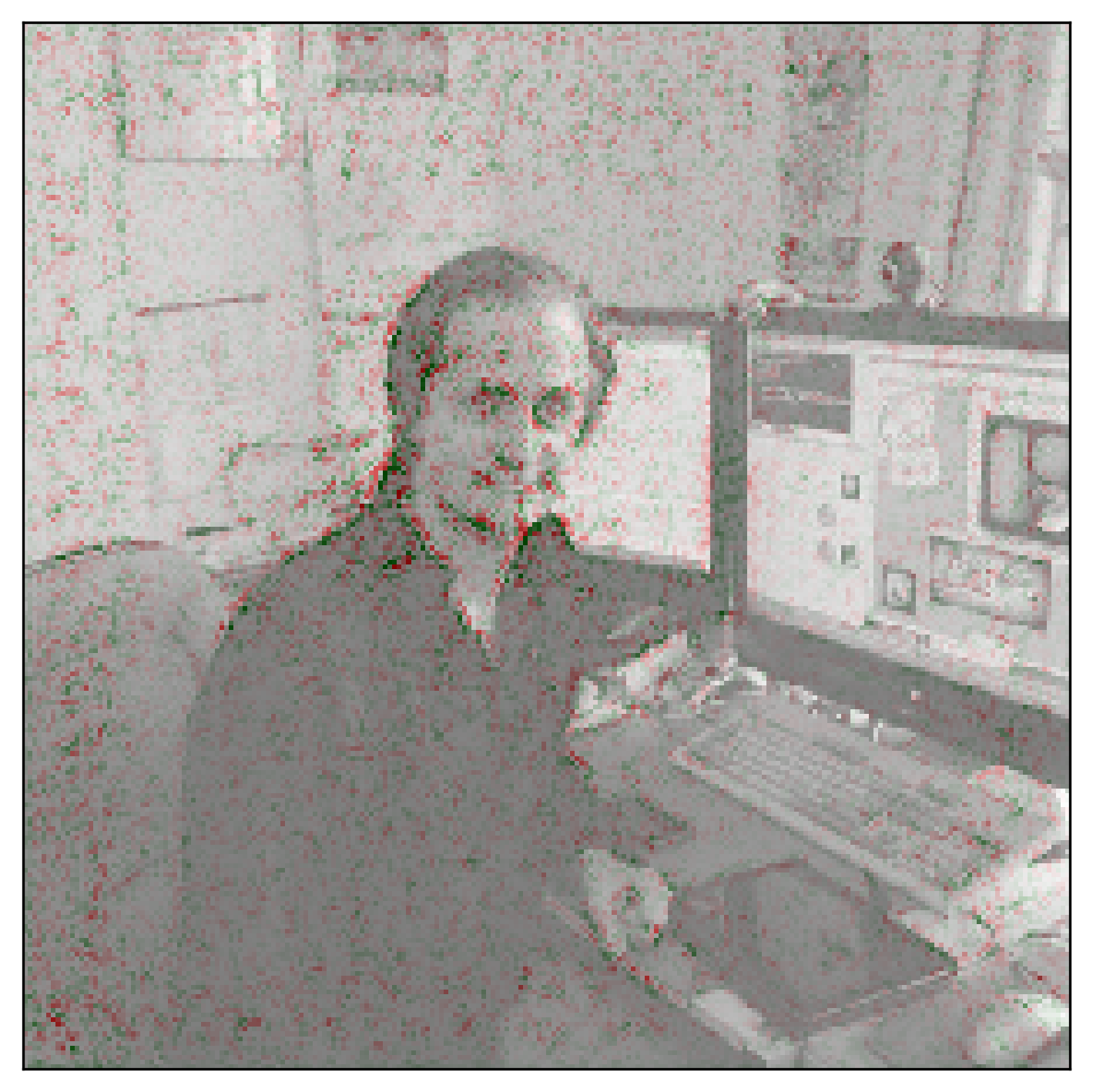} & 
\includegraphics[width=25mm]{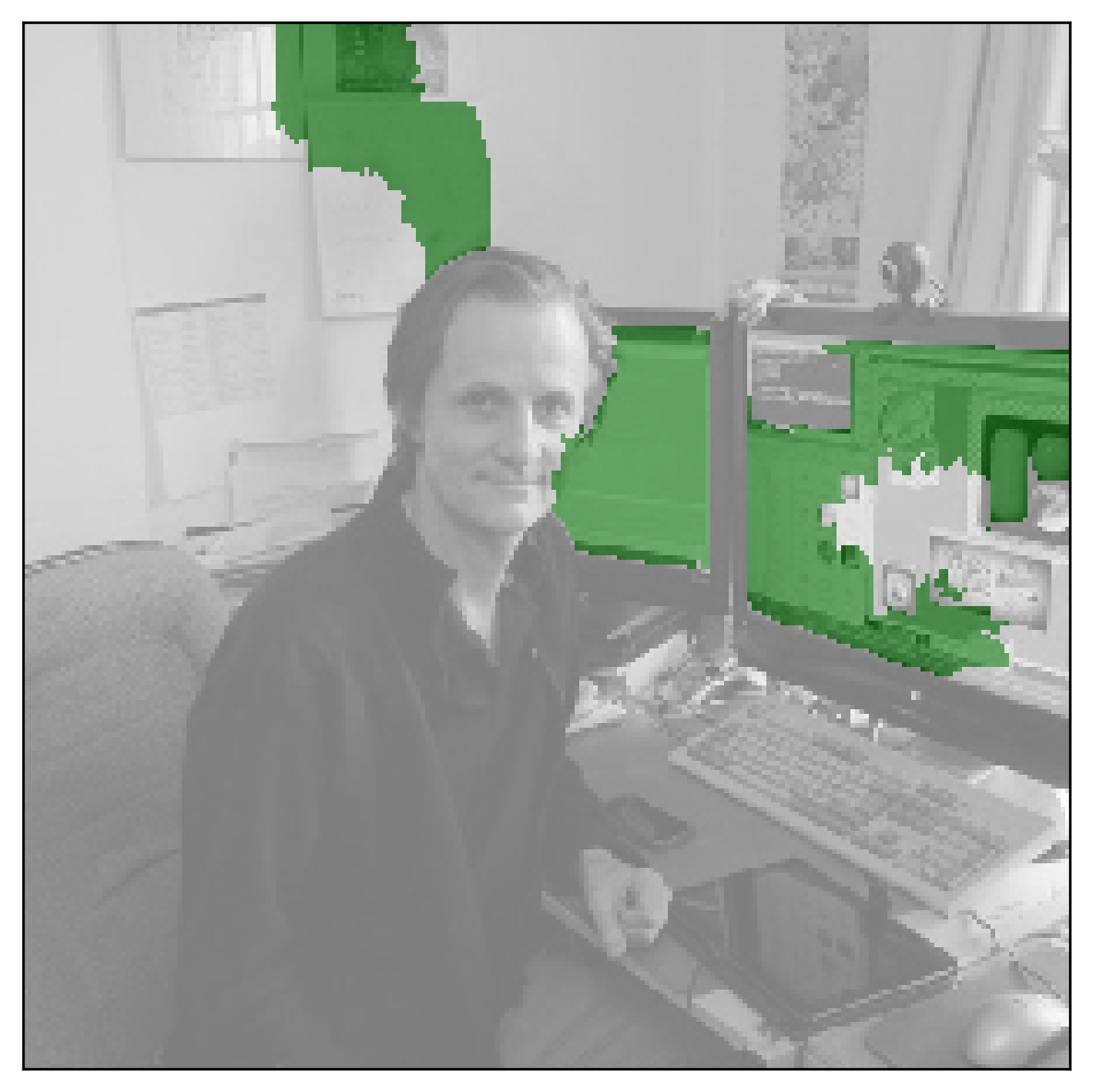} & 
\includegraphics[width=25mm]{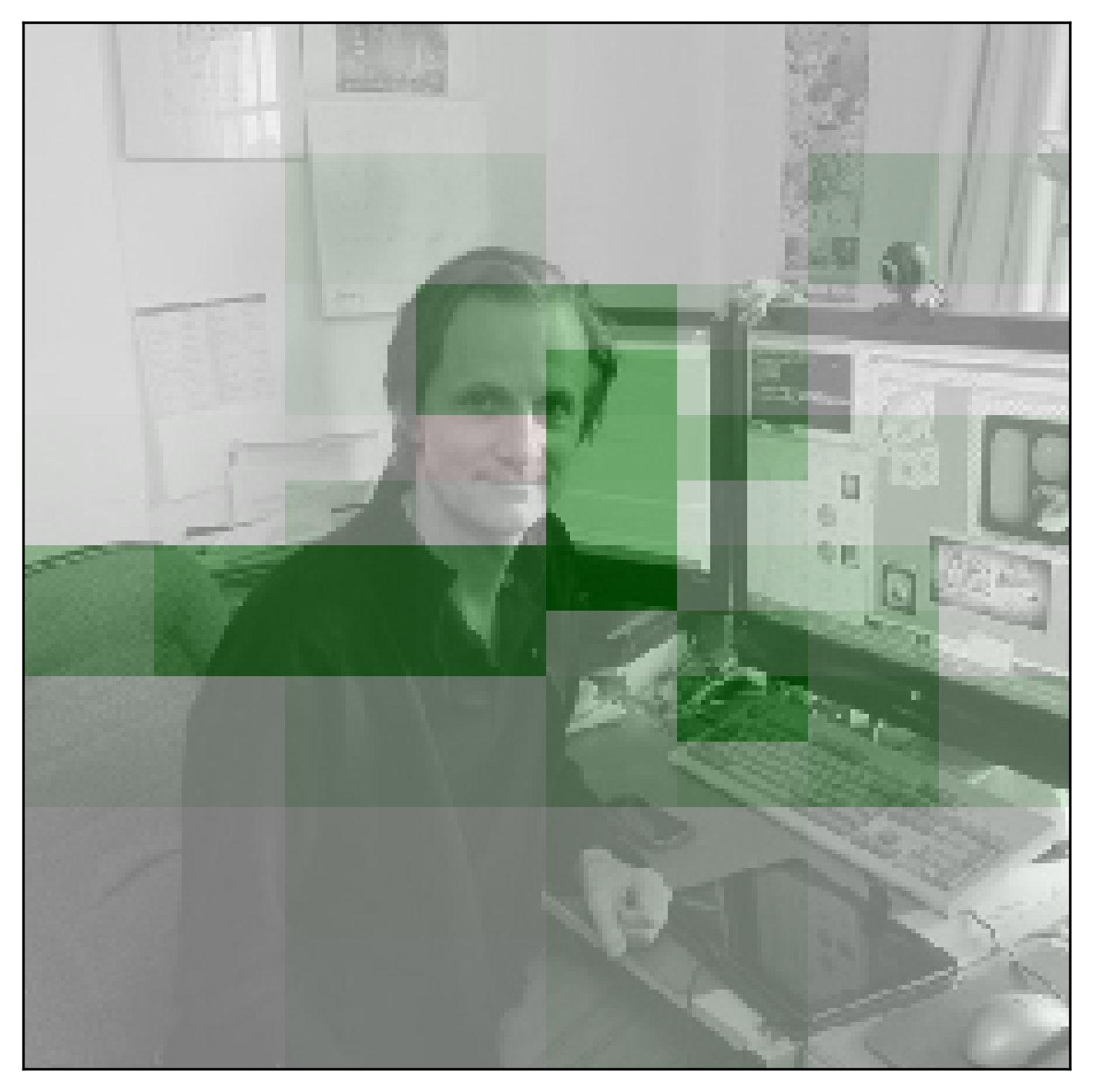} & 
\includegraphics[width=25mm]{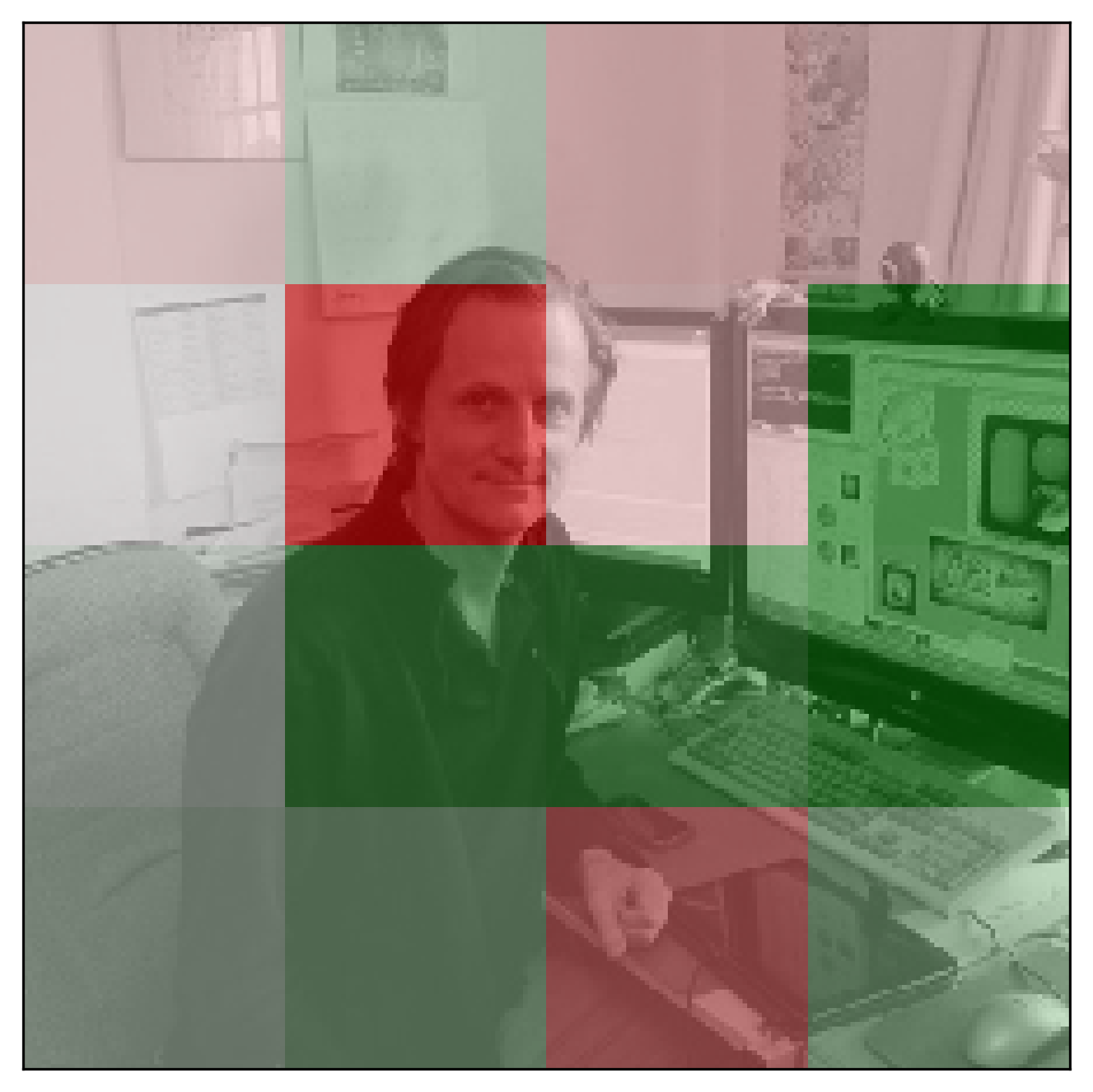} 
\\


\end{tabularx}
\caption{A Comparison of different state-of-the-art local explanation methods for a pre-trained ResNet-50 model~\cite{he2016deep} given random samples from the ImageNet dataset~\cite{deng2009imagenet}. Name of the original classes \textit{according} to the selected model (from top to bottom): \texttt{bittern}, \texttt{Indian elephant}, \texttt{hog}, \texttt{dowitcher}, \texttt{cardigan}, and \texttt{desktop computer}. 
The explanations from the RLE method are less noisy and rank almost all parts of the image with either positive (green) or negative (red) influence.
}
\label{tab:app_visual_experiments}
\end{table*}

\begin{table}[t]
    \centering
    \begin{tabular}{l|c}
    \toprule
    Method  &  Local Explanation  \\
    \midrule
    
    IG \cite{sundararajan2017axiomatic}  & \texttt{The new \hlc[green!50]{design} is \hlc[red!50]{awful}}\\
    LIME \cite{ribeiro2016should}  & \texttt{The new design is \hlc[red!50]{awful}}\\
    SHAP \cite{SHAP}  & \texttt{The new design \hlc[red!50]{is} \hlc[red!50]{awful}}\\
    IH \cite{janizek2021explaining}  & \texttt{The new design \hlc[red!50]{is} \hlc[red!50]{awful}}\\
    \textbf{RLE (ours)}          &  \texttt{The new design is  \hlc[red!50]{awful}} \\
    \midrule

    IG \cite{sundararajan2017axiomatic}  & \texttt{I \hlc[green!50]{love} you and I \hlc[red!50]{hate} you}\\
    LIME \cite{ribeiro2016should}  & \texttt{I \hlc[green!50]{love} you and I \hlc[red!50]{hate} you}\\
    SHAP \cite{SHAP}  & \texttt{I \hlc[green!50]{love} \hlc[green!50]{you} and I \hlc[red!50]{hate} \hlc[green!50]{you}}\\
    IH \cite{janizek2021explaining}  & \texttt{\hlc[red!50]{I} \hlc[green!50]{love} \hlc[green!50]{you} \hlc[green!50]{and} \hlc[red!50]{I}  \hlc[green!50]{hate} you}\\
    \textbf{RLE (ours)}          &  \texttt{I \hlc[green!50]{love} you \hlc[green!50]{and} I \hlc[red!50]{hate} you} \\
    \midrule

    IG \cite{sundararajan2017axiomatic}  & \texttt{\hlc[red!50]{I'm} not \hlc[green!50]{sure} if I \hlc[green!50]{like} the new \hlc[red!50]{design}}\\ 
   LIME \cite{ribeiro2016should}  &   \texttt{I’m \hlc[red!50]{not} sure \hlc[red!50]{if} I \hlc[green!50]{like} the new design}\\
    SHAP \cite{SHAP} &  \texttt{I’m \hlc[red!50]{not} \hlc[red!50]{sure} if I like the new design}\\ 
    IH \cite{janizek2021explaining} &  \texttt{I’m not sure if I \hlc[red!50]{like} \hlc[red!50]{the} \hlc[red!50]{new} design}\\ 
    \textbf{RLE (ours)}          &  \texttt{I’m \hlc[red!50]{not} \hlc[red!50]{sure} if I like the \hlc[red!50]{new} design} \\
    \midrule
    
    IG \cite{sundararajan2017axiomatic}  & \texttt{I \hlc[green!50]{really} like the \hlc[green!50]{new} design of \hlc[red!50]{your} \hlc[red!50]{website}}\\ 
    LIME \cite{ribeiro2016should}     &  \texttt{\hlc[green!50]{I} \hlc[green!50]{really} \hlc[green!50]{like} the new design of your website}\\ 
    SHAP \cite{SHAP}                  &   \texttt{I really \hlc[green!50]{like} the new design of your \hlc[green!50]{website}} \\
    IH \cite{janizek2021explaining}  &   \texttt{I \hlc[green!50]{really} \hlc[green!50]{like} the \hlc[green!50]{new} design of your \hlc[green!50]{website}}\\
    \textbf{RLE (ours)}          &       \texttt{\hlc[green!50]{I} \hlc[green!50]{really} \hlc[green!50]{like} the new \hlc[green!50]{design} of your website} \\
    \midrule

    IG \cite{sundararajan2017axiomatic}  & \texttt{\hlc[red!50]{The} bed was \hlc[green!50]{super} \hlc[green!50]{comfy}. \hlc[green!50]{The} chair wasn't bad, \hlc[red!50]{either}}\\
    LIME \cite{ribeiro2016should}  & \texttt{The \hlc[red!50]{bed} was \hlc[red!50]{super} \hlc[green!50]{comfy}. The chair wasn't bad, \hlc[red!50]{either}}\\
    SHAP \cite{SHAP}  & \texttt{The bed \hlc[red!50]{was} \hlc[green!50]{super} comfy. The chair \hlc[red!50]{wasn't} bad, \hlc[green!50]{either}}\\
    IH \cite{janizek2021explaining}  & \texttt{The bed was super comfy. The \hlc[green!50]{chair} wasn't bad, either}\\
    \textbf{RLE (ours)}          &  \texttt{The bed was super comfy. The chair \hlc[green!50]{wasn't} \hlc[green!50]{bad}, either} \\
    \midrule

    IG \cite{sundararajan2017axiomatic}  & \texttt{Terrible pitching \hlc[red!50]{and} awful \hlc[green!50]{hitting} led to \hlc[red!50]{another} crushing \hlc[green!50]{loss}}\\
    LIME \cite{ribeiro2016should}  & \texttt{\hlc[red!50]{Terrible} pitching and \hlc[red!50]{awful} hitting led to another \hlc[red!50]{crushing} \hlc[red!50]{loss}}\\
    SHAP \cite{SHAP}  & \texttt{\hlc[red!50]{Terrible} \hlc[red!50]{pitching} and \hlc[red!50]{awful} hitting led to \hlc[red!50]{another} crushing loss}\\
    IH \cite{janizek2021explaining}  & \texttt{Terrible pitching and \hlc[red!50]{awful} \hlc[red!50]{hitting}  led to \hlc[red!50]{another} crushing loss}\\
    \textbf{RLE (ours)}          &  \texttt{\hlc[red!50]{Terrible} \hlc[red!50]{pitching} and \hlc[red!50]{awful} \hlc[red!50]{hitting} \hlc[red!50]{led} to another crushing \hlc[red!50]{loss}} \\
    
    \bottomrule

    \end{tabular}
    \caption{A comparison of state-of-art feature attribution approaches to the presented RLE algorithm given a pre-trained DistilBERT model \cite{sanh2019distilbert} for the sentiment analysis task. We highlight the most important words according to each feature attribution method, where the green and red colors indicate the positive and negative impact, respectively. }
    \label{tab:text_bench3}
    \vspace{-6pt}
\end{table}


\end{document}